\documentclass[times, sageh]{TRR}
\usepackage{moreverb,url}
\usepackage[colorlinks,bookmarksopen,bookmarksnumbered,citecolor=blue,urlcolor=blue,linkcolor=blue]{hyperref}
\usepackage{enumitem}
\usepackage{siunitx}
\usepackage{subfigure}
\usepackage[ruled,vlined]{algorithm2e}
\SetAlCapFnt{\small}

\SetCommentSty{mycommfont}
\usepackage{tabularx}
\usepackage{multirow}

\newcolumntype{Y}{>{\centering\arraybackslash}X}
\DeclareMathOperator{\st}{s.t.}
\newcommand{\T}{^\mathsf{T}}
\newcommand\CC{C\nolinebreak[4]\hspace{-.05em}\raisebox{.2ex}{\relsize{0}{\text{++}}}}

\newcommand\BibTeX{{\rmfamily B\kern-.05em \textsc{i\kern-.025em b}\kern-.08em
T\kern-.1667em\lower.7ex\hbox{E}\kern-.125emX}}
\def\volumeyear{2025}

\begin{document}
\runninghead{Huang et al.}
\title{Passage-traversing optimal path planning with sampling-based algorithms}
\author{Jing Huang\affilnum{1,}\affilnum{2},  Hao Su\affilnum{3,}\affilnum{4} and Kwok Wai Samuel Au\affilnum{1,}\affilnum{2}
}
\affiliation{\affilnum{1}Department of Mechanical and Automation Engineering, The Chinese University of Hong Kong, Hong Kong, China  \\
\affilnum{2}Multi-Scale Medical Robotics Center, Hong Kong, China \\
\affilnum{3}Department of Computer Science and Engineering, University of California, San Diego, CA, USA \\
\affilnum{4}Hillbot Inc., USA
}
\corrauth{Jing Huang, Department of Mechanical and Automation Engineering, The Chinese University of Hong Kong, Shatin, N.T., Hong Kong, China \\ Email:\href{mailto:huangjing@mae.cuhk.edu.hk}{huangjing@mae.cuhk.edu.hk}}

\begin{abstract}
This paper introduces a new paradigm of optimal path planning, i.e., passage-traversing optimal path planning (PTOPP), that optimizes paths' traversed passages for specified optimization objectives. In particular, PTOPP is utilized to find the path with optimal accessible free space along its entire length, which represents a basic requirement for paths in robotics. As passages are places where free space shrinks and becomes constrained, the core idea is to leverage the path's passage traversal status to characterize its accessible free space comprehensively. To this end, a novel passage detection and free space decomposition method using proximity graphs is proposed, enabling fast detection of sparse but informative passages and environment decompositions. Based on this preprocessing, optimal path planning with accessible free space objectives or constraints is formulated as PTOPP problems compatible with sampling-based optimal planners. Then, sampling-based algorithms for PTOPP, including their dependent primitive procedures, are developed leveraging partitioned environments for fast passage traversal check. All these methods are implemented and thoroughly tested for effectiveness and efficiency validation. Compared to existing approaches, such as clearance-based methods, PTOPP demonstrates significant advantages in configurability, solution optimality, and efficiency, addressing prior limitations and incapabilities. It is believed to provide an efficient and versatile solution to accessible free space optimization over conventional avenues and more generally, to a broad class of path planning problems that can be formulated as PTOPP. 
\end{abstract}
\maketitle

\section{Introduction}
\noindent Motion and path planning makes a fundamental component in robotics and also many other disciplines. It resolves the basic problem of finding feasible paths connecting the start and goal states in a configuration space populated with infeasible portions, i.e., obstacles. Though path planning is hard from the algorithmic and computational perspectives, its prominent importance has driven the development of diverse path planning methods over the past decades, e.g., grid-based search, artificial potential fields, and sampling-based algorithms \citep{lavalle2006planning, gonzalez2015review, paden2016survey, orthey2023sampling}. Early studies mainly addressed path feasibility and only aimed at collision-free paths. Path optimality was not explicitly considered or simply referred to as path length. Up to date, the dominant planning paradigm has shifted to optimal path planning which seeks feasible paths optimizing specified objectives \citep{karaman2011sampling, orthey2023sampling}.

Despite the prevalence of optimal path planning, limited objectives have been investigated. Path length, formally defined as the total variance of a path, is the most popular objective because it commonly reflects the time or cost to traverse the path. In practice, there are many other desired path properties such as good path smoothness and reasonable terrain selection \citep{rowe2000finding}. A fundamental one among these is to provide sufficient accessible free space along the entire path length \citep{sakcak2021complete, huang2023deformable, huang2024homotopic}. Accessible free space is reminiscent of the broadly used path clearance, the shortest distance between the path and obstacles, but represents a more general and comprehensive concept to gauge holistic accessible free space for a path. The requirement for it arises commonly in path planning in order to accommodate robots and their motions. For instance, manipulation in cluttered sites prefers large free space along the manipulated object's path to reduce collision risk and manipulation complexity \citep{huang2023deformable, huang2024homotopic}. In navigation, large accessible free space along paths is required for the safety and agility of mobile robots \citep{mao2024optimal, wang2025fast}. However, optimizing accessible free space in path planning remains under-investigated.

To grant sufficient free space in path planning, clearance between robots and obstacles is embedded by inflating their volumes in collision check, e.g., \citep{lynch2017modern, paliwal2018maximum}. Alternatively, path clearance is formulated as a path cost term \citep{jaillet2010sampling, davoodi2015clear, davoodi2020path} or gets enlarged by retracting planned paths \citep{geraerts2007creating}. Many works aim to find the maximum clearance, e.g., using visibility-Voronoi complex \citep{wein2007visibility}, corridors in Voronoi diagrams \citep{geraerts2010planning}, and medial axes of the free space \citep{lien2003general, denny2014marrt}. Complex processing is required in geometric computations, costmap building, and path adjustments in these methods. They also suffer from restrictive scalability in high-dimensional spaces, even in 3D space. More importantly, clearance cannot characterize the holistic accessible free space since it only reflects the most constrained situation. Maximizing clearance alone does not ensure adequate free space along the entire path.

Using traversed passages for a comprehensive characterization of the path's accessible free space, passage-traversing optimal path planning (PTOPP) is proposed as a new path planning paradigm in this article, which gives rise to an array of new perspectives and tools in path planning. The primary novelty lies in utilizing the path's passage-traversing status to extract free space accessibility comprehensively. By querying passed passages, an accurate assessment of the accessible free space is established and enables its global optimizability. Firstly, fast passage detection and environment cell decomposition methods are proposed by introducing proximity graphs of obstacles. Optimal path planning with accessible free space objectives is then formulated as PTOPP problems compatible with sampling-based optimal planners. In PTOPP implementation, the core to efficiency is rapid passage traversal check for paths. This is achieved by positioning samples in decomposed cells. To validate its effectiveness and efficiency, this scheme is extensively tested in various setups.  Beyond accessible free space optimization, the proposed algorithms present a systemic and general pipeline for a broad class of path planning problems that can be described by PTOPP. To sum up, the main contributions are as follows.
\begin{enumerate}[leftmargin=*, noitemsep, nolistsep]
    \item A novel and general passage detection and associated free space decomposition method based on proximity graphs. The detected passages and cells are intuitive, sparse, and informative in characterizing the obstacle distribution and free space accessibility. This decomposition enables fast passage-traversing check for paths. 
    \item PTOPP problem formulations compatible with sampling-based optimal planners. Typical accessible free space optimization occasions are formulated as PTOPP problems defined by their path costs. The optimizability condition for path costs is addressed.
    \item Efficient algorithms for PTOPP. 
        Fast passage and cell detection using Gabriel graphs is introduced. In PTOPP, the core module of rapid passage traversal check for paths is realized by positioning samples in subdivided cells, making PTOPP efficient.
    \item Elaborate algorithm implementation and extensive experimental results for effectiveness and efficacy validation. For the common 2D and 3D workspace, all the proposals are implemented for thorough performance tests, which reveals the generality and applicability of PTOPP as a new paradigm of path planning.
\end{enumerate}

A preliminary study called passage-aware optimal path planning was reported as a submodule in homotopic path set planning in \citep{huang2024homotopic}. This article investigates PTOPP independently and systematically, contextualizing it in general path planning problems. Substantial new theories, methods, and algorithms make PTOPP a self-contained new path planning paradigm.

\section{Related work}
A set of contributed studies are related to this work, of which we mainly draw from literature on narrow passage problems, path clearance, and cell decomposition.

\subsection{Passages in path planning}
While there is no unified definition of passages, they are generally recognized as passable regions in the configuration space whose removal changes the connectivity of the space \citep{hsu1998finding, sun2005narrow, wang2018learning, ruan2022efficient}. Specific characterizations of passages exist such as the notion of visibility sets \citep{barraquand1996random, hsu1997path} and concatenation of a series of maximal spheres in free space \citep{guo2023obstacle}. Frequently studied is narrow passage detection. The bridge detection tests based on the randomized bridge builder (RBB) \citep{hsu2003bridge, sun2005narrow, wang2010triple} and its variants, e.g., the randomized star builder (RSB) \citep{zhong2013robot}, bridge line tests \citep{lee2014selective}, and the grid bridge builder (GBB) \citep{guo2023obstacle}, are arguably the most prominent method. If the midpoint of two nearby in-collision points is collision-free, its neighborhood is marked as a passage region. Narrow passages are also identified as regions of large velocities in potential flows \citep{kazemi2004robotic}, high potentials in artificial potential fields \citep{aarno2004artificial}, and watersheds in decomposed space \citep{van2005using}. A sample is considered in a narrow passage if the probabilistic roadmap (PRM) connected component it lies in has a small size \citep{shi2014spark}. Using density-based spatial clustering, passage channels are detected as the sample clusters in narrow-passage-aware Gaussian sampling processes \citep{huang2024robot}.

Narrow passages impose severe challenges for sampling-based path planning since the sampling probability sharply deteriorates inside them, rendering it hard to capture the connectivity of the sampled configuration space. Most works are contributed to sampling strategies for efficient path finding in the presence of narrow passages. In \citep{hsu1998finding}, infeasible samples and connections of PRMs in the dilated free space are adjusted to the original free space via local resampling. Similarly, the free space is expanded by shrinking robots and obstacles in roadmap construction, and colliding portions are repaired by retraction in \citep{saha2005finding, hsu2006multi}. For sampling strategies, the Gaussian sampler is proposed in \citep{boor1999gaussian} in order to densify samples close to obstacles for better space coverage. Obstacle-based PRM (OBPRM) \citep{amato1998obprm, yeh2012uobprm} generates samples near obstacles for better connectivity in narrow passages. Another common avenue is to increase the sampling probability in detected narrow passages \citep{kazemi2004robotic, aarno2004artificial, van2005using} or to bias sampling to passages \citep{zucker2008adaptive}.
Exploiting obstacle or sample distribution information, the expansion methods of rapidly-exploring random tree (RRT) in different positions are designed in \citep{rodriguez2006obstacle, tahirovic2018rapidly} to move through narrow areas efficiently. Retracting invalid samples to obstacle boundaries \citep{zhang2008efficient} or passages \citep{lee2014selective} also helps to find paths traversing passages. In Triple RRTs \citep{wang2010triple, zhong2012triple}, an RRT rooted at the midpoint in the bridge test attempts to connect to two RRTs rooted at the start and goal respectively in RRT-Connect. The analogous idea of growing RRTs in narrow passages is utilized within PRMs in \citep{shi2014spark}. Explicit geometrical modeling of robots, environments \citep{ruan2022efficient}, and learning-based approaches for sampling and planning \citep{wang2018learning, li2023sample, li2024sampling} are also employed to address narrow passage problems.

\subsection{Path clearance in path planning}
Admitting sufficient accessible free space along the path is a fundamental requirement in robot path planning. This requirement is commonly characterized by path clearance, i.e., the shortest distance between the path and obstacles. A basic avenue to attaining clearance is by dilating obstacles and robots in collision check \citep{lynch2017modern, paliwal2018maximum}. The resulting path is automatically endowed with the prespecified clearance. Many works achieve this by constructing generalized Voronoi diagrams (medial axes of the free space) \citep{mark2008computational, geraerts2010planning}. Paths are searched on Voronoi diagrams to possess maximum clearance \citep{o1985retraction, bhattacharya2008roadmap}. A hybrid diagram of the visibility graph and the Voronoi diagram enables the tradeoff between path length and clearance \citep{wein2007visibility}. Without explicitly computing medial axes, Medial Axis PRM (MAPRM) \citep{wilmarth1999maprm, lien2003general} and Medial Axis RRT (MARRT) \citep{denny2014marrt} adapt path nodes towards medial axes so that the formed roadmaps and trees have large clearance. Uniformly sampling on medial axes in PRM is presented in \citep{yeh2014umaprm}. In \citep{guibas1999probabilistic,holleman2000framework}, samples in PRM are restricted near the medial axes to efficiently sample narrow areas and increase path clearance. Similarly, skeleton-biased locally-seeded RRT (skilled-RRT) \citep{dong2018faster} leverages the geometric path on medial axes to bias RRT growth locally.

Alternatively, clearance is optimized in path postprocessing locally performed without changing the path's homotopy class. The external potential field is applied to push paths away from obstacles and maintain clearance in \citep{brock2002elastic}. Path clearance gets increased by retracting path points to medial axes \citep{geraerts2007creating} or introducing path points of larger clearance \citep{davoodi2015clear}. Apart from explicitly specifying or adjusting path clearance, optimization problems are often formulated. Given roadmaps or grid maps, multi-objective graph-search, e.g., multi-objective A$^*$ \citep{stewart1991multiobjective, mandow2005new}, bi-objective Dijkstra's algorithm \citep{sedeno2019biobjective}, and multi-objective evolutionary methods \citep{castillo2007multiple, ahmed2013multi}, are exploited with a clearance objective. For continuous space, optimal RRT (RRT$^*$) and multi-objective evolutionary optimization are blended to asymptotically produce Pareto optimal solutions in \citep{yi2015morrf}. The weighted measure minimizing path length while maximizing path clearance is utilized in path search on Voronoi diagrams in \citep{wein2008planning}. Pareto optimal solutions of the same bi-objective are found by a modified Dijkstra's algorithm in \citep{davoodi2020path}. The integral of the reciprocal of clearance over the path length is minimized by sampling-based planners \citep{strub2022adaptively}. By introducing the Pareto optimal paths graph adopted from the visibility-Voronoi complex, the complete set of Pareto optimal solutions is obtainable in \citep{sakcak2021complete}, making the tradeoff between path length and clearance convenient to take.

\subsection{Cell decomposition in path planning}
Another relevant research topic is cell decomposition for path planning. It is a common practice to decompose the free space into a union of simple and non-overlapping subdivisions, i.e., cells, for better solvability in path planning problems \citep{brooks1985subdivision}. Grid search and sampling-based path planning methods are built on bitmap environment representation with uniform cells \citep{lavalle2004relationship, tsardoulias2016review} or regular cells of adaptive sizes \citep{chen1995planning, van2005using, lingelbach2004path, patle2019review}. Since the environment may not be precisely characterized by grids, e.g., irregular boundaries, grid decomposition is classified as approximate decomposition \citep{choset2001coverage, siegwart2011introduction, galceran2013survey}. More related to this work is exact cell decomposition that exploits environment features to obtain a complete environment representation. Classical decomposition approaches include triangulation and trapezoidal/vertical decomposition in 2D space as well as cylindrical decomposition in higher dimensional spaces \citep{lavalle2006planning, kloetzer2011software, mark2008computational, siciliano2016springer, aggarwal2020path}. They utilize geometric features in polygonal obstacle regions, such as vertices, to construct cells. The adjacency graph encoding cell adjacency relationships is employed in graph-search path planning. Roadmaps are also built using nodes located in cells, e.g., centroids and boundary midpoints, as milestones for path planning. There exist other avenues like Morse decomposition \citep{choset1998coverage, acar2002morse} and visibility-based decomposition \citep{lavalle1997finding, choset2005principles} based on changes of space connectivity and line of sight, which are complex to implement. A systematic comparison of different cell types, graph weights, and waypoint selections on cells in cell decomposing-based path planning can be found in \citep{kloetzer2015optimizing, gonzalez2017comparative}.

Path planning based on cell decomposition is commonly restrictive regarding path adjustability. In addition, most cell decomposition methods scale poorly in high-dimensional space, even 3D space. While partitioned cells make feasible paths easy to find and constraints, e.g., obstacle avoidance, easy to impose \citep{deits2015efficient, ryou2021multi}, paths are not easily optimizable even for basic metrics such as the path length \citep{kloetzer2015optimizing}. Cell decomposition generally gives rise to coarse paths returned by the shortest path search in grids or graphs. Integration with other optimal planners to unleash its potential in optimal path planning has been under-explored. Conventional usages of cells for roadmap construction \citep{lavalle2006planning, mark2008computational, siciliano2016springer, aggarwal2020path} and direct integration in sampling-based path planning methods as samples \citep{rosell2005path, samaniego2017uav} suffer from low resolution unable to fulfill demanding planning requirements well. This article presents a novel and natural cell decomposition resulting from the narrow passage detection results, which is utilized in sampling-based optimal planners for fast sample positioning. Path's passage traversal status is efficiently retrieved in PTOPP.

\section{Passage detection and cell decomposition using Gabriel graphs}
\label{Sec: Passage Detection}
This section first presents a general PTOPP formulation as background. As essential preprocessing, an efficient passage detection and free space decomposition procedure based on proximity graphs of obstacles is then introduced.

\subsection{Passage-traversing optimal path planning (PTOPP) problems}
Consider an optimal path planning (OPP) problem from the start position $\mathbf{x}_{0}$ to the goal position $\mathbf{x}_{g}$. The configuration space $\mathcal{X}$ is populated with $m$ obstacles $\mathcal{O}_1, ..., \mathcal{O}_m$ which constitute the obstacle region $\mathcal{X}_{obs} = \cup_{i = 1}^m \mathcal{O}_i$. Both $\mathbf{x}_0$ and $\mathbf{x}_g$ are in the obstacle-free space $\mathcal{X}_{free} = \text{cl}(\mathcal{X} \backslash \mathcal{X}_{obs})$ where cl$(\cdot)$ denotes the closure of a set. Our development is within the common workspace, $\mathcal{X} \subset \mathbb{R}^2$ and $\mathbb{R}^3$. Formally, a collision-free path is a continuous function $\sigma: [0, 1] \mapsto \mathcal{X}_{free}$ and the path length is defined as 
\begin{equation}
    \text{Len}(\sigma) = \sup_{\{0 = \tau_0 < \tau_1 < ... < \tau_n=1\}} \sum_{i = 1}^n \| \sigma(\tau_i) - \sigma(\tau_{i - 1}) \|
\end{equation}
where $\| \cdot \|$ denotes 2-norm. The path argument $\tau \in [0,1]$ is given by path length parameterization so that Len($\sigma(0 \sim \tau)$) = $\tau$Len($\sigma$) with $\sigma(0 \sim \tau)$ being the path segment from $\sigma(0)$ to $\sigma(\tau)$. Let $\Sigma_{free}$ be the set of collision-free paths. A path $\sigma \in \Sigma_{free}$ is feasible if it connects the start and goal, i.e., $\sigma(0) = \mathbf{x}_0, \sigma(1) = \mathbf{x}_g$. The cost function $f_c: \Sigma_{free} \mapsto \mathbb{R}$ encodes the cost associated with a path to depict the path goodness. OPP aims to find a feasible path minimizing $f_c(\cdot)$. 
Using length parameterization, a concatenated path is 
\begin{equation}
    (\sigma_1 | \sigma_2) (\tau) = 
    \begin{cases}
        \sigma_{1}(\tau / r)                & \tau \in [0, r]  \\
        \sigma_{2}((\tau - r) / (1 - r))    &\tau \in (r, 1]
    \end{cases}
\end{equation}
where $\sigma_1 | \sigma_2$ is the concatenated path of $\sigma_1$ and $\sigma_2$ satisfying $\sigma_1(1) = \sigma_2(0)$. The ratio $r = $Len($\sigma_1$) $/$ Len($\sigma_1 | \sigma_2$).
\begin{figure}[t]
    \minipage{1 \columnwidth}
     \centering
     \includegraphics[width= 0.83 \columnwidth]{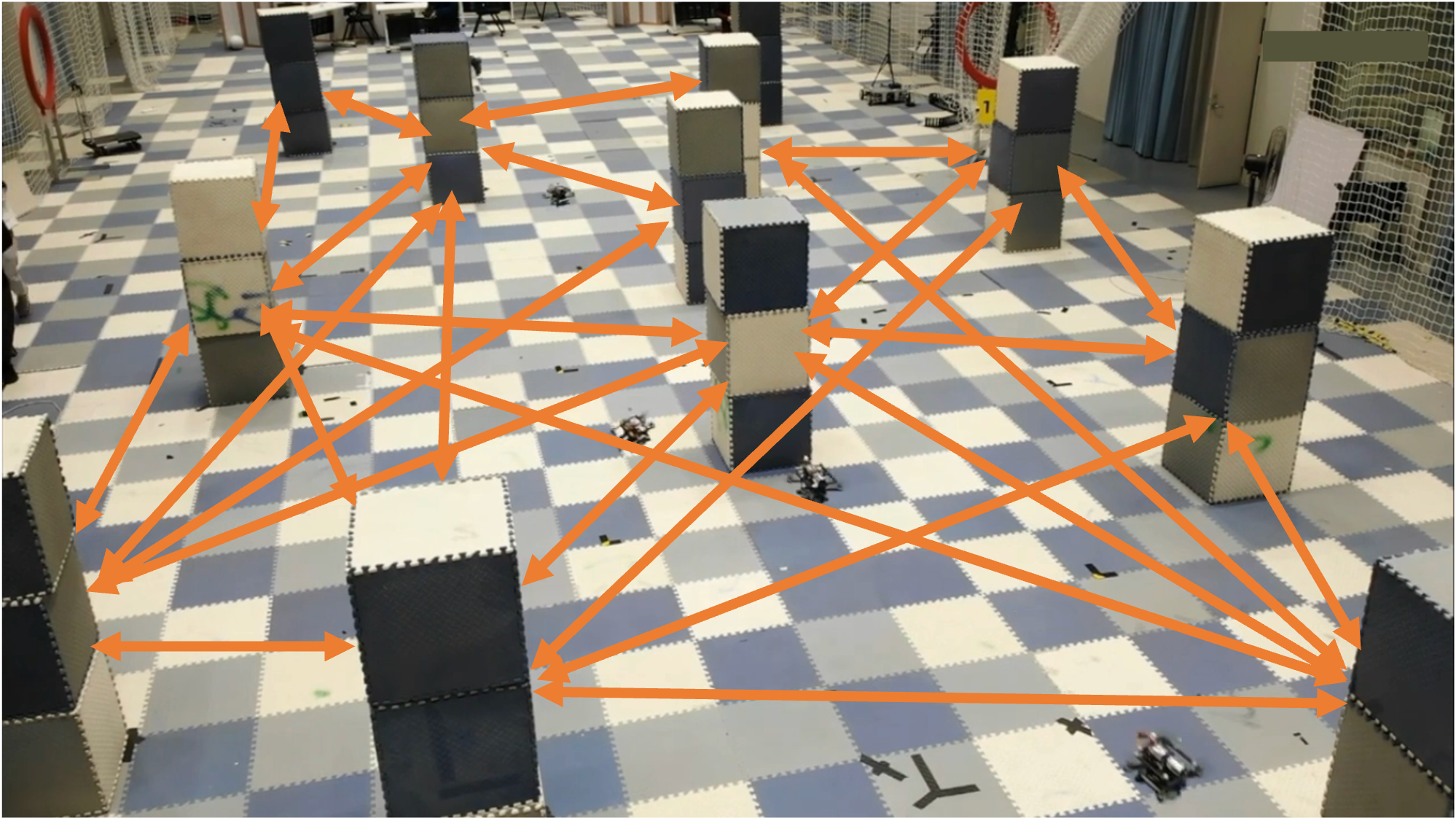}
     \endminipage \hfill
     \caption{Example of dense passage distributions (orange segments) using the pure visibility condition in a laboratory for drone swarm flight, e.g., \citep{mao2024optimal}.}
     \label{Fig: Passage Example in Lab}
\end{figure}

PTOPP problems are a class of OPP that explicitly embed the path's passage traversal status into the path cost $f_c(\cdot)$. Though a unified passage definition is unavailable, a passage $\mathcal{P}$ can be defined as a subset of $\mathcal{X}_{free}$ constructed by a pair of obstacles under some conditions. A path $\sigma$ may traverse $\mathcal{P}$ or not. Cases where $\sigma$ partially or entirely lies in a passage are trivial. Passages passed by $\sigma$ from the start $\sigma(0)$ to a path point $\sigma(\tau)$ form an ordered list $P_\sigma(\tau) = \{\mathcal{P}_i, ..., \mathcal{P}_k\}$ and $P_\sigma(\tau, i)$ indexes the $i$-th passage in $P_\sigma(\tau)$. PTOPP has $f_c(\cdot)$ as a function of $P_\sigma$. For example, narrow passages passed by $\sigma$ largely determine its accessible free space. The minimum passage width in $P_\sigma$ can be formulated in $f_c(\cdot)$ to avoid passing narrow passages. Such requirements are important when a proper path clearance cannot be prespecified, e.g., in deformable object manipulation and swarm navigation in obstacle-dense environments \citep{huang2024homotopic, mao2024optimal, mao2025tube}. To sum up, PTOPP aims to find a path $\sigma^*$ such that $f_c(\sigma^*) = \min\{f_c(\sigma): \sigma \text{ is feasible}\}$ with $f_c(\cdot)$ being a function of $P_\sigma$, and reports failure if no such path exists.

\subsection{Passage definition with Gabriel condition}
Passage detection is the prerequisite for PTOPP. Unlike RBB and its variants \citep{hsu2003bridge, sun2005narrow} that identify narrow passages in the random sampling process, the obstacle distribution is assumed to be known. Intuitively, narrow passages are where free space significantly shrinks and becomes much less accessible due to the existence of nearby obstacles. To detect them, the visibility condition excludes visually invalid passages blocked by obstacles \citep{huang2023deformable}, whereas a large fraction of redundant passages remain, e.g., \hyperref[Fig: Passage Example in Lab]{Figure \ref{Fig: Passage Example in Lab}}. Redundant passages are not informative in evaluating accessible nearby free space, hence inducing uncoordinated distributions and high computational loads. Our goal is a sparse and informative distribution of narrow passages.

\begin{figure}[t]
    \minipage{1 \columnwidth}
     \centering
     \includegraphics[width= 0.94 \columnwidth]{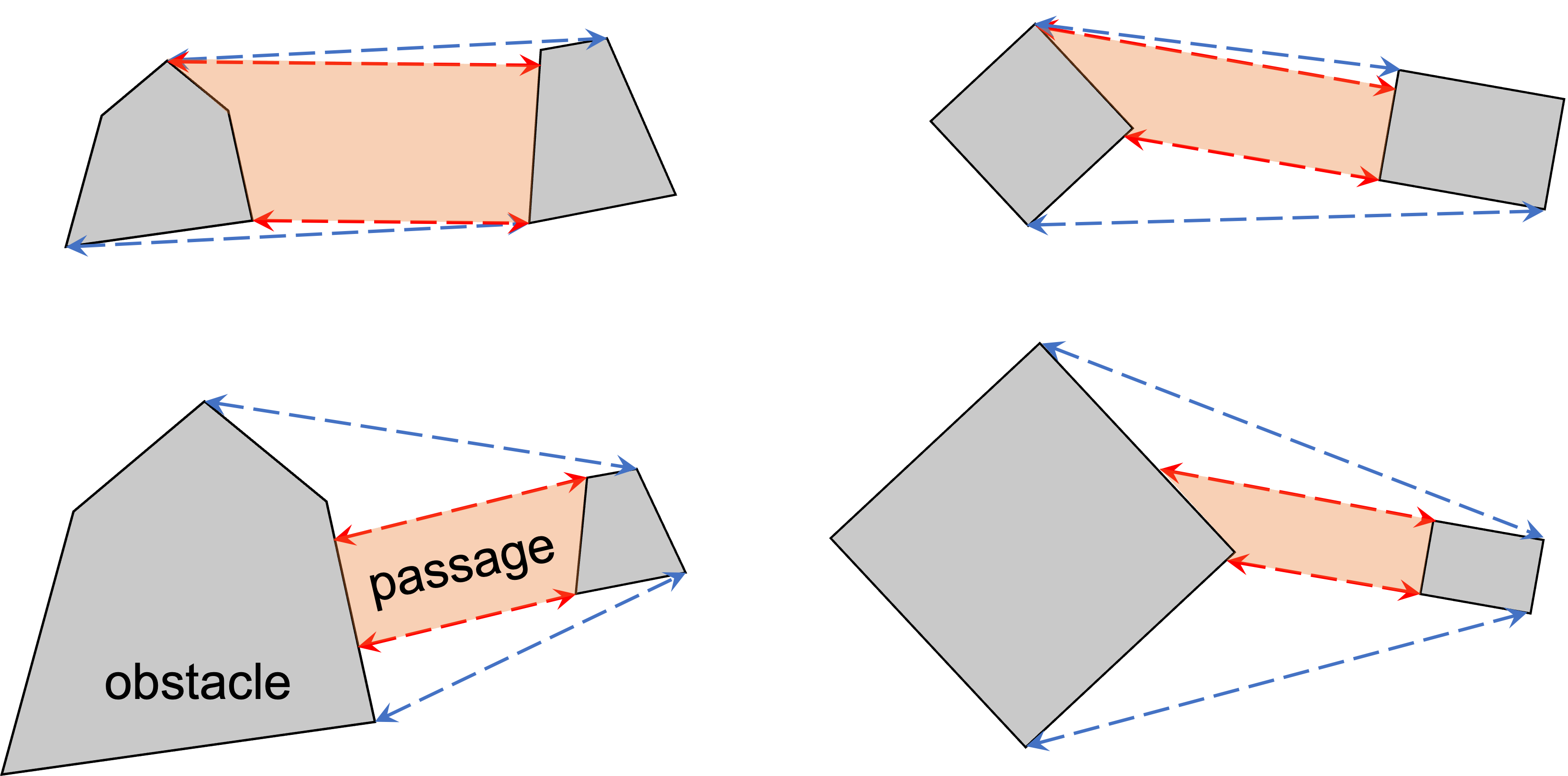}
     \endminipage \hfill
     \caption{Passage regions between obstacles. Blue dashed segments are convex hull sides of obstacles. Passages as obstacles' common projection shadow volume in the distance direction are in orange, which are not sensitive to obstacle size differences.
     }
     \label{Fig: Passage Representation}
\end{figure}

In passage representation, consider the elementary case of two neighboring polygon obstacles. The sharply shrinking free space between them should be classified as the passage region. A natural option is the convex hull enclosed by obstacles. However, the resulting passage region tends to be oversized if obstacle sizes differ dramatically, as exemplified in \hyperref[Fig: Passage Representation]{Figure \ref{Fig: Passage Representation}}. A more conservative measure leverages the most constrained place defined by the shortest distance between obstacles. The obstacle distance is 
\begin{equation}
\label{Eqn: Obstacle Distance}
    (\mathbf{p}_i^*, \mathbf{p}_j^*) = \operatorname*{arg \, min}_{ \mathbf{p}_i \in \mathcal{O}_i, \mathbf{p}_j \in \mathcal{O}_j } \| \mathbf{p}_i - \mathbf{p}_j \|.
\end{equation}
Let $l(\mathbf{p}_i^*, \mathbf{p}_j^*)$ be the segment connecting $\mathbf{p}_i^*$ and $\mathbf{p}_j^*$. The passage region $\mathcal{P}_{i-j}$ should cover $l(\mathbf{p}_i^*, \mathbf{p}_j^*)$ as the most constrained part enclosed by $\mathcal{O}_i$ and $\mathcal{O}_j$. Starting from this observation, $\mathcal{P}_{i-j}$ grows orthogonal to $l(\mathbf{p}_i^*, \mathbf{p}_j^*)$ bidirectionally until it no longer simultaneously intersects with $\mathcal{O}_i$ and $\mathcal{O}_j$. Formally, the resulting passage region is the intersection of two obstacles' projection shadow volumes in $l(\mathbf{p}_i^*, \mathbf{p}_j^*)$ direction, i.e.,
\begin{equation}
\label{Eqn: Passage Region}
    \mathcal{P}_{i-j} := \text{SV}_{\mathbf{e}_{i,j}}(\mathcal{O}_i) \cap \text{SV}_{\mathbf{e}_{j,i}}(\mathcal{O}_j)
\end{equation}
where $\mathbf{e}_{i,j}$ is the unit vector of $\mathbf{p}_j^* - \mathbf{p}_i^*$ and $\mathbf{e}_{i,j}$ is its inverse, both designating the projection directions of the shadow volume SV($\cdot$). In this definition, $\mathcal{P}_{i-j}$ is insensitive to obstacle size differences as in \hyperref[Fig: Passage Representation]{Figure \ref{Fig: Passage Representation}}. 

\begin{figure}[t]
    \minipage{1 \columnwidth}
     \centering 
     \includegraphics[width= 0.71 \columnwidth]{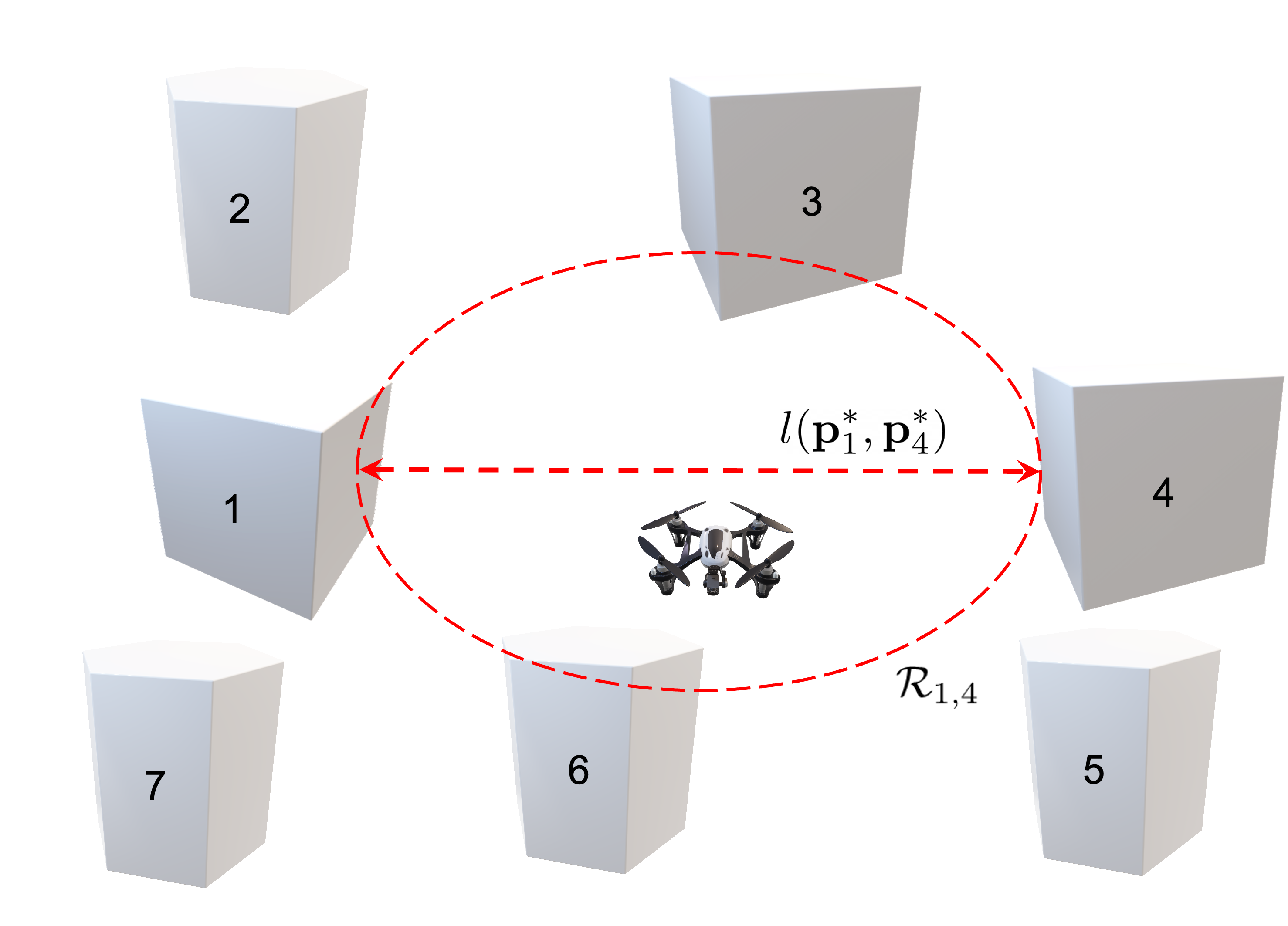} 
     \endminipage \hfill
     \caption{Gabriel condition in passage detection exemplified with mobile robots. Due to intersections between $\mathcal{R}_{1,4}$ and $\mathcal{O}_3$, $\mathcal{O}_6$, $\mathcal{P}_{1-4}$ is not identified as a valid passage.}
     \label{Fig: Gabriel Graph Condition Illustration}
\end{figure}

Not all passages given by (\ref{Eqn: Passage Region}) are informative amid multiple obstacles. The validity of $\mathcal{P}_{i-j}$ is further examined in the presence of other obstacles. To address the passage redundancy in the visibility condition, the extended visibility condition is proposed to enforce more thorough passage filtering in \citep{huang2024homotopic}. Specifically, by examining how robots are restricted inside passages, the maximum free space limited by $l(\mathbf{p}_i^*, \mathbf{p}_j^*)$ is the closed disc $\mathcal{R}_{i,j}$ with $l(\mathbf{p}_i^*, \mathbf{p}_j^*)$ as its diameter. If any other obstacle $\mathcal{O}_k$ intersects with $\mathcal{R}_{i,j}$, the free space shrinks and $\mathcal{O}_k$ forms passages with $\mathcal{O}_{i} $ or $\mathcal{O}_j$ narrower than $\mathcal{P}_{i-j}$, rendering $\mathcal{P}_{i-j}$ invalid. This condition is expressed as
\begin{equation}
\label{Eqn: Extended Visibility Check}
    \mathcal{V}(\mathcal{P}_{i-j}) = \text{False} \;\; \text{if} \;\; \mathcal{O}_k \cap \mathcal{R}_{i,j} \neq \emptyset, \exists \, \mathcal{O}_k
\end{equation}
where $\mathcal{V}(\cdot)$ is a binary variable encoding the passage validity. See \hyperref[Fig: Gabriel Graph Condition Illustration]{Figure \ref{Fig: Gabriel Graph Condition Illustration}} for an example and conceptual illustration of the detection condition.

\begin{figure}[t]
    \minipage{1 \columnwidth}
     \centering
     \includegraphics[width= 0.73 \columnwidth]{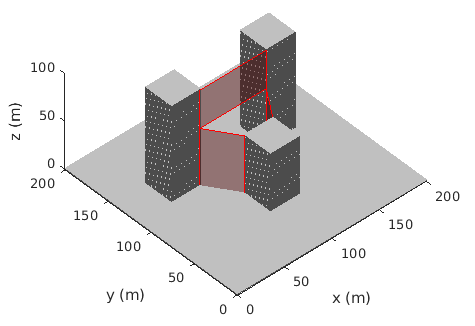} 
     \endminipage \hfill
     \caption{Illustration of spatial passages in 3D space comprising only three obstacles. The passage between two tall obstacles has a valid height range starting at the lower obstacle.}
     \label{Fig: 3D Passage Illustration}
\end{figure}

In 3D space, the most common situation where obstacles are placed on a base ground is of interest first. An interval $[\underline{h}, \overline{h}]$ is used to designate the valid height range of a spatial passage. In \citep{huang2024homotopic}, the height interval of $\mathcal{P}_{i-j}$ is set between the base (zero height) and the lower obstacle height between $\mathcal{O}_i, \mathcal{O}_j$, i.e., $\underline{h} = 0, \overline{h} = \min(h_i, h_j)$ where $h_i$ and $h_j$ are heights of $\mathcal{O}_i$ and $\mathcal{O}_j$, respectively. Simple as it is, this representation does not capture all narrow passages in space and only contains those on the base. As illustrated in \hyperref[Fig: 3D Passage Illustration]{Figure \ref{Fig: 3D Passage Illustration}}, a passage in 3D space exists until some lower obstacle blocks it. Therefore, the lower height bound $\underline{h}$ is the largest height among those obstacles making $\mathcal{V}(\cdot)$ false. After introducing the valid height interval, the detection condition becomes
\begin{equation}
\label{Eqn: Extended Visibility Check 3D}
    \mathcal{V}(\mathcal{P}_{i-j}) = \text{False} \;\; \text{if} \; \mathcal{O}_k \cap \mathcal{R}_{i,j} \neq \emptyset, h_k \geq \overline{h}, \, \exists \, \mathcal{O}_k.
\end{equation}
If there exists any obstacle $\mathcal{O}_k$ blocking $\mathcal{P}_{i-j}$ as in (\ref{Eqn: Extended Visibility Check}) and of a height larger than $\overline{h}$, $\mathcal{P}_{i-j}$ is invalid. Otherwise, $\mathcal{P}_{i-j}$ is valid and its lower height bound $\underline{h}$ equals the largest height of obstacles invalidating $\mathcal{V}(\mathcal{P}_{i-j})$,
\begin{equation}
\label{Eqn: Loewr Height Bound}
    \underline{h} = \max \{h_k : \mathcal{V}(\mathcal{P}_{i-j}, \mathcal{O}_k) = \text{False}\}
\end{equation}
where the two-argument $\mathcal{V}(\mathcal{P}_{i-j}, \mathcal{O}_k)$ encodes the validity of $\mathcal{P}_{i-j}$ tested by $\mathcal{O}_k$. As such, narrow passages at all heights in 3D space are detected with sparsity.

The condition in (\ref{Eqn: Extended Visibility Check}) expresses the proximity of $\mathcal{O}_i$ and $\mathcal{O}_j$ the same as the Gabriel graph that describes point nearness in the Euclidean space \citep{gabriel1969new, norrenbrock2016percolation}. It is thus termed the \textit{Gabriel condition} here. 
Gabriel graph is a subgraph of the Delaunay graph, the dual graph of the Voronoi diagram \citep{mark2008computational}. Although there exist other proximity graphs, e.g., the relative neighborhood graph \citep{toussaint1980relative, melchert2013percolation} and nearest neighbor graph \citep{eppstein1997nearest, prokhorenkova2020graph}, as more selective subgraphs, Gabriel graph is more suitable for passage detection because of its spatial interpretation. To the best of our knowledge, Gabriel graphs and other proximity graphs have not been applied to passage detection. Their characterization capacity of proximity relations is quite instrumental for narrow passage identification. Since only relative positions of obstacles are used, they have better generality than explicit criteria, such as preset passage width limits in RBB. 

\subsection{Passage detection with Delaunay graph}
\begin{figure}[t]
    \minipage{1 \columnwidth}
     \centering
     \includegraphics[width= 0.76 \columnwidth]{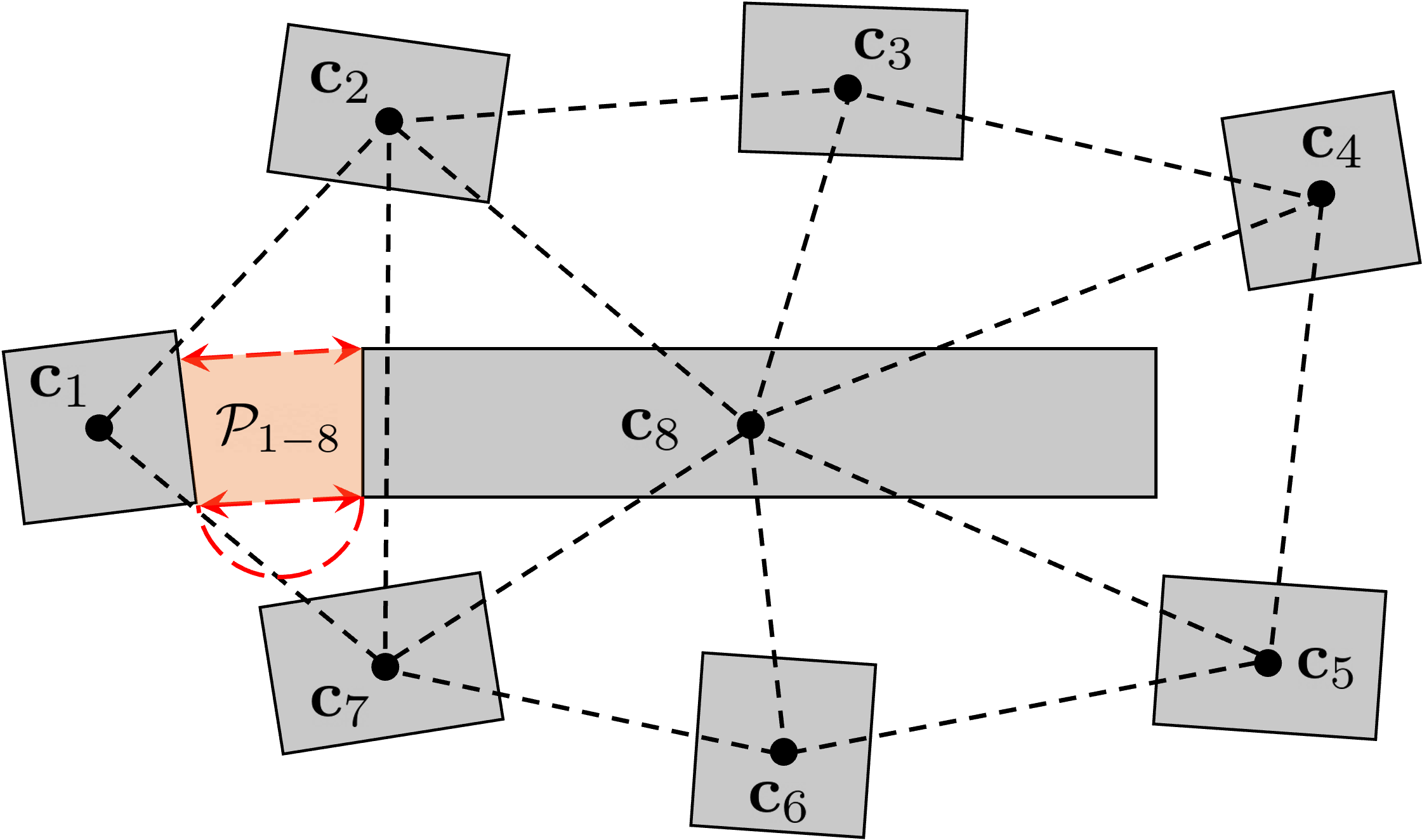}
     \endminipage \hfill
     \caption{Due to obstacle sizes, the Gabriel graph constructed from obstacle centroids may miss some passages defined by (\ref{Eqn: Amended Gabriel Condition}). While $(\mathbf{c}_1, \mathbf{c}_8)$ is not an edge of $\mathcal{DG}(C_c)$ above, $\mathcal{P}_{1-8}$ is a valid passage above.}
     \label{Fig: Obstacle Delaunay Graph Example}
\end{figure}

The Gabriel condition does not take into account how $\mathcal{R}_{i,j}$ is located in $\mathcal{P}_{i-j}$. It is possible that an obstacle $\mathcal{O}_k$ does not intersect with $\mathcal{R}_{i,j}$, but blocks $\mathcal{P}_{i-j}$ to make $\mathcal{V}(\mathcal{P}_{i-j})$ false. To remedy this, $\mathcal{R}_{i,j}$ is augmented to be the union with $\mathcal{P}_{i-j}$ and the Gabriel condition in (\ref{Eqn: Extended Visibility Check}) is amended as
\begin{equation}
\label{Eqn: Amended Gabriel Condition}
    \mathcal{V}(\mathcal{P}_{i-j}) = \text{False} \;\; \text{if} \;\; \mathcal{O}_k \cap \overline{\mathcal{R}}_{i,j} \neq \emptyset, \exists \, \mathcal{O}_k.
\end{equation}
where $\overline{\mathcal{R}}_{i,j} = \mathcal{R}_{i,j} \cup \mathcal{P}_{i-j}$. Operations on obstacles, e.g., computation of $l(\mathbf{p}_i^*, \mathbf{p}_j^*)$ and $\mathcal{P}_{i-j}$, take constant time. To find all valid passages, a direct traversal check of (\ref{Eqn: Amended Gabriel Condition}) over all obstacle pairs consumes $O(m^3)$ time. The Gabriel graph is obtainable in linear time by checking the associated Delaunay graph which takes $O(m \log m)$ time to compute for $m$ points \citep{norrenbrock2016percolation}. However, obstacles and passages are volumetric in reality. The obstacle graph connected by passages need not be isomorphic to the Gabriel graph with obstacles represented by simple nodes due to obstacle sizes (see \hyperref[Fig: Obstacle Delaunay Graph Example]{Figure \ref{Fig: Obstacle Delaunay Graph Example}}). Existing algorithms are thereby not directly applicable.
\begin{algorithm}[t]
\SetNlSty{}{}{:}
\small{
    \nl Input obstacles $\mathcal{O}_1, ..., \mathcal{O}_m$\ as polygons\;
    \nl Get obstacle centroids $C_c = \{\mathbf{c}_1, \mathbf{c}_2, ..., \mathbf{c}_m\}$\;
    \nl Compute the Delaunay graph of $C_c$ as $\mathcal{DG}(C_c)$\;
    \nl \ForEach{$\mathbf{c}_i$} {
        \nl Get the neighbor list $N_i$ with $d_{gd} \leq k_{gd}$\;
        \nl \ForEach{$\mathbf{c}_j \in N_i$} {
        \nl Get the neighbor list $N_j$ with $d_{gd} \leq k_{gd}$\;
        \nl $N_k \leftarrow N_i \cup N_j$\;
        \nl \If{\textnormal{$\mathcal{V}(\mathcal{P}_{i-j})$ == True for $\mathcal{O}_k \in N_k$}} {
            \nl Append $\mathcal{P}_{i-j}$ to the passage list\;
        }
        }
    }
    \nl \Return passage list\;
    }
    \caption{\small Passage Detection Using Delaunay Graph}
    \label{Alg: Valid Passage Detection}
\end{algorithm}
\begin{algorithm}[t]
\SetKw{logicalAnd}{and}
\SetKw{logicalOr}{or}
\SetNlSty{}{}{:}
\small{
    \nl Sort $\mathcal{O}_1, ..., \mathcal{O}_m$\ in descending order of height\;
    \nl Get the associated centroid list $C_s = \{\mathbf{c}_{s,1}, \mathbf{c}_{s,2}, ..., \mathbf{c}_{s,m}\}$\; 
    \nl Initialize the Delaunay graph $\mathcal{DG}(C_s)$ with $\mathbf{c}_{s,1}, \mathbf{c}_{s,2}$\;
    \nl \For{$i \leftarrow 3$ \KwTo $m$} {
        \nl Add $\mathbf{c}_{s,i}$ to $\mathcal{DG}(C_s)$\;
        \nl Get $\mathbf{c}_{s,i}$'s neighbor list $N_i$ with $d_{gd} \leq k_{gd}$\;
        \nl \ForEach{$\mathcal{P}_{p-q} \in$ \textnormal{passage list}} {
            \nl \If{\textnormal{($p \in N_i$ \logicalOr $q \in N_i$) 
                    \\ \logicalAnd height interval of $\mathcal{P}_{p-q}$ is not processed 
                    \\ \logicalAnd $\mathcal{V}(\mathcal{P}_{p-q}, \mathcal{O}_i)$ == \textnormal{False}}} {
                    \nl Assign $\mathcal{P}_{p-q}$'s lower height bound as $\underline{h} = h_i$\;
                    \nl Label $\mathcal{P}_{p-q}$'s height interval processed\;
            }
        }    
        \nl \ForEach{$\mathbf{c}_j \in N_i$} {
        \nl Get the neighbor list $N_j$ with $d_{gd} \leq k_{gd}$\;
        \nl $N_k \leftarrow N_i \cup N_j$\;        
        \nl \If{\textnormal{$\mathcal{V}(\mathcal{P}_{i-j}) ==$ True for $\mathcal{O}_k \in N_k$}}{ 
            \nl Append $\mathcal{P}_{i-j}$ to the passage list with its height interval being $[0, h_i]$\;
            }
        }
    }
    \nl \Return passage list\;
    }
    \caption{\small Passage Detection in 3D Space}
    \label{Alg: 3D Passage Detection}
\end{algorithm}

To reduce detection complexity, the proximity of Delaunay graph nodes is leveraged. An obstacle $\mathcal{O}_i$ is represented as its centroid $\mathbf{c}_i$. The Delaunay graph $\mathcal{DG}(C_c)$ is computed from the centroid set $C_c = \{\mathbf{c}_1, \mathbf{c}_2, ..., \mathbf{c}_m \}$. Although obstacle geometries are indeterminate, the geodesic distance $d_{gd}(\mathbf{c}_i, \mathbf{c}_j)$ between $\mathbf{c}_i$ and $\mathbf{c}_j$ in $\mathcal{DG}(C_c)$, i.e., the number of edges in the shortest path connecting them, is small and bounded if $\mathcal{P}_{i-j}$ is valid. Hence, it suffices to examine limited neighbors of $\mathcal{O}_i$. The smallest geodesic distance $k_{gd}$ to be checked is
\begin{equation}
\label{Eqn: Smallest Geodesic Distance}
    k_{gd} = \max \{d_{gd}(\mathbf{c}_i, \mathbf{c}_j) \, \text{in} \, \mathcal{DG}(C_c) \, | \, \mathcal{V}(\mathcal{P}_{i-j}) = \text{True} \}.
\end{equation}
$k_{gd}$ is affected by obstacle geometries and distributions. In point sets, $k_{gd} = 1$ to check if an edge incident to $\mathbf{c}_i$ is in the Gabriel graph. As obstacles are simplified as points, $k_{gd} = 1$ is no longer sufficient to find all $\mathcal{O}_j$ such that $\mathcal{P}_{i-j}$ is valid, e.g., \hyperref[Fig: Obstacle Delaunay Graph Example]{Figure \ref{Fig: Obstacle Delaunay Graph Example}}. In practice, for random obstacle distributions under limited size variations, $k_{gd}$ is small and assigned as two in our implementation (see \hyperref[Appendix: Geodesic Distance Test]{Appendix \ref{Appendix: Geodesic Distance Test}} for analysis). For a given pair of $\mathcal{O}_i, \mathcal{O}_j$, (\ref{Eqn: Amended Gabriel Condition}) is checked only with obstacles $\mathcal{O}_k$ of $d_{gd}$ to $\mathcal{O}_i$ or $\mathcal{O}_j$ not greater than $k_{gd}$. Let such a neighbor list of $\mathcal{O}_i$ be $N_i(k_{gd}) = \{\mathcal{O}_k | d_{gd}(\mathcal{O}_i, \mathcal{O}_k) \leq k_{gd}\}$. Then passages incident to $\mathcal{O}_i$ are detected between $\mathcal{O}_j \in N_i(k_{gd})$ and
\begin{equation}
\label{Eqn: Amended Gabriel Condition Within Geodesic Distances}
    \begin{split}
    \mathcal{V}(\mathcal{P}_{i-j}) &= \text{False} \;\; \text{if} \;\; \mathcal{O}_k \cap \overline{\mathcal{R}}_{i,j} \neq \emptyset, \\
    \exists \, \mathcal{O}_k &\in N_i(k_{gd}) \cup N_j(k_{gd}).
    \end{split}
\end{equation}
The expected degree of a Delaunay graph node is at most six \citep{mark2008computational}. Therefore, the number of edges to be examined is linear in the number of obstacles and the cardinality $|N_i(k_{gd})|$ is a constant on average. The overall $O(m \log m)$ passage detection algorithm using Delaunay graph of obstacles is outlined in Algorithm \ref*{Alg: Valid Passage Detection}.

Instead of using 3D proximity graphs, determining spatial passages with height intervals fits into the randomized incremental construction process of planar Delaunay graphs. Obstacles are first sorted in descending order of height. When computing the Delaunay graph, obstacles are inserted in the sorted order and passages are updated locally near the newly added obstacle $\mathcal{O}_i$. The update consists of two parts. First, the validity of previously detected passages that contain obstacles in $N_i(k_{gd})$ is examined. Second, new passages between $\mathcal{O}_i$ and obstacles in $N_i(k_{gd})$ are checked. This loop ends until all obstacles are added (see Algorithm \ref*{Alg: 3D Passage Detection}). As such, the core routine of passage detection is limited to the neighborhood of the newly added obstacle. Only a constant number of previously detected and new passages are processed in each iteration. Therefore, the overall detection procedure preserves the complexity of Delaunay graph construction as $O(m \log m)$. 

\subsection{Free space decomposition with Gabriel cells}
\begin{figure}[t]
    \minipage{1 \columnwidth}
     \centering
     \frame{\includegraphics[width= 0.78 \columnwidth]{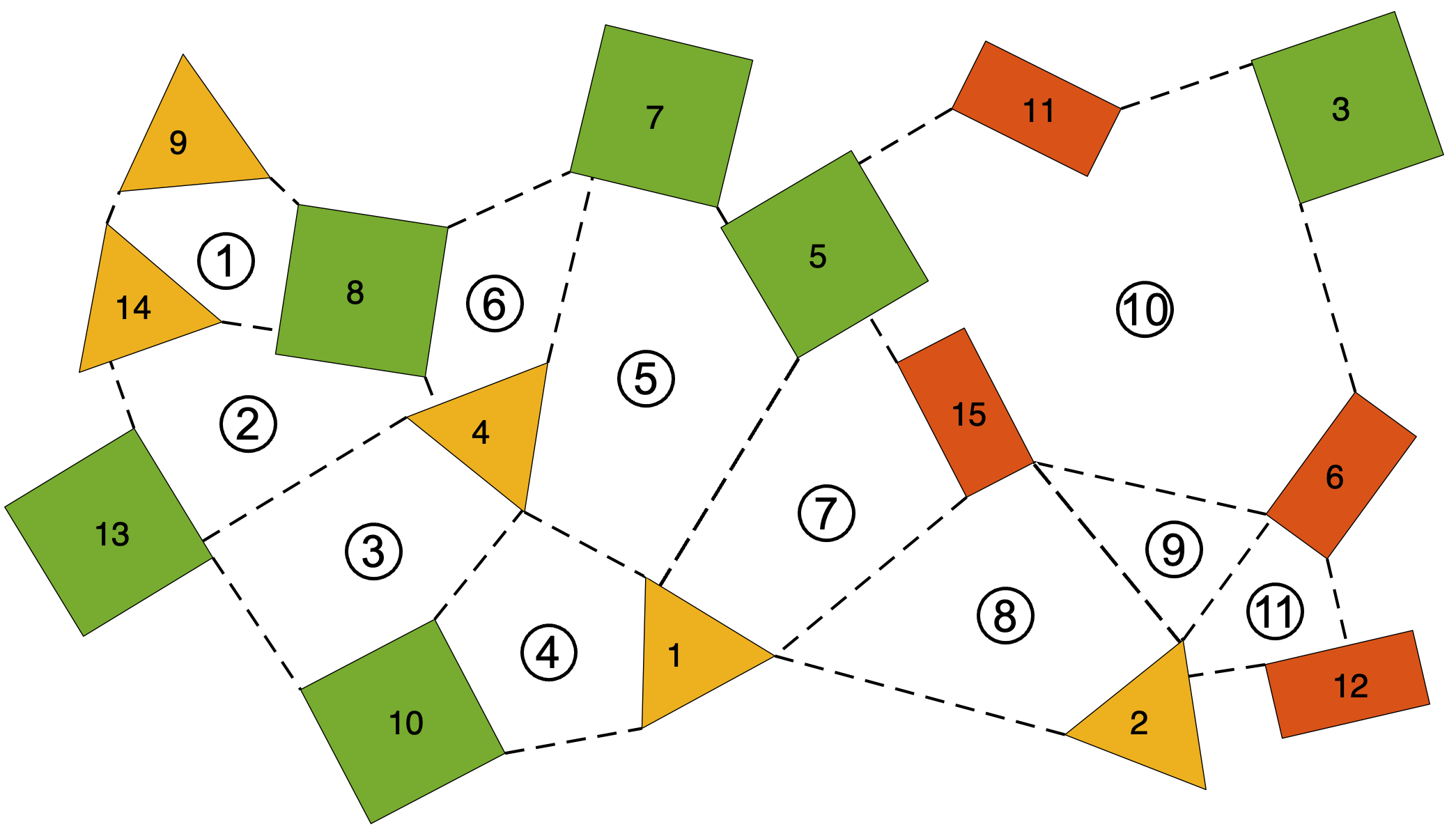}}
     \endminipage \hfill
     \caption{Free space is divided into Gabriel cells denoted by circled numbers (environment boundaries are not considered). Cells are enclosed by passage segments and obstacles. Adjacent cells have passage segments as common edges. }
     \label{Fig: Gabriel Cells Illustration}
\end{figure}

A prominent advantage of the proposed passage detection procedure is that it brings a natural free space subdivision. Delaunay graphs are planar (i.e., no intersecting edges) and connected (i.e., no isolated nodes). As a result, the passage distribution essentially defines a decomposition of the free space. The cells are bounded by passages and have obstacles as their vertices. Together with environment boundaries, a cell decomposition of the entire free space is obtained. The free space $\mathcal{X}_{free}$ is divided into separate regions $\mathcal{C}_i$ such that $\mathcal{X}_{free} = \cup_{i = 1}^{n_c} \mathcal{C}_i$ where $n_c$ designates the number of cells. For cell adjacency,  $\forall i \neq j$
\begin{equation}
\label{Eqn: Gabriel cell intersection}
    \mathcal{C}_i \cap \mathcal{C}_j = 
    \begin{cases}
        \emptyset    & \text{if} \, \mathcal{C}_i \, \text{and} \, \mathcal{C}_j \, \text{are not adjacent} \\
        l(\mathbf{p}_p^*, \mathbf{p}_q^*)    & \text{otherwise}
    \end{cases}
\end{equation}
where $l(\mathbf{p}_p^*, \mathbf{p}_q^*)$ is the neighboring passage segment of $\mathcal{C}_i$ and $\mathcal{C}_j$. Namely, obstacles and passage segments comprise cell edges, but cells only have passage segments as common edges (see \hyperref[Fig: Gabriel Cells Illustration]{Figure \ref{Fig: Gabriel Cells Illustration}}). These cells are termed Gabriel cells here, which enable many cell-based path planning methods \citep{choset2005principles, lavalle2006planning}. When integrated into sampling-based PTOPP, they can dramatically speed up passage traversal check of paths via sample localization in cells, making PTOPP computationally efficient without extra decomposition steps.

Gabriel cells always contain passage segments as sides. They additionally have sides from obstacle boundaries in most situations for non-negligible obstacle sizes, e.g., \hyperref[Fig: Gabriel Cells Illustration]{Figure \ref{Fig: Gabriel Cells Illustration}}. Detected passages define a passage graph. To report all cells, face detection is carried out in this graph. Adjacency lists in the passage graph are first sorted counter-clockwise (or clockwise). Let $A_i$ be the passage adjacency list of $\mathcal{O}_i$, i.e., $A_i = \{j \, | \, \mathcal{V}(\mathcal{P}_{i-j}) = \text{True} \}$. $A_i$ is sorted such that 
\begin{equation}
\label{Eqn: Clockwise Comparer}
    j \prec k \;\; \text{if } \texttt{Oriented}(\mathbf{c}_{i,j}, \mathbf{c}_{i,k}) = \text{True} \;\; \forall j, k \in A_i
\end{equation}
and $\mathbf{c}_{i,j} = \mathbf{c}_j - \mathbf{c}_i$. \texttt{Oriented}() is a transitivity-preserving comparer telling if it is counter-clockwise from  $\mathbf{c}_{i,j}$ to $\mathbf{c}_{i,k}$. Later the passage graph is traversed in a depth-first search (DFS) way in which edges maintain a clockwise order until a face is enclosed. Cell vertices are associated passage segment endpoints and are readily retrievable. See Algorithm \ref*{Alg: Gabriel Cell Detection} for the Gabriel cell detection procedure. Sorting $O(m)$ items takes $O(m \log m)$ time, and graph traversal takes $O(m)$ time, resulting in overall $O(m \log m)$ complexity.

Although Gabriel cell decomposition is transplantable to 3D space to decompose the space into voxels in obstacle height intervals, the resulting cells have complex boundaries and adjacency relations. To use cells for PTOPP in 3D space, cells on the base are employed. Specifically, planar cells on the base are found considering passages rooted on the base ($\underline{h} = 0$). For midair passages with $\underline{h} > 0$, their projections on the base intersect with base cells. Since the projection may fall into the obstacle region, obstacle region is treated as planar cells (obstacle cross sections are assumed invariant), resulting in compound Gabriel cell decomposition in Algorithm \ref*{Alg: Compound Cell Detection 3D}. Finding intersected cells takes $O(1)$ time by calling the passage traversal check routine. Midair passage number does not exceed $O(m)$. Therefore, compound cell detection also takes $O(m)$ time. Midair passages are treated as virtual sides of base cells which their projections intersect with. When checking passed passages of paths in 3D space, virtual sides of cells are additionally checked. 
\begin{algorithm}[t]
\SetNlSty{}{}{:}
\small{
    \nl Get passage adjacency lists $A_1,..., A_m$ from passages\;
    \nl Sort each $A_i$ using comparer (\ref{Eqn: Clockwise Comparer})\;
    \nl Assign all-false arrays $U_1, ..., U_m$ with cardinality $| U_i | = | A_i |$ to record edge visit status\;
    \nl $C \leftarrow \emptyset$\;
    \nl \For{$i \leftarrow 1$ \KwTo $m$} {
        \nl \For{$j \leftarrow 1$ \KwTo $| A_i |$} {
            \nl \If {$U_{i,j}$ == True} {
            \nl continue\;
            }
            \nl $\mathcal{C}_t \leftarrow \emptyset$; $u \leftarrow i; e \leftarrow j $\;
            \nl \While{$U_{u,e}$ == False} {
                \nl $U_{u,e} \leftarrow \text{True}; \mathcal{C}_t \leftarrow \mathcal{C}_t \cup \{u\}; v \leftarrow A_{u,e}$\;
                \nl $e' \leftarrow $ The next element in $A_v$ after $u$ by (\ref{Eqn: Clockwise Comparer})\;
                \nl $u \leftarrow v; e \leftarrow e'$\;
            }
        \nl $C \leftarrow C \cup \{\mathcal{C}_t\}$\;    
        }
    }
    \nl \Return cell list $C$\;
    }
    \caption{\small Gabriel Cell Detection}
    \label{Alg: Gabriel Cell Detection}
\end{algorithm}
\begin{algorithm}[t]
\SetNlSty{}{}{:}
\small{
    \nl Detect Gabriel cells using passages starting at base\; 
    \nl Add obstacle base sections as cells\;
    \nl \ForEach{$\mathcal{P}_{i-j}$ \textnormal{with} $\underline{h} > 0$} {
        \nl Get cells the projection of $\mathcal{P}_{i-j}$ intersects with\;
        \nl Add $\mathcal{P}_{i-j}$ as virtual sides to these cells\;
        }
    \nl \Return cell list\;
    }
    \caption{\small Compound Cell Detection in 3D Space}
    \label{Alg: Compound Cell Detection 3D}
\end{algorithm}

\section{PTOPP categories}
\label{Sec: Problem Categories}
Before developing the final planning algorithms, this section introduces typical PTOPP problems characterized by path cost formulations, which are proposed to be compatible with sampling-based optimal planners.

\subsection{Minimum passage width PTOPP}
The primary objective of PTOPP in this article is to find paths with large accessible free space while being suboptimal in other crucial criteria, of which the most prominent is the path length. Such planning goals are embedded in the cost function $f_c(\cdot)$ in OPP. In PTOPP, $f_c(\cdot)$ is structured as a function of the path passage traversal list $P_{\sigma}$, and various constructions exist. Since the narrowest passage gauges the most constrained region the path should avoid, the minimum passage width passed by $\sigma$ is an important metric in $f_c(\cdot)$. The width of passage $\mathcal{P}_{i-j}$, denoted by $\| \mathcal{P}_{i-j} \|$, is defined as the distance between $\mathcal{O}_i$ and $\mathcal{O}_j$, i.e., $\| \mathcal{P}_{i-j} \| := \| l(\mathbf{p}_i^*, \mathbf{p}_j^*) \|$. The minimum passage width traversed by $\sigma$ is $f_p(\sigma) = \min \| \mathcal{P}_{i-j} \|, \, \mathcal{P}_{i-j} \in P_{\sigma}(1)$. Analogously, $f_p(\sigma)$ as a function of the path argument $\tau$ is 
\begin{equation}
\label{Eqn: Minimum Passage Width on Path}
    f_p(\sigma, \tau) = \min \| \mathcal{P}_{i-j} \| \;\; \mathcal{P}_{i-j} \in P_{\sigma}(\tau)
\end{equation}
which encodes the narrowest passage width passed from $\sigma(0)$ to $\sigma(\tau)$. Problems solely formulating the minimum passage width in the path cost are classified as \textit{minimum passage width PTOPP} (MPW-PTOPP). MPW-PTOPP aims to make $f_p$ large enough.

Different planning objectives are often conflicting. For instance, a path not traversing narrow paths need not be short and vice versa. Improving one objective usually comes with deteriorating the others \citep{sakcak2021complete}. To handle this, a trade-off is often adopted. One versatile cost used in \citep{huang2024homotopic} is
\begin{equation}
\label{Eqn: Weighted MPW And Length Cost}
    f_c(\sigma) = \text{Len}(\sigma) - w_p f_p(\sigma)
\end{equation}
where $w_p > 0$ is the weight adjusting the dominance between Len$(\sigma)$ and $f_p(\sigma)$, offering (\ref{Eqn: Weighted MPW And Length Cost}) good adjustability over other trade-off costs, e.g., the ratio cost $f_c(\sigma) = \text{Len}(\sigma) / f_p(\sigma)$ in \citep{huang2023deformable}. Path costs need to have certain properties. Specifically, they must be monotonically non-decreasing and bounded in path concatenation to be optimizable in sampling-based optimal planners \citep{karaman2011sampling}. To achieve monotonicity of MPW-PTOPP costs, $f_p(\sigma)$ is initialized as a large value when $\sigma$ has not passed passages yet, i.e.,
\begin{equation}
\label{Eqn: Passage Width Upper Bound}
    f_p(\sigma) = \bar{\varepsilon}_p \;\; \text{if} \;\; P_{\sigma}(1) = \emptyset
\end{equation}
where $\bar{\varepsilon}_p$ is a value no smaller than the maximum passage width in the environment. As such, $f_c(\sigma)$ is non-decreasing after traversing any passage.

Clearly, $f_c(\cdot)$ above satisfies monotonicity and boundedness (Len$(\cdot)$ is monotonically increasing. $f_p(\cdot)$ is monotonically non-increasing). However, it proves to be insufficient. The cost is still not asymptotically optimizable by sampling-based optimal planners for not being order-preserving. Formally, suppose path $\sigma_1$ and $\sigma_2$ reach the same endpoint, i.e., $\sigma_1(1) = \sigma_2(1)$. Both are concatenated with path $\sigma_3$ at $\sigma_3(0) = \sigma_1(1) = \sigma_2(1)$. A path cost $f_c(\cdot)$ is said to be order-preserving if the following holds 
\begin{equation}
\label{Eqn: Order-Preserving Definition}
    f_c(\sigma_1) \geq f_c(\sigma_2) \Rightarrow f_c(\sigma_1 | \sigma_3) \geq f_c(\sigma_2 | \sigma_3)
\end{equation}
\citep{pearl1984heuristics}, which can be regarded as a generalized and less demanding condition of the broadly used additivity property of path costs. Sampling-based optimal planners implicitly rely on the order-preserving property of costs to find the optimal solution. A basic hypothesis is that reducing the cost of an intermediate path segment decreases the entire path's cost. This is utilized in optimal path search in roadmaps, e.g., optimal PRM (PRM$^*$), and locally rewiring trees, e.g., RRT$^*$. Suboptimal paths are returned for non-order-preserving path costs. To remedy this, a path cost in PTOPP is said to be \textit{compatible} if it is monotonically non-decreasing, bounded, and order-preserving under path concatenation.

\begin{figure}[t]
    \minipage{1 \columnwidth}
     \centering
     \includegraphics[width= 0.75 \columnwidth]{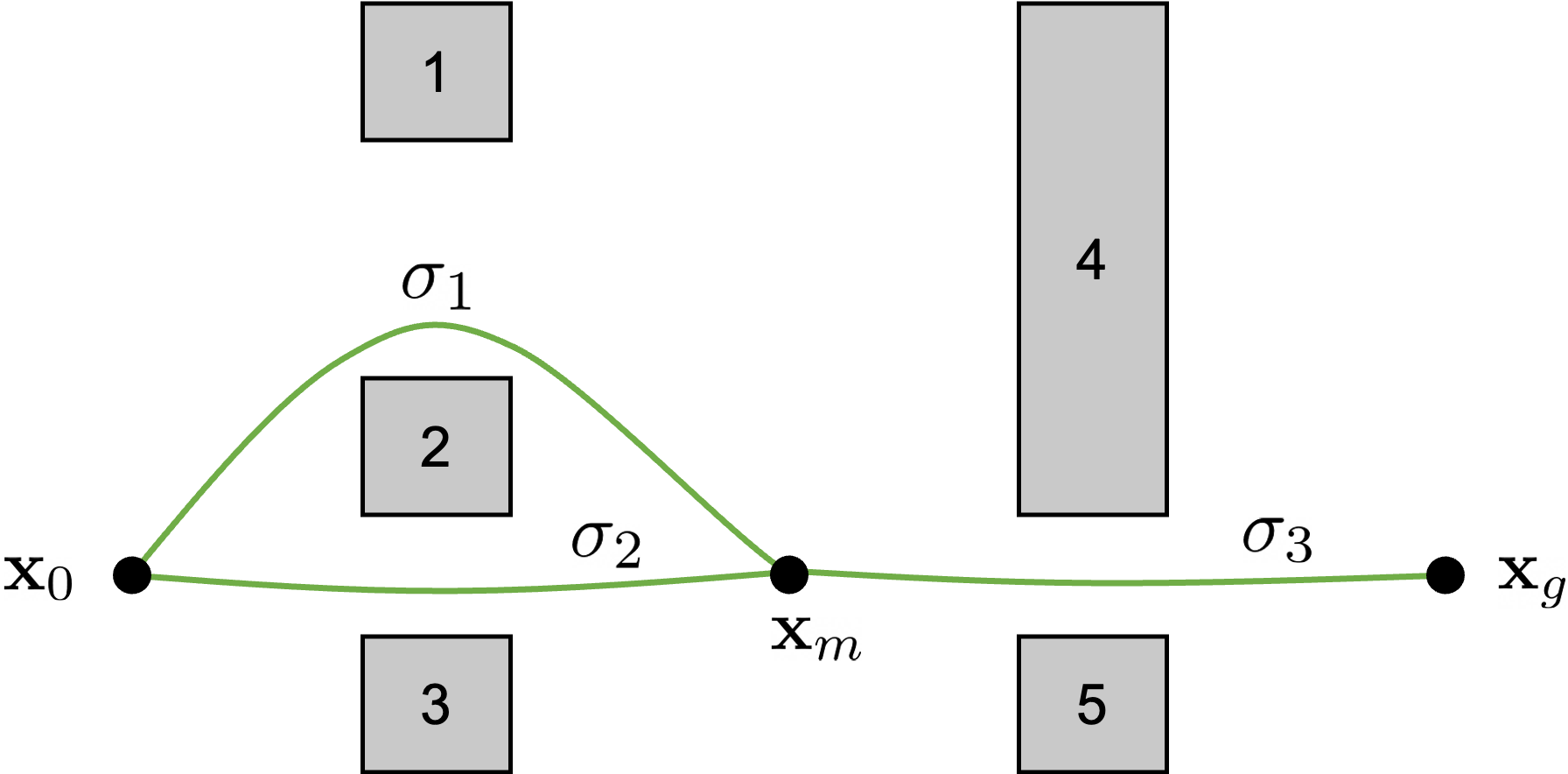}
     \endminipage \hfill
     \caption{Example of lack of optimizability for non-compatible costs by sampling-based optimal planners. Consider cost (\ref{Eqn: Weighted MPW And Length Cost}) with a large weight $w_p$ so $f_p(\sigma)$ dominates the cost. Passages $\| \mathcal{P}_{1-2} \| > \| \mathcal{P}_{2-3} \| = \| \mathcal{P}_{4-5} \|$. $\mathbf{x}_m$ is an intermediate path node. The globally optimal path is the joint path $\sigma* = \sigma_2 | \sigma_3$, but $\sigma_1 | \sigma_3$ is found since the optimal path from $\mathbf{x}_0$ to $\mathbf{x}_m$ is found as $\sigma_1$.}
     \label{Fig: Noncompatible Cost Example}
\end{figure}

The compound cost in (\ref{Eqn: Weighted MPW And Length Cost}) is not order-preserving due to the interaction between Len$(\sigma)$ and $w_p f_p(\sigma)$. This can be noticed by investigating the occasion where $f_p(\sigma)$ is achieved at $\sigma(\tau_p)$. To be globally optimal, the optimal path segment from $\mathbf{x}_0$ to $\sigma(\tau_p)$ is the shortest path to attain the minimum Len$(\sigma)$, which, nevertheless, generally cannot be found via optimizing (\ref{Eqn: Weighted MPW And Length Cost}). See \hyperref[Fig: Noncompatible Cost Example]{Figure \ref{Fig: Noncompatible Cost Example}} for an intuitive example. An order-preserving cost for MPW-PTOPP is taking negative $f_p(\sigma)$, i.e.,
\begin{equation}
\label{Eqn: Negative MPW-PTOPP Cost}
    f_c(\sigma) = -\min \| \mathcal{P}_{i-j} \| \;\; \mathcal{P}_{i-j} \in P_{\sigma}(1).
\end{equation}
While $f_c(\cdot)$ above is piece-wise constant since it gets updated only if $\sigma$ has passed a passage narrower than before, it is easy to verify that the cost is compatible. Paths traversing the same minimum-width passage have an equal cost, leading to a non-zero measure of optimal paths. Let $\Sigma^*$ be the set of optimal paths and $\mathcal{X}^*$ be the set of states that an optimal path in $\Sigma^*$ passes through
\begin{equation}
\label{Eqn: Optimal State Space}
    \mathcal{X}^* = \{ \mathbf{x} \in \mathcal{X}_{free} \, | \, \mathbf{x} = \sigma^*(\tau), \, \exists \, \sigma^* \in \Sigma^*, \tau \in [0, 1]\}
\end{equation}
and $\mu(\mathcal{X}^*)$ denotes the hypervolume of $\mathcal{X}^*$. A widely used assumption in OPP is zero-measure of optimal paths, i.e., $\mu(\mathcal{X}^*) = 0$, to imply that an optimal path cannot be found in a finite number of iterations. For discretely updated cost (\ref{Eqn: Negative MPW-PTOPP Cost}) in PTOPP, $\mu(\mathcal{X}^*) > 0$ as $\mu(\mathcal{X}^*) \geq \text{Len}(\sigma^*) \mu(\mathcal{P}^*) > 0$ where $\mu(\mathcal{P}^*) > 0$ is the hypervolume of the minimum-width passage in dim$(\mathcal{X})$ - 1 dimensional space. 

A positive $\mu(\mathcal{X}^*)$ infers a positive probability of finding an optimal path even after a finite number of sampling iterations. But returned optimal paths differ significantly in terms of length and topology properties from run to run by solely maximizing $f_p(\sigma)$. To remove this indeterminacy and enhance optimality, extra planning objectives are introduced while the optimal path cost is imposed as a constraint. In particular, for (\ref{Eqn: Negative MPW-PTOPP Cost}), the optimal path is found in a length-regulated form as 
\begin{equation}
\label{Eqn: Constrained MPW-PTOPP Formulation}
\begin{split}
    \sigma^* &= \operatorname*{arg \, min}_{\sigma \in \mathcal{H}} \, \text{Len}(\sigma) \\
    \st  \; &f_c(\sigma^*) = \min f_c(\sigma) \\
\end{split}
\end{equation}
where $\mathcal{H}$ is the solution path set for $\min f_c(\sigma)$ returned by planners.  $\mathcal{X}^*$ now has zero measure because of minimization of Len$(\sigma^*)$. Though $\mu(\mathcal{X}^*) = 0$ does not imply a single optimal path, the stochasticity of the optimal path is removed. $f_c(\cdot)$ has a higher priority over Len$(\cdot)$ and hence the global optimality. The optimal path is reachable since both $f_c(\cdot)$ and Len$(\cdot)$ are compatible. The cost compatibility is thereafter ensured. In fact, MPW-PTOPP finds paths homotopic to those with the maximum clearance because the clearance of a path is half $f_p(\sigma)$ in planner cases.

\subsection{Global passage width PTOPP} 
\begin{figure}[t]
    \minipage{1 \columnwidth}
     \centering
     \includegraphics[width= 0.86 \columnwidth]{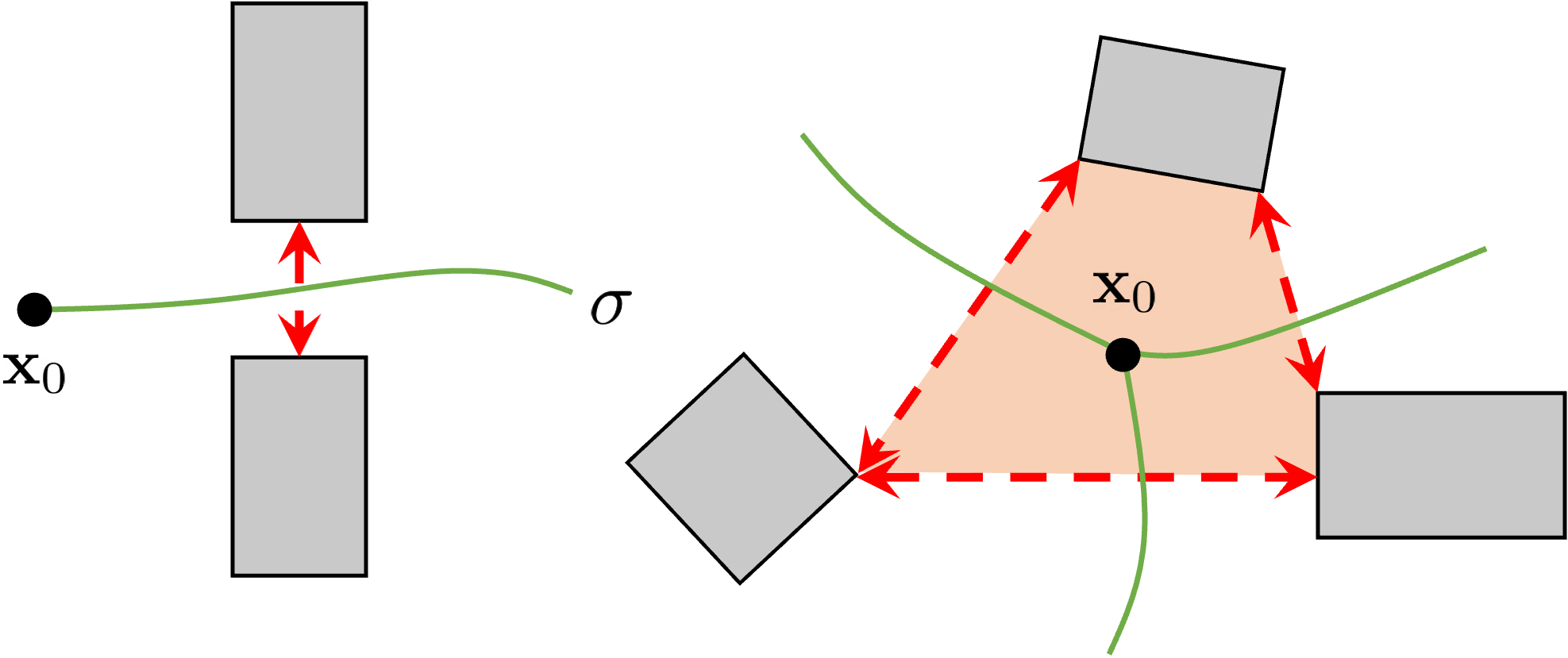}
     \endminipage \hfill
     \caption{On the left, the start and target are on two sides of the narrowest passage. Feasible paths must traverse this passage. The start position on the right is enclosed by passages. Any feasible path will pass one of them.}
     \label{Fig: Invertible Passage Example}
\end{figure}

Despite its simplicity and versatility, only taking into account $f_p(\sigma)$ may not reach high-quality paths for ignoring passages other than $f_p(\sigma)$. For example, $f_p(\sigma)$ is an inevitable passage from the start $\mathbf{x}_0$ in \hyperref[Fig: Invertible Passage Example]{Figure \ref{Fig: Invertible Passage Example}}. MPW-PTOPP essentially degrades to shortest path planning. This further scales to occasions where feasible paths have inevitable passages. A typical situation is enclosure of $\mathbf{x}_0$ by passages. Paths must traverse one surrounding passage (see \hyperref[Fig: Invertible Passage Example]{Figure \ref{Fig: Invertible Passage Example}}). Outer passages wider than them are muted in costs, highlighting the limitation of MPW-PTOPP. To alleviate this degradation, a comprehensive coverage of $P_{\sigma}$ is required. Conceptually, if traversed passage widths more than $f_p(\sigma)$ in $P_{\sigma}(1)$ are incorporated in path costs, problems are called \textit{global passage width PTOPP} (GPW-PTOPP).

Cost formalisms in GPW-PTOPP are non-trivial to involve multiple passages properly. Some common measures are not directly applicable as costs. For instance, maximizing the sum of passed passage widths incurs an ergodic passage traversal, and the average passage width lacks monotonicity. For better comparability of passage widths, denote $\mathbf{p}_\sigma(\tau)$ the passed passage widths in ascending order and $\bar{P}_{\sigma}(\tau)$ the sorted permutation of $P_{\sigma}(\tau)$ such that $\| \bar{P}_\sigma(\tau, i) \| \leq \| \bar{P}_\sigma(\tau, j) \|$ for $i < j$, then
\begin{equation}
\label{Eqn: Sorted Passage Width Vector}
    \mathbf{p}_{\sigma}(\tau) = [\| \bar{P}_\sigma(\tau, 1) \|, \| \bar{P}_\sigma(\tau, 2) \|, ...]\T
\end{equation}
where $\mathbf{p}_\sigma(\tau, i) = \| \bar{P}_\sigma(\tau, i) \|$. To involve all these traversed passage widths in costs, the path cost summing them up is framed as 
\begin{equation}
\label{Eqn: Weighted GPW-PTOPP Cost}
    f_c(\sigma) = - \mathbf{w}_p\T \mathbf{p}_\sigma(1)
\end{equation}
where $\mathbf{w}_p \succeq \mathbf{0}$ with nonnegative entries is the weight vector. How $f_c(\sigma)$ updates is the key to compatibility, but it is not well-defined since the dimension of $\mathbf{p}_\sigma$ generally increases in path concatenation. An important step is to adopt a fixed and large dimension of $\mathbf{p}_\sigma$. Say the dimension is $\textit{q}$, $\mathbf{p}_{\sigma}$ is initialized as $\bar{\varepsilon}_p \mathbf{1}_{q}$ where $\bar{\varepsilon}_p$ is a large value as in (\ref{Eqn: Passage Width Upper Bound}). $\mathbf{1}_{q}$ is the \textit{q}-dimensional all-one vector. Every sorted passage width is now assigned a weight. Elements in $\mathbf{w}_p$ are in descending order so that narrower passages have higher priorities to be avoided, i.e., $\mathbf{w}_p = [w_{p,1}, w_{p,2},..., w_{p, q}]\T$, $w_{p, i} < w_{p, i + 1}$. 

While $\mathbf{p}_\sigma$ now preserves its dimensionality and (\ref{Eqn: Weighted GPW-PTOPP Cost}) is monotonically non-increasing, it is easy to verify that $f_c(\cdot)$ is not order-preserving as a sum of paired $w_{p,i} \mathbf{p}_{\sigma}(1, i)$. To resolve this, the lexicographic order is utilized for comparing $\mathbf{p}_\sigma$ and $f_c(\cdot)$ is specially designed to preserve this order via configuring $\mathbf{w}_p$. Formally, if $\mathbf{p}_{\sigma_1}$ is lexicographically smaller than $\mathbf{p}_{\sigma_2}$, then
\begin{equation}
\label{Eqn: Lexical Order}  
    \mathbf{p}_{\sigma_1} \prec_{\text{lex}} \mathbf{p}_{\sigma_2} \Leftrightarrow \mathbf{p}_{\sigma_1, j} < \mathbf{p}_{\sigma_2, j}, \mathbf{p}_{\sigma_1, i} = \mathbf{p}_{\sigma_2, i}, \forall \, 1 \leq i < j.
\end{equation}
When $\mathbf{p}_{\sigma_1} \prec_{\text{lex}} \mathbf{p}_{\sigma_2}$, the narrowest passage $\sigma_2$ passes is wider than $\sigma_1$ except those common passages for $\sigma_1, \sigma_2$. A smaller cost is thus expected from $f_c(\sigma_2)$. This can be done by setting large differences between successive $\mathbf{w}_p$ items so that $w_{p, i} / w_{p, i+1} \gg 1$. For a given passage distribution, by setting sufficiently large differences between $w_{p, i}$, the following holds 
\begin{equation}
\label{Eqn: Lexical Order to Cost} 
    \mathbf{p}_{\sigma_1} \prec_{\text{lex}} \mathbf{p}_{\sigma_2} \Rightarrow f_c(\sigma_1) > f_c(\sigma_2)
\end{equation}
for (\ref{Eqn: Weighted GPW-PTOPP Cost}) (see \hyperref[Appendix: Path Cost Compatibility]{Appendix \ref{Appendix: Path Cost Compatibility}}). The lexical order of $\mathbf{p}_{\sigma_1}$ and $\mathbf{p}_{\sigma_2}$ is preserved under path concatenation because inserting the same value does not change their lexical order. Therefore, (\ref{Eqn: Lexical Order to Cost}) ensures that the cost function is compatible.

It is convenient to set $w_{p, i}$ as a geometric sequence with a small positive common ratio. As before, optimal paths still have a positive measure. Cast to the length-regularized form, (\ref{Eqn: Weighted GPW-PTOPP Cost}) becomes 
\begin{equation}
\label{Eqn: Constrained GPW-PTOPP Formulation}
\begin{split}
    \sigma^* &= \operatorname*{arg \, min}_{\sigma \in \mathcal{H}} \, \text{Len}(\sigma) \\
    \st  \; &f_c(\sigma^*) = \max \mathbf{w}_p\T \mathbf{p}_\sigma(1).
\end{split}
\end{equation}
In practice, \textit{q} can be fairly small to consider top \textit{k} narrowest passages and such problems are called \textit{k}-minimum passage width PTOPP (\textit{k}-MPW-PTOPP), as a generation of (\ref{Eqn: Negative MPW-PTOPP Cost}).
Directly composing a compatible trade-off cost analogous to (\ref{Eqn: Weighted MPW And Length Cost}) is attainable as $f_c(\sigma) = \text{Len}(\sigma) - \mathbf{w}\T \mathbf{p}_\sigma(1)$.
However, weights in $\mathbf{w}$ are set equal to be order-preserving, rendering it unable to prioritize avoiding narrower passages (see \hyperref[Appendix: Path Cost Compatibility]{Appendix \ref{Appendix: Path Cost Compatibility}} for more analysis of compatible GPW-PTOPP path cost design).

\subsection{Constrained passage width PTOPP} 
The last category of PTOPP focuses on problems with constraints on passed passage widths, termed \textit{constrained passage width PTOPP} (CPW-PTOPP). The most representative case is that there exists a minimum admissible passage width $\underline{\varepsilon}_p$ to accommodate robots, similar to the preset clearance requirement. Technically, this is solvable by labeling passage regions of widths smaller than $\varepsilon_p$ as obstacles to prohibit traversing. In PTOPP, such constraints are readily imposed on path costs without the labeling processing above. Firstly, CPW-PTOPP can be easily integrated into shortest path planning. If the path passes a passage narrower than $\underline{\varepsilon}_p$, a penalty is imposed by adding a large value $K_p$ to the path cost, which becomes
\begin{equation}
\label{Eqn: Passage Width Cost on Path Length}
    f_c(\sigma) = \text{Len}(\sigma) + K_p |\mathbf{p}_\sigma|_{\underline{\varepsilon}_p}
\end{equation}
where $|\mathbf{p}_\sigma|_{\underline{\varepsilon}_p}$ is the number of widths in $\mathbf{p}_{\sigma}$ smaller than $\underline{\varepsilon}_p$. The cost compatibility is easy to verify.

Next, constrained passage width could be enforced in PTOPP categories by truncating passage widths smaller than $\underline{\varepsilon}_p$ to be a negative value (e.g., $-1 / \| l(\mathbf{p}_i^*, \mathbf{p}_j^*) \| $ here) so that widths in $\mathbf{p}_\sigma$ are updated as follows
\begin{equation}
\label{Eqn: Passage Width Update Rule}
    \| \mathcal{P}_{i-j} \| = 
    \begin{cases}
        \| l(\mathbf{p}_i^*, \mathbf{p}_j^*) \|    & \| l(\mathbf{p}_i^*, \mathbf{p}_j^*) \| > \underline{\varepsilon}_p  \\
        -1 / \| l(\mathbf{p}_i^*, \mathbf{p}_j^*) \|   & \text{otherwise}.
    \end{cases}
\end{equation}
This causes the path cost $f_c(\sigma)$ to soar after passing passages narrower than $\underline{\varepsilon}_p$, excluding them from the optimal path's route. $f_c(\sigma)$ refers to any cost above. Hence, CPW-PTOPP is adapted from previous PTOPP categories by simply applying (\ref{Eqn: Passage Width Update Rule}), making passage width constraints quite convenient to enforce. Note that both MPW-PTOPP and GPW-PTOPP maximize the smallest passage width $f_p(\sigma)$ under the proposed costs. If $f_p(\sigma) > \underline{\varepsilon}_p$, (\ref{Eqn: Passage Width Update Rule}) does not change the optimal path. If $f_p(\sigma) \leq \underline{\varepsilon}_p$, (\ref{Eqn: Passage Width Update Rule}) helps select wider passages and decrease the number of passages narrower than $\underline{\varepsilon}_p$ for a smaller cost.

\section{Primitive procedures in PTOPP}
\label{Sec: PTOPP Primitives}
PTOPP shares common subroutines with sampling-based optimal planners such as the nearest neighbor search and collision check. This part details other primitive procedures PTOPP relies on, in particular, node attributes and primitive routines for passage traversal status computation.

\subsection{Path node attributes} 
Sampling-based planners maintain a graph or tree to organize samples. A sample $\mathbf{x}$ is referred to as a path/graph node whose position is drawn from a random sampling process. In OPP, basic node attributes include the predecessor (\texttt{parent}) and descendants/neighbors (\texttt{children}), the optimal path cost (\texttt{cost}) from the start node $\mathbf{x}_0$ to $\mathbf{x}$, and the associated path length (\texttt{len}). Attributes of other data structures, e.g., left/right child in kd-tree, are also assigned. In PTOPP, nodes additionally carry attributes of passage traversal and cell positioning status. \texttt{passedWidths} is the array of passage widths in the order traversed by the edge $(\mathbf{x}_{parent}, \mathbf{x})$. The number of passages an edge passes through is generally small but may exceed one (see \hyperref[Fig: Edge and Cell Position]{Figure \ref{Fig: Edge and Cell Position}}). \texttt{sortedWidths} corresponds to $\mathbf{p}_\sigma$ with $\sigma$ being the optimal path from $\mathbf{x}_0$ to $\mathbf{x}$ and is dynamically maintained as a linked list. \texttt{cellIndex} stores the Gabriel cell index where the node is located. In initialization, \texttt{sortedWidths} entries are assigned large values. Data Structure outlines these core path node attributes in {\CC} style.

\subsection{PTOPP primitives} 
\subsubsection{Passage traversal check} Given a passage $\mathcal{P}_{i-j}$ and an edge $(\mathbf{x}_1, \mathbf{x}_2)$, \texttt{Traverse}$(\mathcal{P}_{i-j}, \mathbf{x}_1, \mathbf{x}_2)$ returns True if $(\mathbf{x}_1, \mathbf{x}_2)$ passes $\mathcal{P}_{i-j}$ and False otherwise. While $\mathcal{P}_{i-j}$ represents a region, this examination is carried out between $l(\mathbf{p}_i^*, \mathbf{p}_j^*)$ and $(\mathbf{x}_1, \mathbf{x}_2)$ so that binary results exist. In 3D space, $l(\mathbf{p}_i^*, \mathbf{p}_j^*)$ becomes a plane as a gate. Under the general position assumption of nodes, corner cases of colinearity or coplanarity between passage segments and edges are not considered. Passage traversal check is essentially equivalent to an intersection check as follows
\begin{equation}
\label{Eqn: Segment Intersection Check}
\begin{split}
    \texttt{Traverse}(\mathcal{P}_{i-j}&, \mathbf{x}_1, \mathbf{x}_2) = \text{True} \Leftrightarrow \\ 
    l(\mathbf{p}_i^*, \mathbf{p}_j^*) &\cap (\mathbf{x}_1, \mathbf{x}_2) \neq \emptyset
\end{split}
\end{equation}
which can be determined via efficient geometric intersection check primitives.

\subsubsection{Passage traversal check by node positioning in Gabriel cells} 
\begin{figure}[t]
    \minipage{1 \columnwidth}
    \centering
     \includegraphics[width= 0.85 \columnwidth]{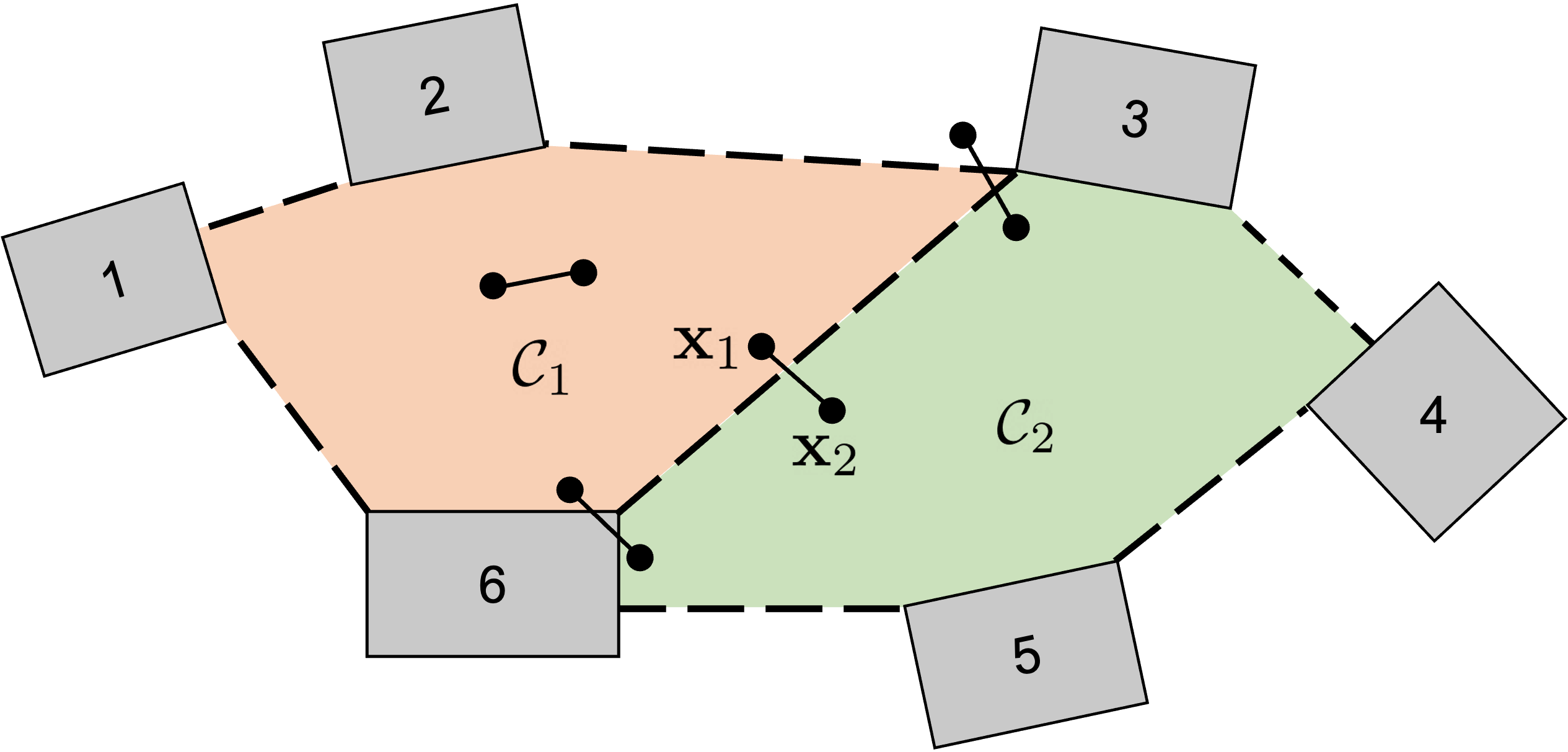}
     \endminipage \hfill
     \caption{If edge $(\mathbf{x}_1, \mathbf{x}_2)$ traverses any passage, $\mathbf{x}_1, \mathbf{x}_2$ are in different Gabriel cells. The reverse also holds on the condition that $(\mathbf{x}_1, \mathbf{x}_2)$ is feasible. If $\mathbf{x}_1, \mathbf{x}_2$ are in different cells but $(\mathbf{x}_1, \mathbf{x}_2)$ passes no passages, $(\mathbf{x}_1, \mathbf{x}_2)$ must collide with obstacles.}
     \label{Fig: Edge and Cell Position}
\end{figure}

For a collision-free edge $(\mathbf{x}_1, \mathbf{x}_2) \subset \mathcal{X}_{free}$ and the Gabriel cell $\mathcal{C}_1$ in which $\mathbf{x}_1$ is located ($\mathbf{x}_1 \in \mathcal{C}_1$), \texttt{SameCell}$(\mathbf{x}_1, \mathbf{x}_2)$ returns True if $\mathbf{x}_2$ also lies inside $\mathcal{C}_1$ ($\mathbf{x}_2 \in \mathcal{C}_1$) and False otherwise. This procedure fast examines if $(\mathbf{x}_1, \mathbf{x}_2)$ passes through any passage because only passages enclosing $\mathcal{C}_1$ need to be tested. Namely, the following holds
\begin{equation}
\label{Eqn: Same Cell Check}
\begin{split}
    \texttt{SameCell}(\mathbf{x}_1, \mathbf{x}_2) &= \text{True} \Leftrightarrow \\ 
    \texttt{Traverse}(\mathcal{P}_{i-j}, \mathbf{x}_1, \mathbf{x}_2) &= \text{False}, \, \forall \, \mathcal{P}_{i-j} \in \mathcal{E}_1
\end{split}
\end{equation}
where $\mathcal{E}_1$ is the set of passage edges enclosing $\mathcal{C}_1$.
$\mathcal{C}_1$ usually contains edges from obstacle boundaries.  
(\ref{Eqn: Same Cell Check}) holds due to the premise that $(\mathbf{x}_1, \mathbf{x}_2)$ is collision-free, which excludes the possibility that $(\mathbf{x}_1, \mathbf{x}_2)$ collides with an obstacle, passes no passages, and $\mathbf{x}_1$ and $\mathbf{x}_2$ belongs to two different cells. The illustration of various placements for an edge in cells is shown in \hyperref[Fig: Edge and Cell Position]{Figure \ref{Fig: Edge and Cell Position}}. 

Based on this routine, node positioning in Gabriel cells is completed rapidly. Nodes need to be located in cells, i.e., determining the \texttt{cellIndex} attribute, to retrieve passages enclosing them. Without iteratively examining all cells or relying on complicated point location query algorithms such as trapezoidal maps \citep{mark2008computational}, the proximity of adjacent path nodes enables fast sequential node positioning. Suppose $\mathbf{x}_1$ is previously located and a new node $\mathbf{x}_2$ is linked to $\mathbf{x}_1$ through edge $(\mathbf{x}_1, \mathbf{x}_2)$. To determine $\mathbf{x}_2.$\texttt{cellIndex}, it suffices to investigate what cells $(\mathbf{x}_1, \mathbf{x}_2)$ traverses and the cells to be tested are thus limited. As in Algorithm \ref*{Alg: Node Positioning in Cells}, starting from the cell $\mathbf{x}_1$ is inside, the script examines each Gabriel cell in the order $(\mathbf{x}_1, \mathbf{x}_2)$ traverses until reaching the one $\mathbf{x}_2$ is located in. Since the edge is generally short compared with cell sizes, only a small number of cells will be passed before termination, resulting in an efficient node positioning and passage traversal check procedure. Meanwhile, other related attributes of $\mathbf{x}_2$ are determined in this positioning procedure as in Algorithm \ref*{Alg: Node Positioning in Cells}.
\begin{algorithm}[t]   	
\SetNlSty{}{}{:}
\SetAlgoLined
\NoCaptionOfAlgo
\setcounter{algocf}{\getrefnumber{Alg: Gabriel Cell Detection}}
\small{
    \SetKwProg{kdtreenode}{Struct}{:}{\nl end}
    \nl \kdtreenode{Node}{
    \nl Point \texttt{position}\;
    \nl Node *\texttt{parent}, *\texttt{leftChild}, *\texttt{rightChild}\;
    \nl list$<$Node*$>$ \texttt{children} (or \texttt{adjacencyList})\;
    \nl int \texttt{cellIndex}\;
    \nl double \texttt{len}, \texttt{cost}\;
    \nl array$<$double$>$ \texttt{passedWidths}\;
    \nl list$<$double$>$ \texttt{sortedWidths}\;
    }
    }
    \caption{\small \textbf{Data Structure}: Path Node Structure}
\end{algorithm}        
\begin{algorithm}[t]
\SetKw{logicalAnd}{and}
\SetKw{logicalOr}{or}
\SetNlSty{}{}{:}
\small{
    \nl Input $\mathbf{x}_1, \mathbf{x}_2$ with known $\mathbf{x}_1.$\texttt{cellIndex} and $l(\mathbf{x}_1, \mathbf{x}_2)$ being collison-free\;
    \nl $c_{id} \leftarrow \mathbf{x}_1.\texttt{cellIndex}; \, \text{positioned} \leftarrow \text{False}$\;
    \nl $C \leftarrow \emptyset; P \leftarrow \emptyset$\tcp*{record passed cell indices and passages}
    \nl \While{\textnormal{positioned} == False}{
        \nl $P_{t} \leftarrow$ set of passages $l(\mathbf{x}_1, \mathbf{x}_2)$ traverses in $\mathcal{E}_{c_{id}}$\;
        \nl \If{$P_{t} == \emptyset$ \logicalOr $P_{t} \subset P$} {
            \nl positioned $\leftarrow$ True\;
            \nl $\mathbf{x}_2.\texttt{cellIndex} \leftarrow c_{id}$\;
            \tcp{assign other attributes}
            \nl $\mathbf{x}_2.\texttt{parent} \leftarrow \mathbf{x}_1; \, \mathbf{x}_1.$\texttt{children}.insert($\mathbf{x}_2$)\;
            \nl $\mathbf{x}_2.\texttt{len} \leftarrow \mathbf{x}_1.\texttt{len} + \text{Len}(l(\mathbf{x}_1, \mathbf{x}_2))$\;
            \nl $\mathbf{x}_2.\texttt{passedWidths} \leftarrow$ passage widths in $P$\;
            \nl $\mathbf{x}_2.\texttt{sortedWidths} \leftarrow \mathbf{x}_1.\texttt{sortedWidths}$\;
            \nl Insert passage widths in $P$ to $\mathbf{x}_2.\texttt{sortedWidths}$\;
            \nl $\mathbf{x}_2.\texttt{cost} \leftarrow \texttt{Cost}(\mathbf{x}_2)$\;
        }
        \nl \Else {
            \nl $\mathcal{P} \leftarrow$ the newly passed passage in $P_{t} \backslash P$\;
            \nl $c_{id} \leftarrow$ new cell index incident to $\mathcal{P}$ and not in $C$\; 
            \nl $P \leftarrow P \cup P_{t}; \, C \leftarrow C \cup \{ c_{id} \}$\;
        }
    }
    }
    \caption{\small \texttt{PositionNode}$(\mathbf{x}_1, \mathbf{x}_2)$}
    \label{Alg: Node Positioning in Cells}
\end{algorithm}

\subsubsection{Cost computation}
Denote the current optimal path from the start $\mathbf{x}_0$ to $\mathbf{x}$ as $\sigma'_\mathbf{x}$. \texttt{Cost}($\mathbf{x}$) returns the optimal path cost, i.e., $\texttt{Cost}(\mathbf{x}) = f_c(\sigma'_\mathbf{x})$, and this cost is stored in $\mathbf{x}.\texttt{cost}$. $l(\mathbf{x}_1, \mathbf{x}_2)$ represents the straight-line path from $\mathbf{x}_1$ to $\mathbf{x}_2$. Suppose $l(\mathbf{x}_1, \mathbf{x}_2)$ is collision-free, \texttt{NewCost}$(\mathbf{x}_1, \mathbf{x}_2)$ computes the cost of the concatenated path $\sigma'_{\mathbf{x}_1} | l(\mathbf{x}_1, \mathbf{x}_2)$ as the cost to come of $\mathbf{x}_2$ through $\mathbf{x}_1$. As $l(\mathbf{x}_1, \mathbf{x}_2)$ is newly concatenated, variables used in cost computation are first updated. For example, a sorted passage width vector $\mathbf{p}' = \mathbf{p}_{\sigma'_{\mathbf{x}_1}}$ is assigned for the joined path.
$\mathbf{p}'$ is updated if $l(\mathbf{x}_1, \mathbf{x}_2)$ traverses passages. Then the cost is calculated accordingly. See the detailed procedure in Algorithm \ref*{Alg: Compute New Cost} where the cost type is easily configurable.

\subsubsection{Subtree node attributes update}
A key operation to attain path optimality in many sampling-based planners is modifying edge connections locally in the neighborhood of new samples, e.g., the rewire routine in RRT$^*$. The effect of changing a node's edge connection propagates to its descendants. Assume edge $(\mathbf{x}_1, \mathbf{x}_2)$ is newly established by changing $\mathbf{x}_2$'s parent to $\mathbf{x}_1$. Paths to $\mathbf{x}_2$'s descendants change accordingly. $\mathbf{x}_2$ is the root of the subtree composed of its descendant nodes. Subtree nodes' path attributes are updated using the new path $\sigma'_{\mathbf{x}_1} | l(\mathbf{x}_1, \mathbf{x}_2) | \sigma'(\mathbf{x}_2 \sim \mathbf{x})$, where $\sigma'(\mathbf{x}_2 \sim \mathbf{x})$ represents the path from $\mathbf{x}_2$ to $\mathbf{x}$ in the subtree. This is done by traversing the subtree with breadth-first search (BFS) or DFS. \texttt{SubtreeUpdate}() in Algorithm \ref*{Alg: Subtree Update} shows a BFS backbone to update subtree node attributes of \texttt{len}, \texttt{cost}, and \texttt{sortedWidths}.

\section{Sampling-based algorithms for PTOPP}
This part elaborates on sampling-based optimal planners for PTOPP integrating previous components. Integration with the two arguably most influential asymptotically optimal planners, optimal PRM (PRM$^*$) and optimal RRT (RRT$^*$), is detailed. Integration principles with their major variants are also briefly discussed.

\subsection{PRM\texorpdfstring{$^*$}{*} and variants for PTOPP}
Primarily aimed at multi-query applications, PRM$^*$ consists of the preprocessing phase constructing a roadmap from random samples in $\mathcal{X}_{free}$ and the query phase searching for the optimal path in this roadmap. In PTOPP, roadmap construction remains unchanged to capture the connectivity of $\mathcal{X}_{free}$. The key to achieving the optimal path lies in modifying the path search procedure. Let the undirected graph $G = (V, E)$ with vertex set $V$ and edge set $E$ represent the roadmap. The nodes in $G$ that $\mathbf{x}_0$ and $\mathbf{x}_g$ connect to are first found by minimizing the cost $f_c(\cdot)$ under path length regulation. Suppose $\mathbf{x}_0$ and $\mathbf{x}_g$ are linked to $\mathbf{x}_{G, 0}, \mathbf{x}_{G, g} \in V$ respectively. 
Alternatively, $\mathbf{x}_0$ and $\mathbf{x}_g$ may also be directly added to the roadmap as nodes. The query phase searches for the optimal path from $\mathbf{x}_{G, 0}$ to $\mathbf{x}_{G, g}$ in $G$. Because of the path cost monotonicity in PTOPP, $G$ is a non-negatively weighted graph. Therefore, optimal path finding equals the shortest path problem where edge weights depend on passages passed by edges. Shortest path finding techniques in weighted graphs can be adopted readily with modifications to compute node costs in PTOPP. 
\begin{algorithm}[t]
\SetNlSty{}{}{:}
\small{
    \nl Input $\mathbf{x}_1, \mathbf{x}_2$ with $l(\mathbf{x}_1, \mathbf{x}_2)$ being collison-free\;
    \nl $\mathbf{p}' \leftarrow \mathbf{x}_1.\texttt{sortedWidths}$\;
    \nl $\text{len} = \mathbf{x}_1.\texttt{len} + \text{Len}(l(\mathbf{x}_1, \mathbf{x}_2))$\;
    \nl \If{\textnormal{\texttt{SameCell}$(\mathbf{x}_1, \mathbf{x}_2)$} == False}{
        \nl $P_t \leftarrow$ passages $l(\mathbf{x}_1, \mathbf{x}_2)$ traverses\;
        \nl Insert passage widths in $P_t$ to $\mathbf{p}'$\;
    }
    \nl \Return path cost in PTOPP\;
    }
    \caption{\small \texttt{NewCost}$(\mathbf{x}_1, \mathbf{x}_2)$}
    \label{Alg: Compute New Cost}
\end{algorithm}
\begin{algorithm}[t]
\SetKw{logicalNot}{not}
\SetNlSty{}{}{:}
\small{
    \nl Input $\mathbf{x}_1, \mathbf{x}_2$ where $\mathbf{x}_1$ is the new parent of $\mathbf{x}_2$\;
    \nl \If{\textnormal{$\mathbf{x}_2.\texttt{parent} \neq$} NULL} {
        \nl $\mathbf{x}_2.$\texttt{parent}.\texttt{children}.remove($\mathbf{x}_2$)\;
        }
    \nl lenChange $\leftarrow \mathbf{x}_1.\texttt{len} + \text{Len}(l(\mathbf{x}_1, \mathbf{x}_2)) - \mathbf{x}_2.\texttt{len}$\;
    \nl \texttt{PositionNode}$(\mathbf{x}_1, \mathbf{x}_2)$\;
    \nl nodeQueue $\leftarrow \{ \mathbf{x}_2 \}$\;
    \nl \While{\textnormal{nodeQueue} $\neq \emptyset$}{
        \nl $\mathbf{x} \leftarrow$ nodeQueue.front()\;
        \nl nodeQueue.removeFront()\;
        \nl \ForEach{\textnormal{$\mathbf{x}_{child} \in \mathbf{x}$.\texttt{children}}} {
            \nl $\mathbf{x}_{child}.\texttt{len} \leftarrow \mathbf{x}_{child}.\texttt{len} + \text{lenChange}$\;
            \nl $\mathbf{x}_{child}.\texttt{sortedWidths} \leftarrow \mathbf{x}.\texttt{sortedWidths}$\;
            \nl \If{$\mathbf{x}_{child}.\textnormal{\texttt{passedWidths}} \neq \emptyset$} {
                \nl Insert $\mathbf{x}_{child}.\texttt{passedWidths}$ into $\mathbf{x}_{child}.$\texttt{sortedWidths}\;
            }
            \nl $\mathbf{x}_{child}.\texttt{cost} \leftarrow \texttt{Cost}(\mathbf{x}_{child})$\;
            \nl nodeQueue.push($\mathbf{x}_{child}$)\;
        }
    }
    }
    \caption{\small \texttt{SubtreeUpdate}$(\mathbf{x}_1, \mathbf{x}_2)$}
    \label{Alg: Subtree Update}
\end{algorithm}

PTOPP built on PRM$^*$ is outlined in Algorithm \ref*{Alg: PRM* for PTOPP}. 
After passage and cell preprocessing, node localization in Gabriel cells is performed once only for $\mathbf{x}_{G,0}$. The query phase adopts Dijkstra's algorithm with two core substitute subroutines. First, the temporary path cost through node $\mathbf{u}$ to its neighbor $\mathbf{v}$ is calculated by \texttt{NewCost}$(\mathbf{u, v})$. Second, if this cost is smaller than $\mathbf{v}.$\texttt{cost}, \texttt{SubtreeUpdate}$(\mathbf{u, v})$ is invoked to update edge connections and node attributes. In PRM$^*$, nodes maintain their adjacent nodes instead of $\texttt{children}$, \texttt{SubtreeUpdate}$(\mathbf{u, v})$ thus terminates early without entering the BFS procedure. Under path length regulation, node precedence is compared lexicographically by the tuple (\texttt{cost}, \texttt{len}). $\mathbf{u}$ is considered to have a smaller cost than $\mathbf{v}$ if
\begin{equation}
\label{Eqn: Node Cost Comparison}
      (\mathbf{u}.\texttt{cost}, \mathbf{u}.\texttt{len}) \prec_{\text{lex}} (\mathbf{v}.\texttt{cost}, \mathbf{v}.\texttt{len}).
\end{equation}
This is utilized when determining the minimum cost node (line \ref*{AlgLine: Minimum Cost} in Algorithm \ref*{Alg: PRM* for PTOPP}) and comparing the temporary cost with node cost (line \ref*{AlgLine: Cost Update Condition} in Algorithm \ref*{Alg: PRM* for PTOPP}).

PRM$^*$ for PTOPP depicts a generic graph-based solution framework comprising roadmap construction and optimal path search. It applies to other optimal PRM$^*$ variants, e.g., simplified PRM (sPRM), which only differ in their roadmap construction. In addition, one important branch is pure cell-based methods enabled by cell decomposition \citep{choset2005principles, lavalle2006planning}. They are particularly promising in PTOPP because no extra cell decomposition step is required. Rather than random samples, the roadmap uses sparse non-random samples in cells, e.g., centroids and edge midpoints, as vertices to capture the connectivity of divided regions. Thanks to the dramatic decrease in node number, pure cell-based methods enjoy high computational efficiency, though heavy path postprocessing is usually needed.
\begin{algorithm}[t]
\SetKw{logicalAnd}{and}
\SetKw{logicalOr}{or}
\SetNlSty{}{}{:}
\small{
    \tcp{preprocessing phase}
    \nl Input $\mathbf{x}_0, \mathbf{x}_g$ and conduct passage and cell preprocessing.\;
    \nl Construct $G = (V, E)$ via random sampling.\;
    \tcp{query phase}
    \nl $\mathbf{x}_{G, 0}, \mathbf{x}_{G, g} \leftarrow X; Q \leftarrow V; \mathbf{x}_{G, 0}.\texttt{cost} \leftarrow \texttt{Cost}(\mathbf{x}_{G, 0})$\;
    \nl Locate $\mathbf{x}_{G, 0}$ in Gabriel cells\;
    \nl \While{$Q \neq \emptyset$} {
        \nl \label{AlgLine: Minimum Cost}$\mathbf{u} \leftarrow$ node in $Q$ with the minimum \texttt{cost}\; 
        \nl \If{$\mathbf{u}$ == $\mathbf{x}_{G,g}$} {
            \nl break\;
        }
        \nl $Q \leftarrow Q \backslash \{ \mathbf{u} \}; A_u \leftarrow$ adjacency list of $\mathbf{u}$\;
        \nl \ForEach{$\mathbf{v} \in A_u$ \logicalAnd $\mathbf{v} \in Q$ } {
            \nl $\text{tempCost} \leftarrow \texttt{NewCost}(\mathbf{u, v})$\;
            \nl \label{AlgLine: Cost Update Condition}\If{\textnormal{tempCost} $< \mathbf{v}.\textnormal{\texttt{cost}}$ \\ \logicalOr \textnormal{(}\textnormal{tempCost} == \textnormal{$\mathbf{v}.\texttt{cost}$ \logicalAnd $\mathbf{u}.\texttt{Len} + \text{len}(l(\mathbf{u, v})) < \mathbf{v}.\texttt{len}$)}} {
                \nl \texttt{SubtreeUpdate}$(\mathbf{u, v})$;
            }
        }
    }
    \nl $ \sigma \leftarrow \texttt{GetPath}(\mathbf{x}_{G,0}, \mathbf{x}_{G,g})$\;
    \nl \Return $l(\mathbf{x}_0, \mathbf{x}_{G, 0}) \, | \, \sigma \, | \, l(\mathbf{x}_{G, g}, \mathbf{x}_g)$\;
    }
    \caption{\small PRM$^*$ for PTOPP}
    \label{Alg: PRM* for PTOPP}
\end{algorithm}

\subsection{RRT\texorpdfstring{$^*$}{*} and variants for PTOPP}
RRT$^*$ emerges as the asymptotically optimal version of RRT. It introduces a rewire procedure that locally modifies existing edges every time a new node is inserted, ensuring that all nodes are reached with the minimum cost from the root. 
To do this in PTOPP, the backbone of RRT$^*$ is retained. At first, $\mathbf{x}_0$ is localized in cells. Core modifications are introduced similarly to above to correctly maintain the minimum cost path to each node. Again, the addition of \texttt{NewCost}() and \texttt{SubtreeUpdate}() is crucial for cost computation and dynamic tree maintenance. The proposed algorithms are cost-agnostic and the cost type is conveniently customizable in \texttt{NewCost}(). \texttt{SubtreeUpdate}() is invoked when creating the edge along the minimum cost path to the newly sampled $\mathbf{x}_{new}$ and adjusting existing edges in the rewire routine, which alter edges and propagate attribute changes in the tree. Lexicographic comparison of (\texttt{cost}, \texttt{len}) is used in choosing the parent node and the rewire condition (line \ref*{AlgLine: Rewire Condition} in Algorithm \ref*{Alg: Rewiring Based on Cost and Length}). RRT$^*$ for PTOPP  in the architecture of vanilla RRT$^*$ is detailed in Algorithm \ref*{Alg: RRT* for PTOPP Concise}.

Another asymptotically optimal algorithm closely related to RRT$^*$ is rapidly exploring random graph (RRG) which incrementally builds an undirected graph. $\mathbf{x}_{new}$ connects to all collision-free nearby nodes $\mathbf{x}_{near} \in X_{near}$, leading to a much richer edge set that contains RRT$^*$ edge set as its subset in the same sampling sequence. To adapt RRG for PTOPP, its graph construction phase remains unchanged. Subsequent optimal path search follows the query phase in Algorithm \ref*{Alg: PRM* for PTOPP}. Namely, RRG for PTOPP fits into the generic pipeline in Algorithm \ref*{Alg: PRM* for PTOPP} and merely differs in graph construction. Additionally, the asymptotically optimal \textit{k}-nearest variants of PRM$^*$, RRT$^*$, and RRG, in which new edges are sought to \textit{k} nearest neighbors of $\mathbf{x}_{new}$ and \textit{k} varies with the vertex number $| V |$, are also compatible with PTOPP. They only differ in the near neighbors search procedure. 
\begin{algorithm}[t]
\SetKw{logicalAnd}{and}
\SetKw{logicalOr}{or}
\SetNlSty{}{}{:}
\small{
    \nl Input $\mathbf{x}_{new}$ and its conlision-free neighbor set $X_{near}$\;
    \nl \ForEach{$\mathbf{x}_{near} \in X_{near}$} {
        \nl \label{AlgLine: Rewire Condition} \If{\textnormal{\texttt{NewCost}$(\mathbf{x}_{near}, \mathbf{x}_{new}) < \mathbf{x}_{near}.\texttt{cost}$ \logicalOr (\texttt{NewCost}$(\mathbf{x}_{near}, \mathbf{x}_{new}) == \mathbf{x}_{near}.\texttt{cost}$ \logicalAnd $\mathbf{x}_{new}.\texttt{len} + \text{Len}(l(\mathbf{x}_{new}, \mathbf{x}_{near})) < \mathbf{x}_{near}.\texttt{len}$)}} {
                \nl $E \leftarrow E \backslash \{ (\mathbf{x}_{near}.\texttt{parent}, \mathbf{x}_{near}) \} \cup \{ (\mathbf{x}_{new}, \mathbf{x}_{near}) \}$\; 
                \nl \texttt{SubtreeUpdate}$(\mathbf{x}_{new}, \mathbf{x}_{near})$\;
            }
        }
    }
    \caption{\small Rewire With Length-Regulation in RRT$^*$}
    \label{Alg: Rewiring Based on Cost and Length}
\end{algorithm}
\begin{algorithm}[t]
\SetKw{logicalNot}{not}
\SetKw{logicalAnd}{and}
\SetKw{logicalOr}{or}
\SetNlSty{}{}{:}
\small{
    \nl Input $\mathbf{x}_0, \mathbf{x}_g$ and conduct passages and cell preprocessing\;
    \nl $V \leftarrow \{ \mathbf{x}_0 \}$; $E \leftarrow \emptyset$; $f_c \leftarrow \infty$\;
     \nl Locate $\mathbf{x}_0$ in Gabriel cells and $\mathbf{x}_0.\texttt{cost} \leftarrow \texttt{Cost}(\mathbf{x}_0)$\;
    \nl \For {$i \leftarrow 1$ \KwTo $n$} {
        \nl $\mathbf{x}_{rand} \leftarrow \texttt{SampleFree}()$\;
        \nl $\mathbf{x}_{nearest} \leftarrow \texttt{Nearest}(G = (V, E), \mathbf{x}_{rand})$\;
        \nl $\mathbf{x}_{new} \leftarrow \texttt{Steer}(\mathbf{x}_{nearest}, \mathbf{x}_{rand})$\;
        \nl \If{\textnormal{\texttt{ObstacleFree}$(\mathbf{x}_{nearest}, \mathbf{x}_{new})$}} {
            \nl $X_{near} \leftarrow \texttt{Near}(G = (V, E), \mathbf{x}_{new}, r_{RRT^*})$\;
            \nl $\mathbf{x}_{min} \leftarrow \texttt{GetParent}(X_{near}, \mathbf{x}_{new})$\;
            \nl \texttt{SubtreeUpdate}$(\mathbf{x}_{min}, \mathbf{x}_{new})$\;
            \nl Rewire\;
            \nl \If{\textnormal{Len$(l(\mathbf{x}_{new}, \mathbf{x}_g)) < r_{RRT^*}$
            \logicalAnd \texttt{ObstacleFree}$(\mathbf{x}_{new}, \mathbf{x}_{g})$ 
            \logicalAnd $\texttt{NewCost}(\mathbf{x}_{new}, \mathbf{x}_g) < f_c$}}{
                \nl $\mathbf{x}_g.\texttt{parent} \leftarrow \mathbf{x}_{new}$\;
                \nl $f_c \leftarrow \texttt{NewCost}(\mathbf{x}_{new}, \mathbf{x}_g)$\;
                }
        }
        }
        \nl \Return \texttt{GetPath}$(\mathbf{x}_0, \mathbf{x}_g)$\;
    }
    \caption{\small RRT$^*$ for PTOPP}
    \label{Alg: RRT* for PTOPP Concise}
\end{algorithm}

\subsection{Algorithm analysis}
\subsubsection{Probabilistic completeness} 
Assume there is a goal region $\mathcal{X}_{goal} \subset \mathcal{X}_{free}$, a path $\sigma$ reaches the goal if $\sigma(1) \in \mathcal{X}_{goal}$. Probabilistic completeness requires the probability that an algorithm finds a feasible path in any robustly feasible problem, i.e., problems having a feasible path of positive clearance to obstacles, converges to one as the sample number $n$ goes to infinity. It is easy to see that PRM$^*$ for PTOPP, together with its variants, is probabilistic complete since the graph construction phase is not modified, and so is RRG. For RRT$^*$-based avenues, though the resulting edge set $E$ differs from the vanilla RRT$^*$ due to new path costs, the vertex set $V$ is identical for the same sampling sequence. Meanwhile, the graph is connected by construction. Hence, its probabilistic completeness directly follows from that of RRT$^*$. Moreover, because the constructed graphs are unchanged or connected, the exponential decay in the failure probability of finding a path for a feasible planning problem with $n$ still holds.

\subsubsection{Asymptotic optimality}
Suppose an OPP problem admits a robustly optimal solution, i.e., the optimal path $\sigma^*$ is homotopic to some feasible path with positive clearance.  
An algorithm is asymptotically optimal if the minimum cost it finds converges to the optimal cost as $n$ approaches infinity. Since graph construction remains unchanged from original PRM$^*$, RRG, and their variants in PTOPP, the algorithms' asymptotic optimality is inherited because path costs are compatible. More specifically, the unchanged graph structures ensure that a sequence of paths converging to the optimal path exists with probability one for all large $n$. Using the monotonicity of the best path cost with respect to $n$, the minimum cost these algorithms return converges to $f_c(\sigma^*)$ almost surely. For PTOPP based on RRT$^*$, only path cost computations are customized. It is guaranteed that the best path starting from $\mathbf{x}_0$ and reaching each new vertex $\mathbf{x}_i$ is found and each vertex has a unique parent. Then the asymptotical optimality follows. 
\begin{figure*}[t]
    \centering
    \minipage{2 \columnwidth}
    \centering
    \subfigure[] {
    \label{Fig: Passage and Cell Detection Results a}
    \frame{\includegraphics[width= 0.32 \columnwidth]{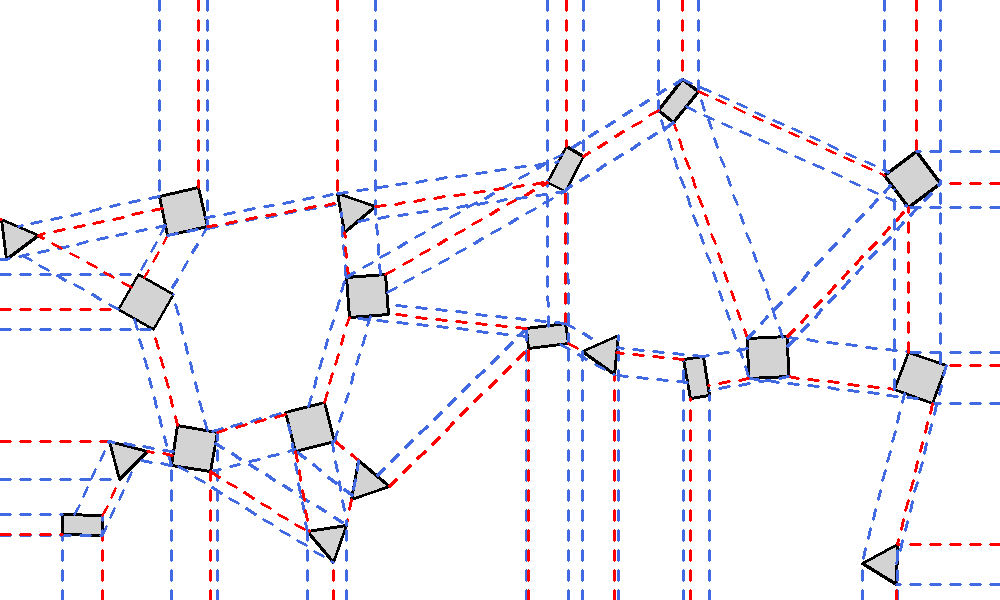}}}
    \subfigure[] {
    \label{Fig: Passage and Cell Detection Results b}
    \frame{\includegraphics[width= 0.32 \columnwidth]{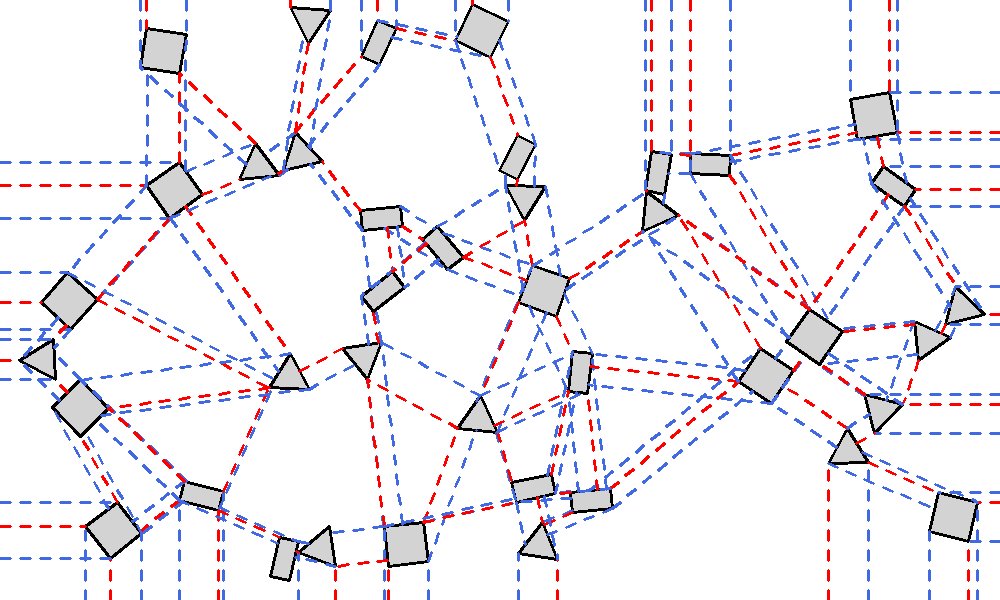}}}
    \subfigure[] {
    \label{Fig: Passage and Cell Detection Results c}
    \frame{\includegraphics[width= 0.32 \columnwidth]{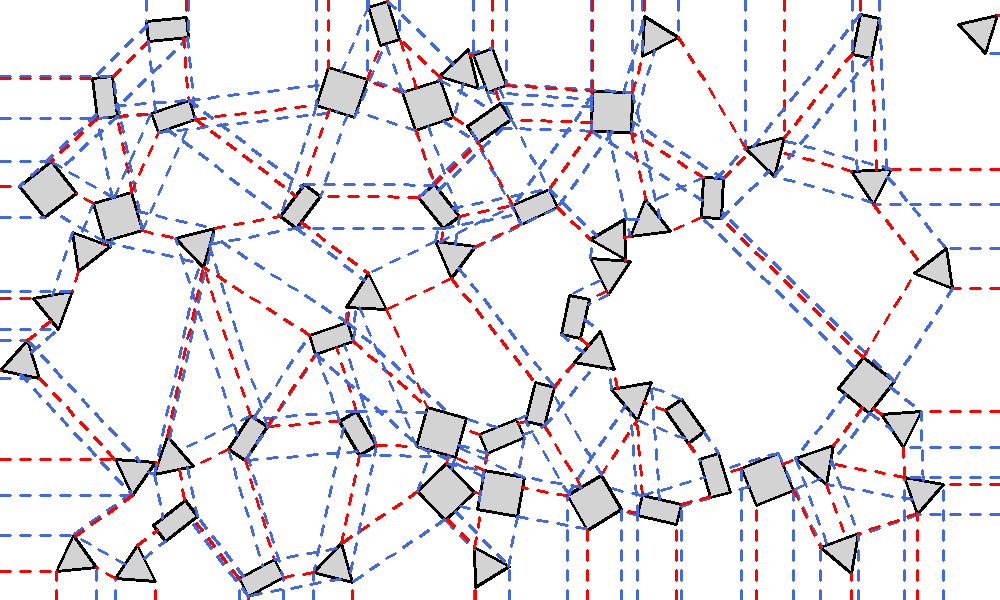}}}      
    \subfigure[] {
    \label{Fig: Passage and Cell Detection Results d}
    \frame{\includegraphics[width= 0.32 \columnwidth]{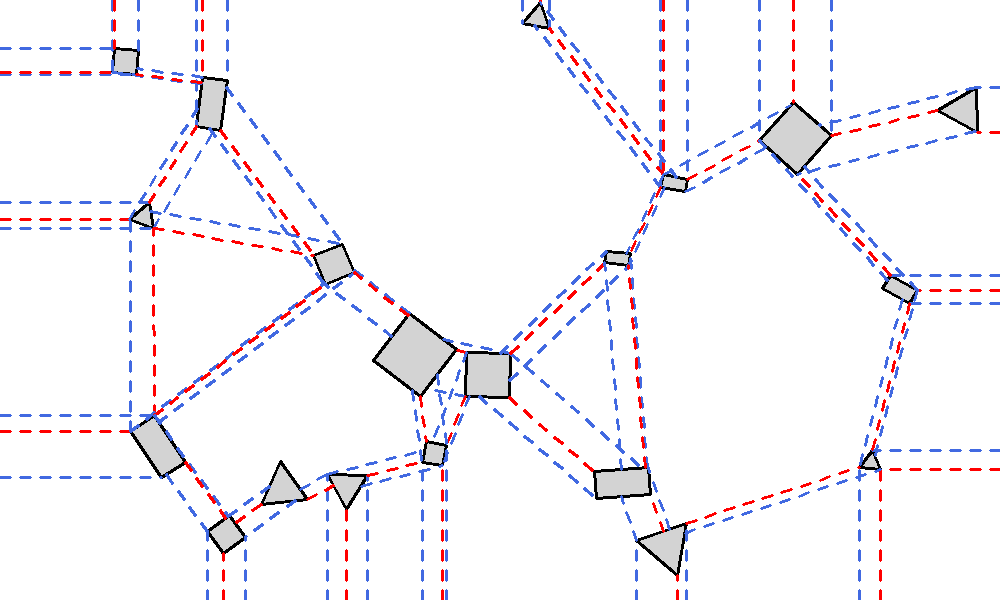}}}
    \subfigure[] {
    \label{Fig: Passage and Cell Detection Results e}
    \frame{\includegraphics[width= 0.32 \columnwidth]{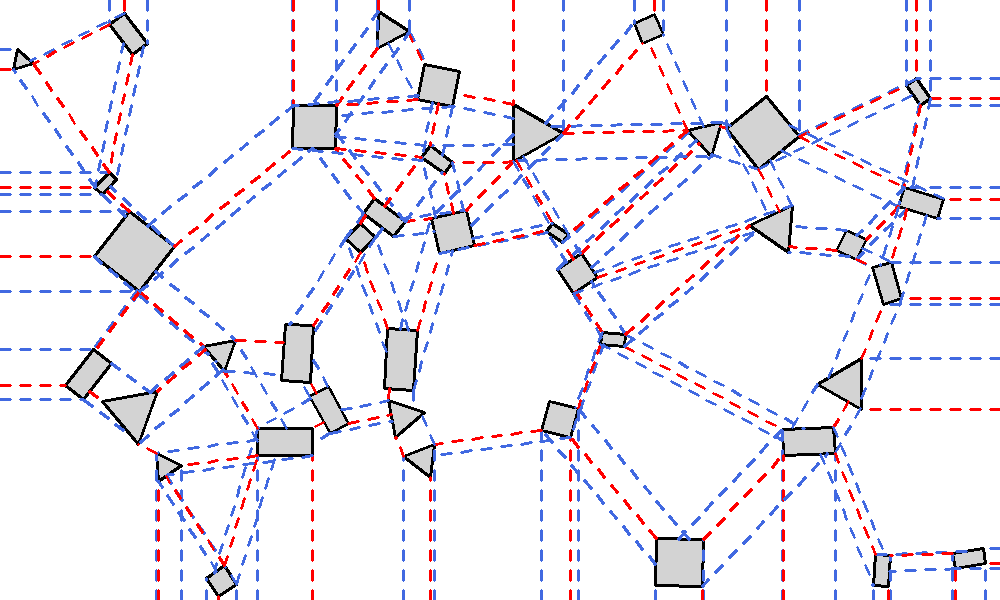}}}
    \subfigure[] {
    \label{Fig: Passage and Cell Detection Results f}
    \frame{\includegraphics[width= 0.32 \columnwidth]{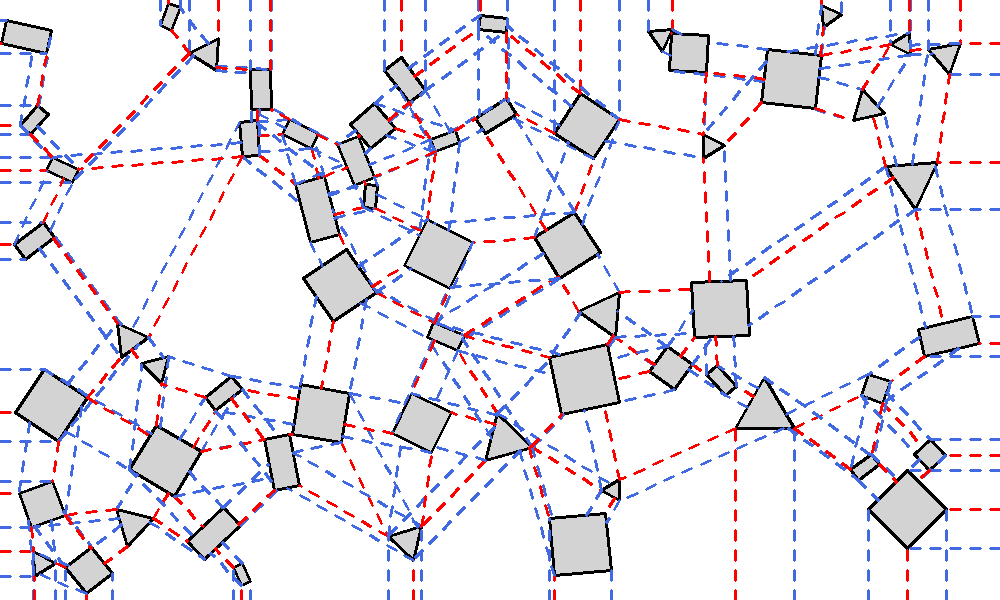}}}  
    \endminipage  \hfill 
    \caption{Examples of passage and Gabriel cell distributions in obstacle-dense environments. Passage region boundaries and passage segments are dashed segments in blue and red, respectively. (a)-(c) have similar obstacle sizes (side length is 40) and the obstacle number $m =$ 20, 40, and 60, respectively. (d)-(f) have obstacle side lengths randomized in [20, 60].}
    \label{Fig: Passage and Cell Detection Results}
\end{figure*}

\subsubsection{Computational complexity} 
Graph construction takes $O(n \log n \log^d m)$ time under efficient implementations of core procedures.  
However, the complexity of subtree update in RRT$^*$ is not taken into account in the original analysis in \citep{karaman2011sampling}. The total edge number $|E|$ in RRT$^*$ is of $O(n)$, implying that the average edge changes in the rewiring procedure are of $O(1)$. Doing graph traversal, the complexity of \texttt{SubtreeUpdate}() is determined by the node number of the subtree branch. RRT$^*$ maintains a tree similar to the random online nearest neighbor graph. 
The expected subtree node number rooted at a random node is no more than $\log n$ \citep{lichev2024new}. The total complexity thus remains.

New computational overhead in PTOPP is introduced by passage and cell detection, and passage traversal check. Passage and cell detection using Delaunay graphs takes $O(m \log m)$ time. As $n \gg m$, most new computations occur in passage traversal check. One-time check in (\ref{Eqn: Segment Intersection Check}) takes constant time and the number of passages examined for each edge decides the cumulative computations. The total passage number after Gabriel condition can be approximated by the edge number $O(m)$ in Gabriel graphs \citep{mark2008computational, huang2024homotopic}. Brute-force traversal yields $O(nm)$ complexity and dramatically increases the planning time. By positioning nodes in Gabriel cells, only a constant number of passages need to be examined. The average cell number an edge passes does not exceed $O(1)$ given $n \gg m$ in $\mathbb{R}^2$ (the cell size is $O(1/m) \times O(1/m)$ and the edge length is $O((\log (n) / n)^\frac{1}{2})$ in a unit square. Meanwhile, $O(1)$ passage sides exist in a cell. 
Thereby, the time complexity of graph construction remains. Summing up, the overall time complexity is $O(n \log n \log^d m + m \log m)$ for passage and cell detection, and graph construction in PTOPP.

In the path query phase to extract the optimal path from the graph, the time complexity is the same as the shortest path problem with nonnegative edge lengths, whose asymptotic optimal bound is $O(|V| \log |V| + |E|)$ \citep{schrijver2003combinatorial}. Trivially, $|V| \in O(n)$ for all algorithms. $|E| \in O(n \log n)$ for PRM$^*$, RRG, and their \textit{k}-nearest variants. $|E| \in O(n)$ for RRT$^*$ and its \textit{k}-nearest variant \citep{karaman2011sampling}. Since node parent is maintained online in RRT$^*$, no separate path finding routine is required. In summary, after efficient passage detection, PTOPP is solved without increasing the time complexity of sampling-based optimal planners. As for space complexity, apart from $O(|V| + |E|)$ space for graph storage, only $O(m)$ extra space is needed to store passages and cells.

\section{Experimental results}
The proposed PTOPP algorithms, including their dependent core modules, are implemented and thoroughly evaluated. This section reports comprehensive evaluation results. Code is updated at \href{https://github.com/HuangJingGitHub/PTOPP}{https://github.com/HuangJingGitHub/PTOPP}.

\subsection{Passage detection and cell decomposition using Gabriel graphs}
Passage detection and Gabriel cell decomposition are crucial preprocessing steps before PTOPP. Experiments in this part aim to reveal the resulting passage and cell distributions using the proposed detection condition and the detection efficiency improvement over the previous direct check strategy. To this end, experiments are carried out in a wide variety of obstacle-dense environments. Passage detection relies on the Delaunay graph of obstacle centroids. After casting obstacles into the centroid set $C_c$, many efficient Delaunay triangulation implementations are available, such as OpenCV and CGAL \citep{boissonnat2000triangulations} libraries. When extracting passages from the Delaunay graph $\mathcal{DG}(C_c)$, the geodesic distance threshold $k_{gd} = 2$ is utilized in passage validity check. Throughout experiments, computations are run on a PC with Ubuntu 20.04, Intel Core i7-14650HX CPU$@\SI{2.40}{\GHz} \times 24$, and 32 GB of RAM.

\subsubsection{Passage detection and cell decomposition results}
\begin{figure*}[t]
    \centering
    \minipage{2 \columnwidth}
    \centering
    \subfigure[] {
    \label{Fig: Passage and Cell Detection Results 3D a}
    {\includegraphics[width= 0.32 \columnwidth]{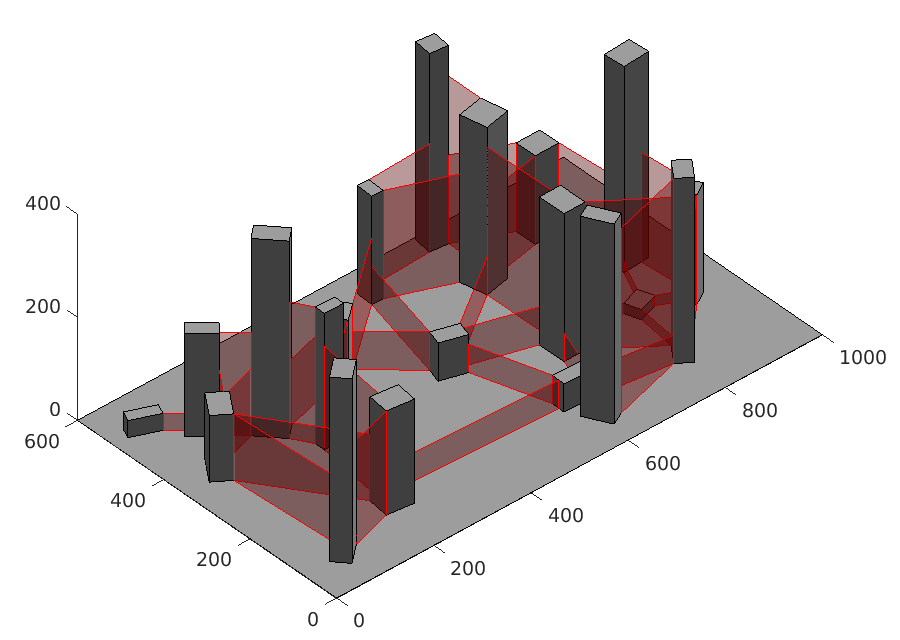}}}
    \subfigure[] {
    \label{Fig: Passage and Cell Detection Results 3D b}
    {\includegraphics[width= 0.32 \columnwidth]{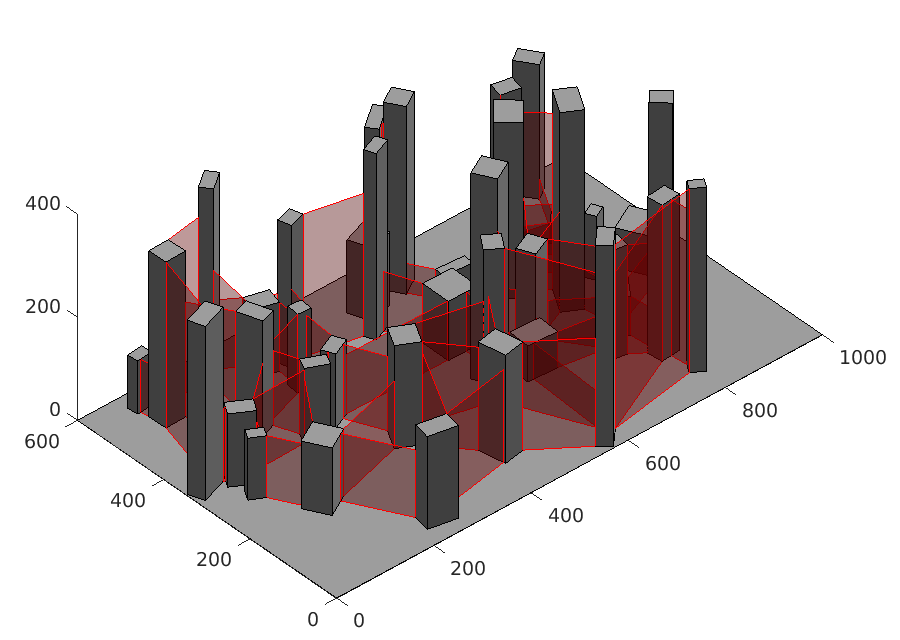}}}
    \subfigure[] {
    \label{Fig: Passage and Cell Detection Results 3D c}
    {\includegraphics[width= 0.32 \columnwidth]{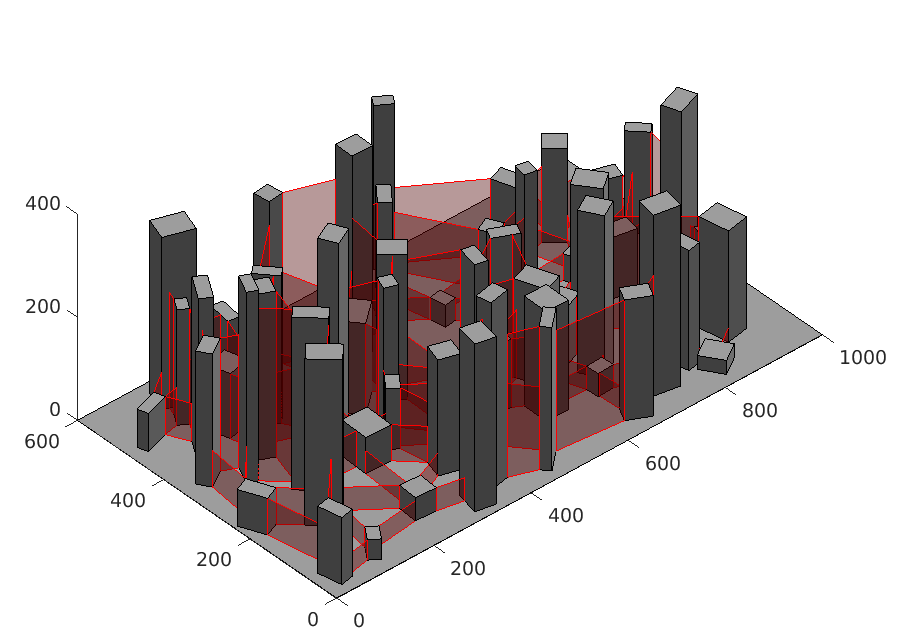}}}  
    \subfigure[] {
    \label{Fig: Passage and Cell Detection Results 3D d}
    {\includegraphics[width= 0.32 \columnwidth]{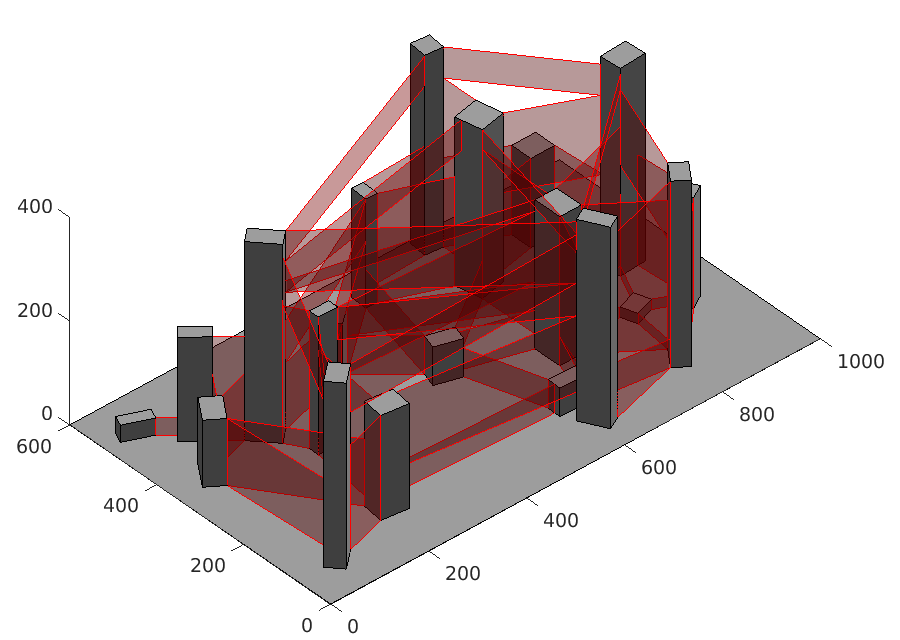}}}
    \subfigure[] {
    \label{Fig: Passage and Cell Detection Results 3D e}
    {\includegraphics[width= 0.32 \columnwidth]{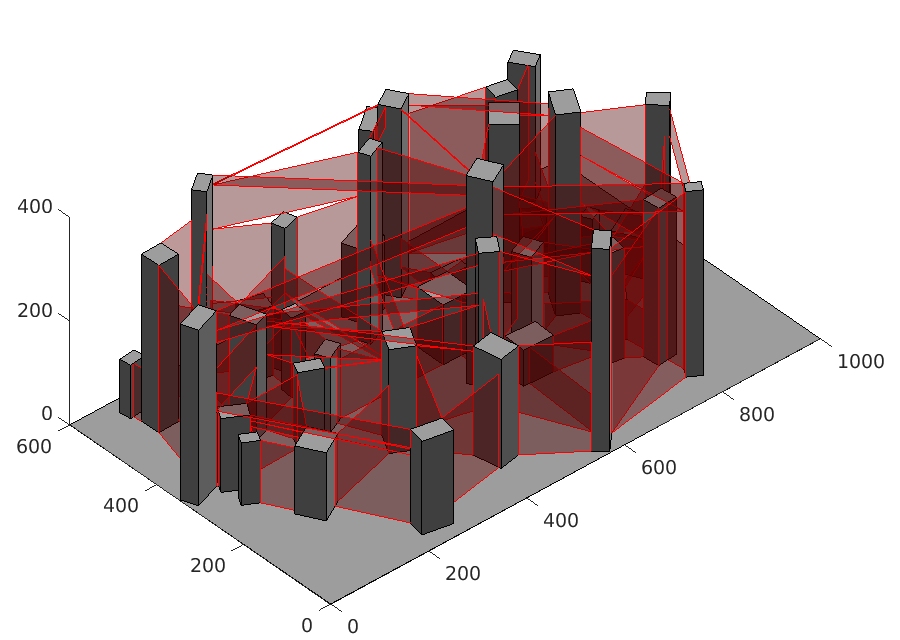}}}
    \subfigure[] {
    \label{Fig: Passage and Cell Detection Results 3D f}
    {\includegraphics[width= 0.32 \columnwidth]{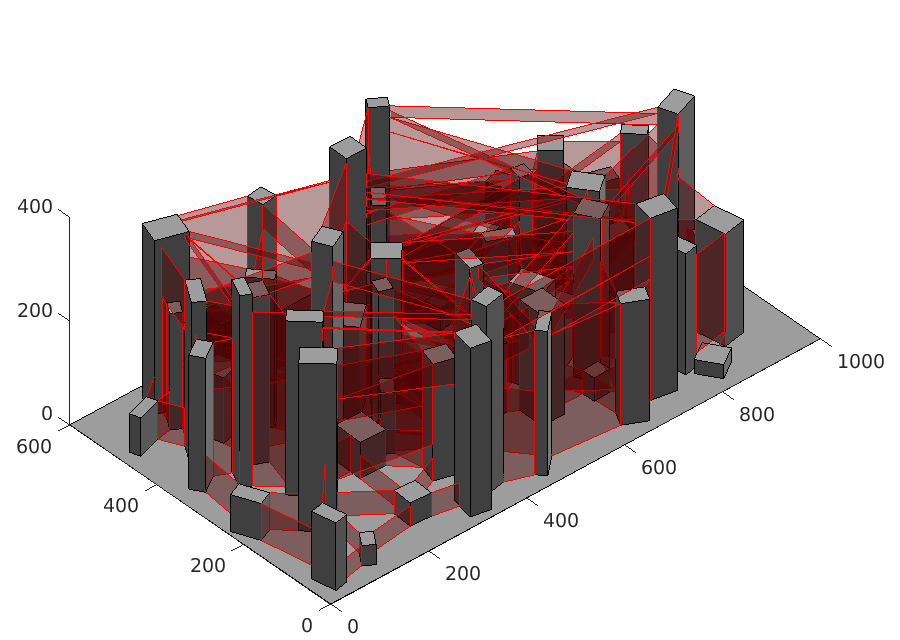}}}    
    \endminipage  \hfill 
    \caption{Examples of passage and Gabriel cell distributions in 3D obstacle-dense environments. Passages are depicted as red transparent planes. (a)-(c) depict passages detected on the ground and the passage height equals the lower incident obstacle height. $m = 20, 40$, and 60, respectively. (d)-(f) depict passages in all height intervals.}
    \label{Fig: Passage and Cell Detection Results 3D}
\end{figure*}

\hyperref[Fig: Passage and Cell Detection Results]{Figure \ref{Fig: Passage and Cell Detection Results}} showcases examples of passages and cells detected amid 2D random obstacles (map size $1000 \times 600$). Passage region bounds are blue dashed lines. Red dashed lines are passage segments, i.e., distances, between obstacle pairs that construct valid passages. Two segment types may overlap. In \hyperref[Fig: Passage and Cell Detection Results a]{Figure \ref{Fig: Passage and Cell Detection Results a}}-\hyperref[Fig: Passage and Cell Detection Results c]{\ref{Fig: Passage and Cell Detection Results c}}, densifying obstacles have an identical side length and thus similar sizes. To get a complete map decomposition, environment walls are processed as adjacent obstacles. In \hyperref[Fig: Passage and Cell Detection Results d]{Figure \ref{Fig: Passage and Cell Detection Results d}}-\hyperref[Fig: Passage and Cell Detection Results f]{\ref{Fig: Passage and Cell Detection Results f}}, obstacles additionally vary in size to enrich randomness. Detected passages and cells are sparse and informative to depict confined locations. They provide passage descriptions more elaborate than simple segments. The proposed passage region definition in (\ref{Eqn: Passage Region}) also endows passages with sizes insensitive to significant obstacle size differences in \hyperref[Fig: Passage and Cell Detection Results d]{Figure \ref{Fig: Passage and Cell Detection Results d}}-\hyperref[Fig: Passage and Cell Detection Results f]{\ref{Fig: Passage and Cell Detection Results f}}.
Spatial passages in 3D environments featuring random obstacles are shown in \hyperref[Fig: Passage and Cell Detection Results 3D]{Figure \ref{Fig: Passage and Cell Detection Results 3D}} (map size $1000 \times 600 \times 400$). Passage segments evolve into gates in 3D space. For clarity, passages incident to the environment walls are not depicted. \hyperref[Fig: Passage and Cell Detection Results 3D a]{Figure \ref{Fig: Passage and Cell Detection Results 3D a}}-\hyperref[Fig: Passage and Cell Detection Results 3D c]{\ref{Fig: Passage and Cell Detection Results 3D c}} plot passages only considering obstacle distribution on the base as in \citep{huang2024homotopic}. They start from the base and end at the lower incident obstacle height. In the bottom row are corresponding spatial passages that provide a comprehensive description of places with restricted accessible free space. 
A notable observation is that passages are sparser at large heights because fewer obstacles are there. Complex decompositions make direct node positioning hard and necessitate the proposed node projection localization in base cells. 

Statistics of passages and cells are showcased in \hyperref[Fig: Passage and Cell Detection Data]{Figure \ref{Fig: Passage and Cell Detection Data}}. 
Since relative sizes are indeterminate, fixed and varying obstacle sizes are tested separately with no evident statistical differences. For brevity, results in common varying-size setups are reported. In comparison, the detection condition that solely utilizes passage segments as a simplified passage representation is used. For each obstacle number $m$, data are collected from 50 random distributions. The linearity between passage segment number and $m$ has been shown in \citep{huang2024homotopic}. \hyperref[Fig: Passage and Cell Detection Data a]{Figure \ref{Fig: Passage and Cell Detection Data a}} demonstrates that the linear increase of passage numbers with respect to $m$ also holds in the amended Gabriel condition. Because of the linearity between triangle and edge numbers in random Delaunay graphs, the linearity between Gabriel cells and passages is expected.
The cell number is linear in the passage number and further, in the obstacle number in \hyperref[Fig: Passage and Cell Detection Data b]{Figure \ref{Fig: Passage and Cell Detection Data b}}. Its value is roughly half the passage number and close to the obstacle number.

For regions and segments in passage representation, while passage regions lead to fewer passages and cells for imposing a more restrictive validity condition, the resulting numbers differ insignificantly among sparse obstacles. 
The reduction becomes more notable in dense obstacles. For example, when $m = 200$, the passage and cell numbers drop \SI{11.3}{\percent} and \SI{20.6}{\percent} respectively. Fewer passages and cells lighten the computational burden in planning. A potential drawback of segment representation is that it fails to reflect how passage and cell change under obstacle size variations. In \hyperref[Fig: Passage and Cell Detection Data c]{Figure \ref{Fig: Passage and Cell Detection Data c}}, consider a fixed number of obstacles (40 herein) with side length in $[10, 100]$. When obstacles expand, passage and cell numbers found via segments increase slightly. This is attributed to the fact that passage segments shrink as obstacle sizes increase. In contrast, fewer passages are presented using passage regions because regions expand laterally with obstacles, which is more favorable.

Statistics of passages in 3D environments are reported in \hyperref[Fig: Passage And Cell Detection Data 3D]{Figure \ref{Fig: Passage And Cell Detection Data 3D}}. Obstacles are assigned random heights in $[0, 400]$. For comparison, planar passage and cell numbers on the base are listed. The data show that the 3D passage number is also linear in the obstacle number. In other words, $O(m)$ valid spatial passages exist. This linearity originates from the fact that all intermediate edges in the incremental Delaunay graph construction process are checked. The ratio of the spatial passage number to the base passage number is $1.7$ on average, suggesting good sparsity of spatial passages. 
Sparity of passages in planar and 3D space is crucial to enabling efficient planning in PTOPP.

\subsubsection{Detection efficiency}
\begin{figure*}[t]
    \centering
    \minipage{2 \columnwidth}
    \centering
    \subfigure[] {
    \label{Fig: Passage and Cell Detection Data a}
    \includegraphics[width= 0.32 \columnwidth]{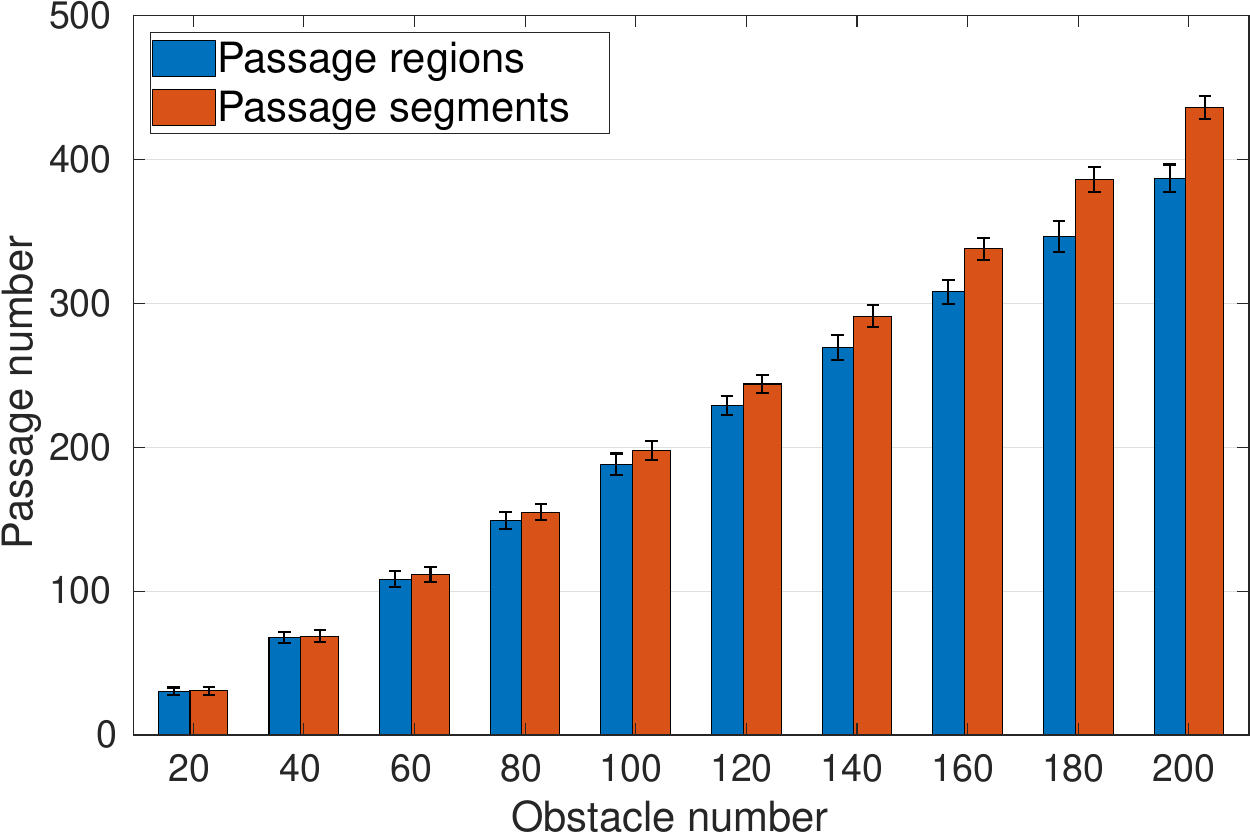}} 
    \subfigure[] {
    \label{Fig: Passage and Cell Detection Data b}
    \includegraphics[width= 0.32 \columnwidth]{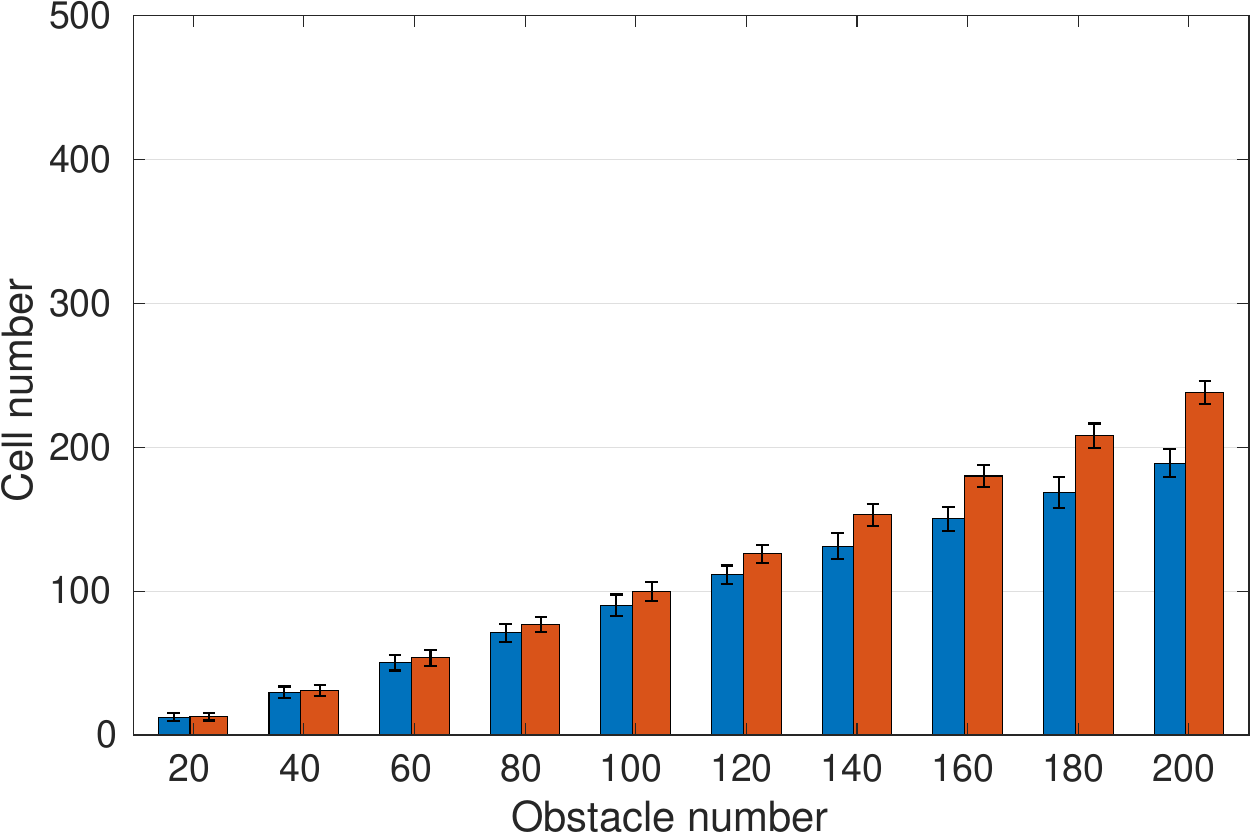}}
    \subfigure[] {
    \label{Fig: Passage and Cell Detection Data c}
    \includegraphics[width= 0.32 \columnwidth]{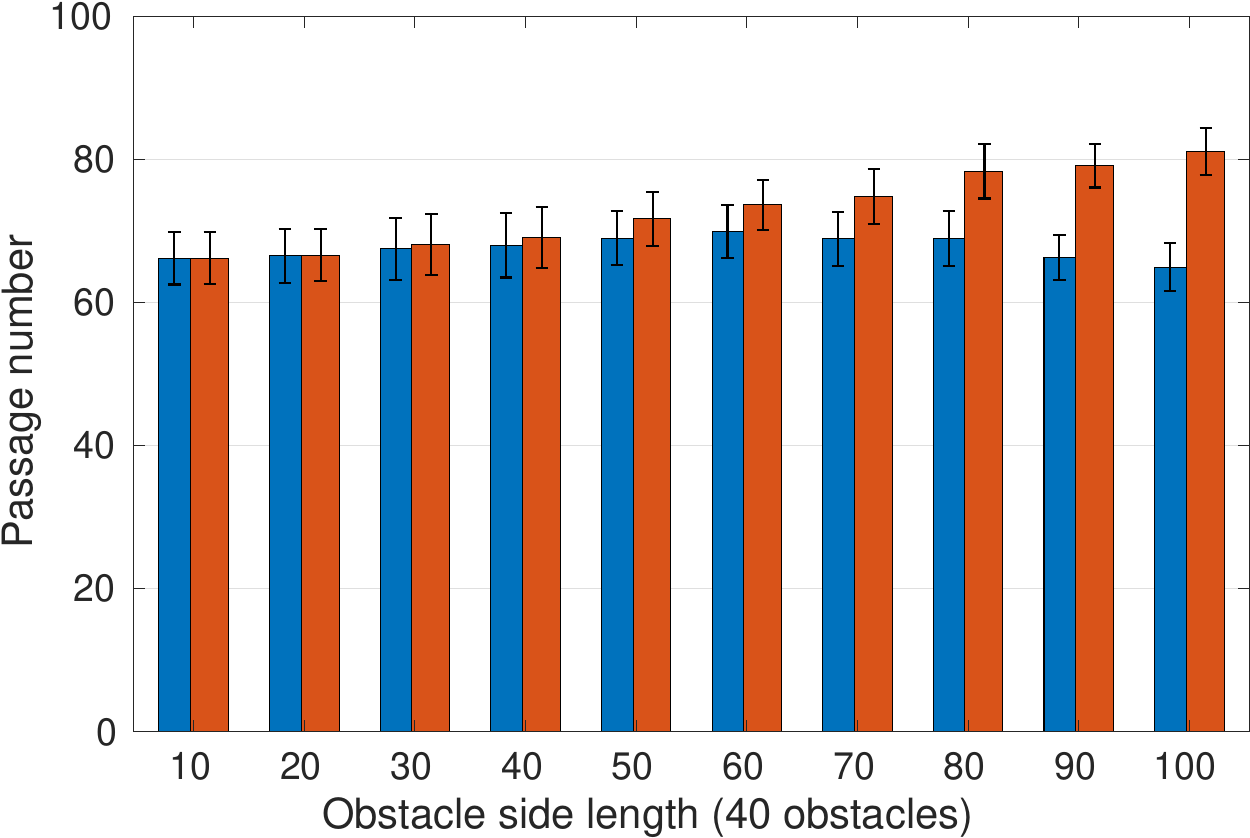}}
    \subfigure[]{
    \includegraphics[width= 0.315 \columnwidth]{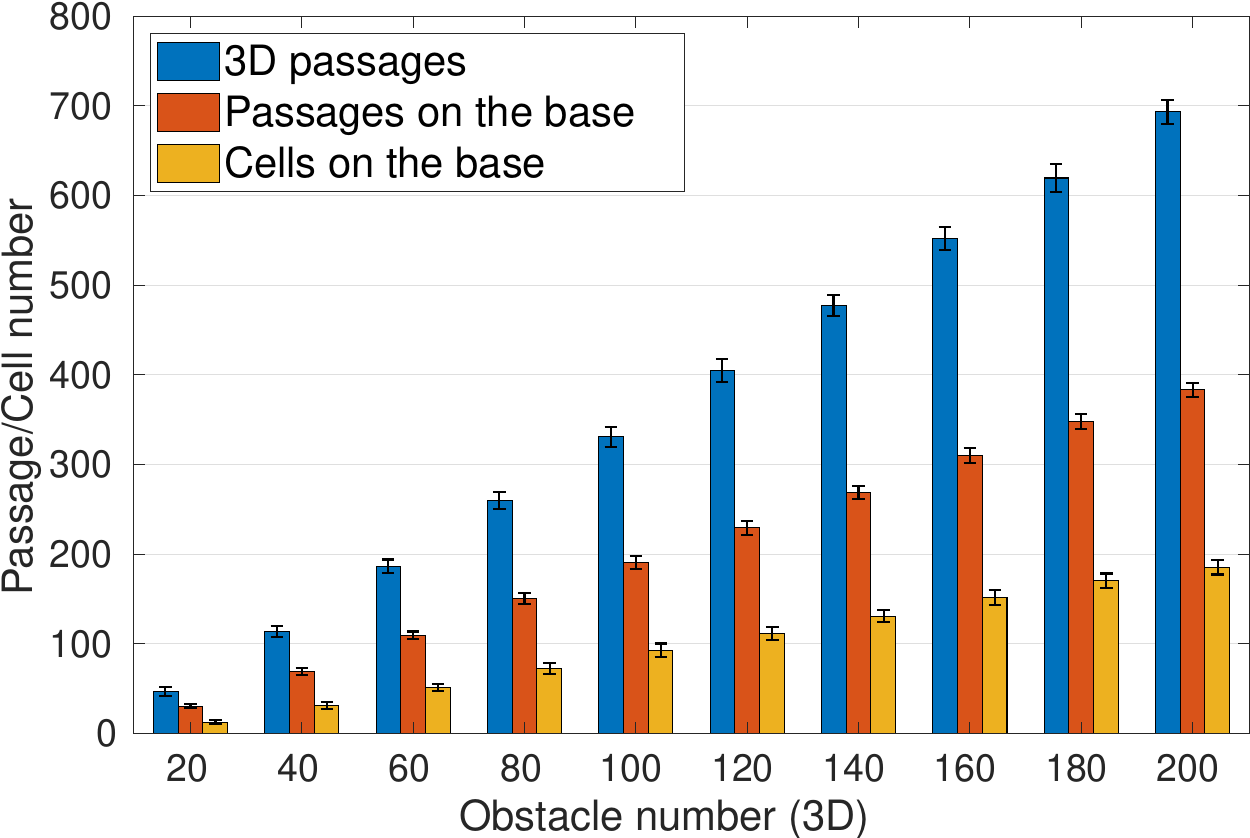}
    \label{Fig: Passage And Cell Detection Data 3D}
    }
    \subfigure[]{
    \includegraphics[width= 0.315 \columnwidth]{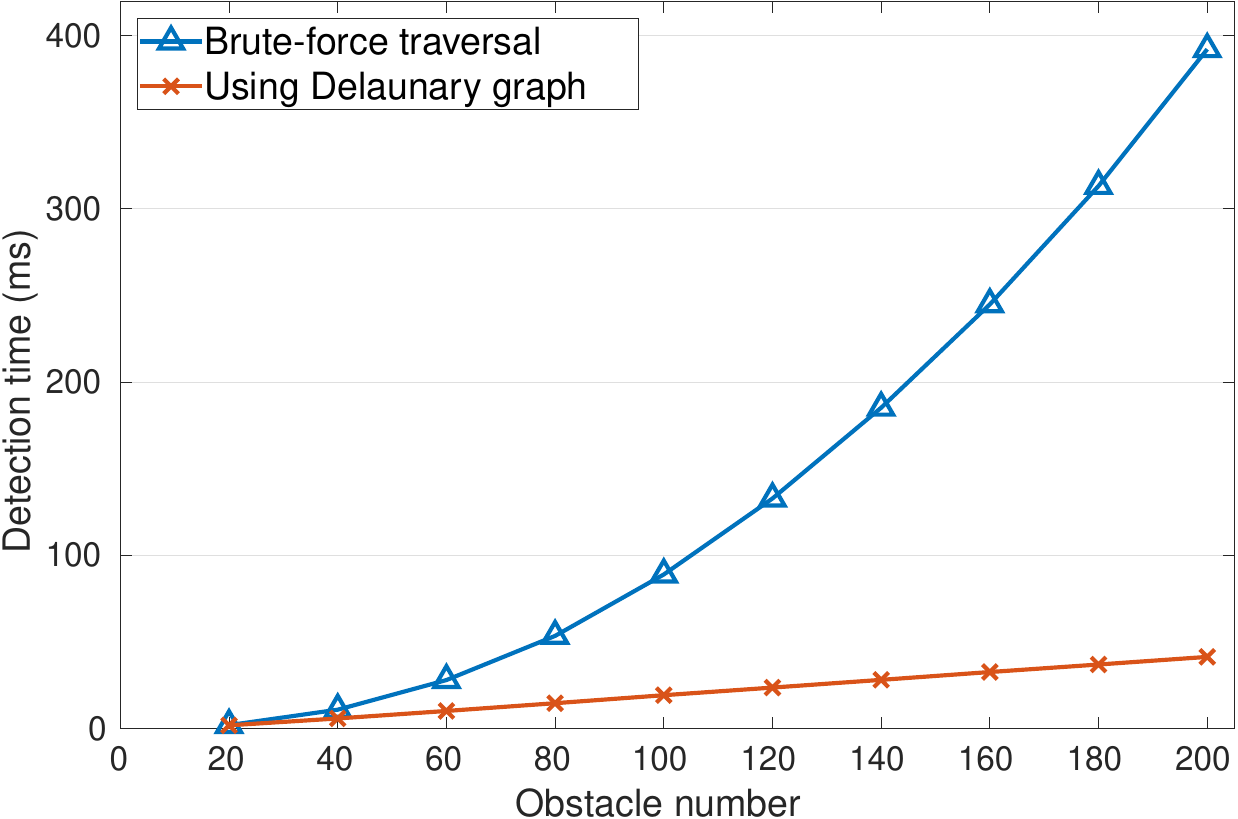}
    \label{Fig: Passage Detection Varying Obstacle Sizes}
    }
    \subfigure[]{
    \includegraphics[width= 0.315 \columnwidth]{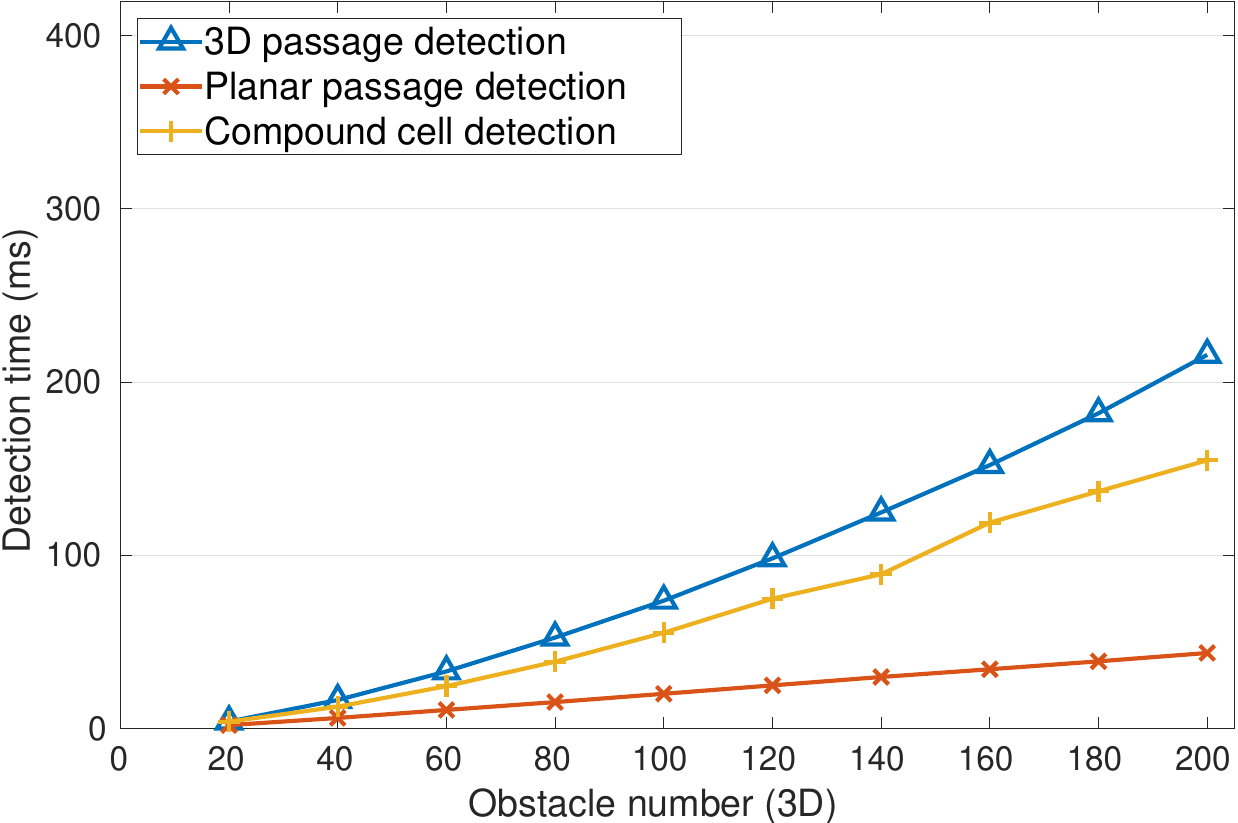}
    \label{Fig: Passage Detection Time Varying 3D}
    }    
    \endminipage  
    \hfill 
    \caption{Statistics in passage and cell detection. (a), (b) 2D passage and cell numbers under varying obstacle number $m$. (c) Passage numbers when $m = 40$ and obstacles vary in size.
    (d) Passage and cell numbers under varying $m$ in 3D space. (e) Passage detection time based on Delaunay graphs and direct traversal check. (f) Passage and compound cell detection time in 3D space.}
    \label{Fig: Passage and Cell Detection Data}
\end{figure*}

\hyperref[Fig: Passage Detection Varying Obstacle Sizes]{Figure \ref{Fig: Passage Detection Varying Obstacle Sizes}} and \hyperref[Fig: Passage Detection Time Varying 3D]{\ref{Fig: Passage Detection Time Varying 3D}} detail the passage and Gabriel cell detection time. Passage detection time using Delaunay graphs and direct brute-force traversal is reported. 
How Delaunay graphs improve the detection efficiency and how the detection time varies with the obstacle number are of interest. Because of fast Delaunay graph construction and the following linear passage check routine within the graph, significant time is saved compared to directly traversing all obstacle pairs in \hyperref[Fig: Passage Detection Varying Obstacle Sizes]{Figure \ref{Fig: Passage Detection Varying Obstacle Sizes}}. This improvement becomes more evident when obstacles increase and up to an order of magnitude speedup is reached, e.g., when $m = 200$. The passage detection time rising trend in Delaunay graphs is much less drastic than the direct check method, which has cubic time complexity. Due to the dominance of linear complexity in $O(m\log m)$, linearity between detection time and obstacle number is obvious in Delaunay graphs.

Subsequent Gabriel cell detection only performs sorting and DFS of the detected passage graph once. It is thus fast and completes within less than \SI{0.1}{\ms} in experiments. In other words, Gabriel cell detection consumes negligible computations compared to passage detection. Passage and cell detection time in 3D space is displayed in \hyperref[Fig: Passage Detection Time Varying 3D]{Figure \ref{Fig: Passage Detection Time Varying 3D}}. The corresponding planar passage detection time is listed as a benchmark. The results show that detecting spatial passages is more time-consuming than planar passages because all intermediate Delaunay graph edges are examined. Although the computational complexity remains, the detection time rises more drastically with obstacle number. Compound cell detection in 3D space is faster than 3D passage detection, but costs much longer time than 2D cells. This time increase originates from processing midair passages. For each midair passage, the base cells its projection intersects with are computed, which accounts for most of the compound cell detection time.

\subsection{PTOPP results}
After passage and cell detection, all PTOPP categories are implemented and thoroughly tested. This part reports PTOPP results and particularly focuses on accessible free space optimization results in different path costs. As a new path planning paradigm leveraging passage properties, PTOPP depends on new functionalities that have not been realized before. Despite many efficient planner implementations available such as the open motion planning library \citep{sucan2012open}, there are no existing interfaces for passages to develop PTOPP algorithms readily. To address this, planners are implemented independently with all subroutines in {\CC}. PRM$^*$ and RRT$^*$ for PTOPP are tested as representatives. Primitives are developed as invokable components and problem categories are easily configurable via specifying cost types. Though the chief objective of PTOPP is to enable large accessible free space along paths, clearance is not embedded explicitly in the planning stage. But it is easily achievable by postprocessing. 
For instance, raw paths may be locally translated to passage centers to enlarge clearance efficiently \citep{huang2023deformable, huang2024robot}. For consistency, path clearance is not explicitly constrained in the planning stage of PTOPP.

\subsubsection{PTOPP results by categories}
\begin{figure*}[t]
    \minipage{2 \columnwidth}
    \centering
    \subfigure[]{
    \includegraphics[width= 0.315 \columnwidth]{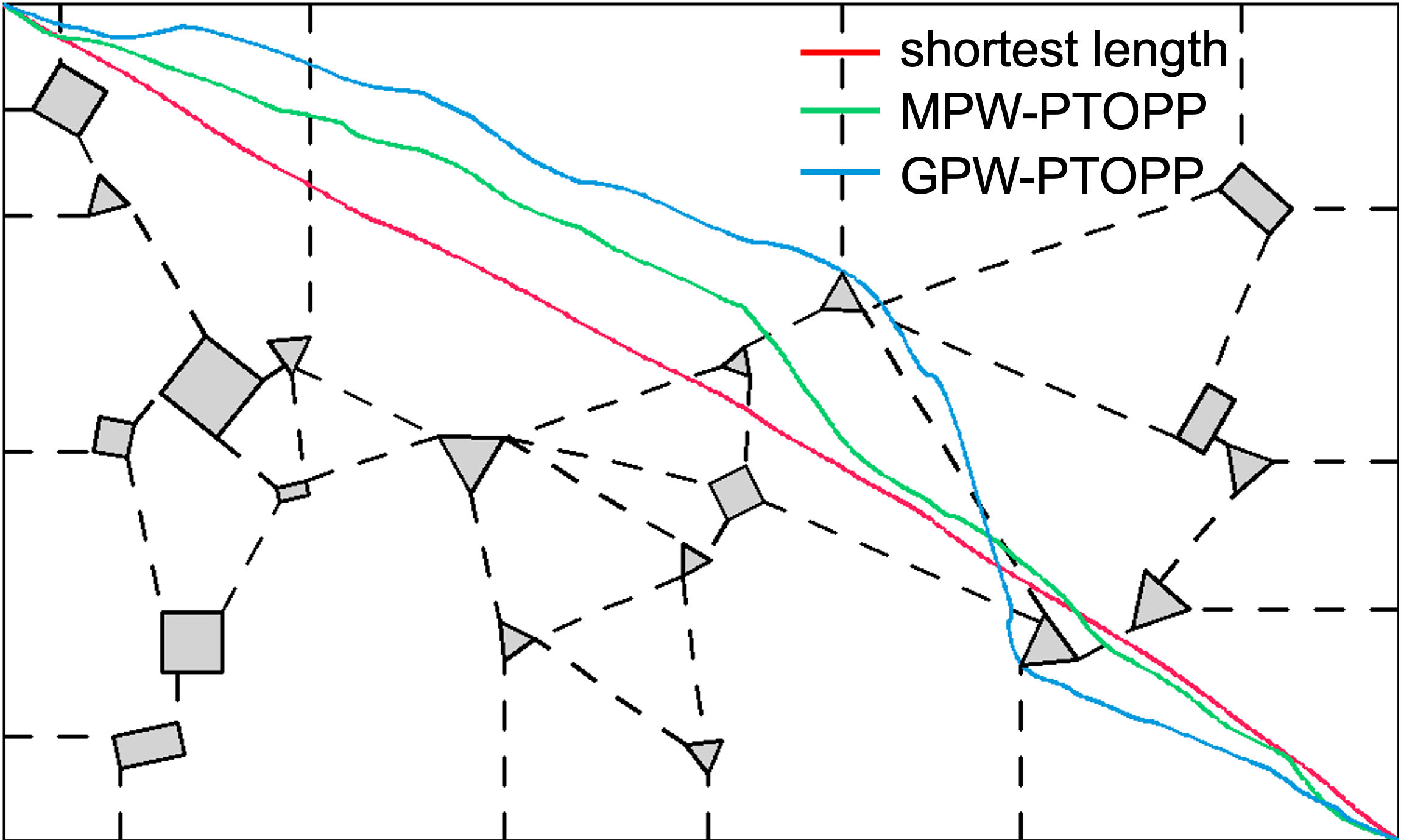}
    \label{Fig: Plannar Planning Examples Obstacle 20}
    }
    \subfigure[]{
    \includegraphics[width= 0.315 \columnwidth]{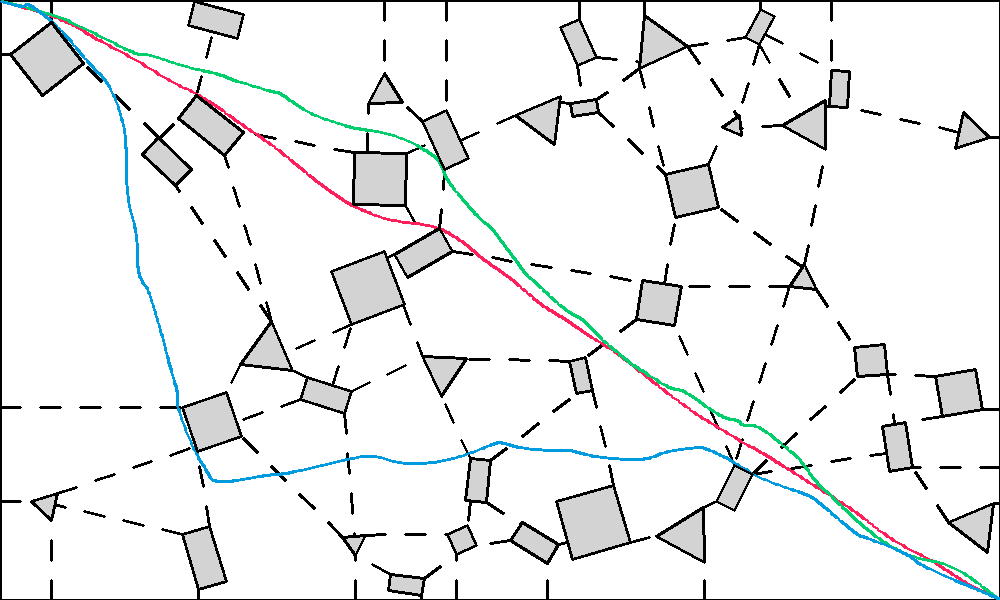} 
    \label{Fig: Plannar Planning Examples Obstacle 40}
    }
    \subfigure[]{
    \includegraphics[width= 0.315 \columnwidth]{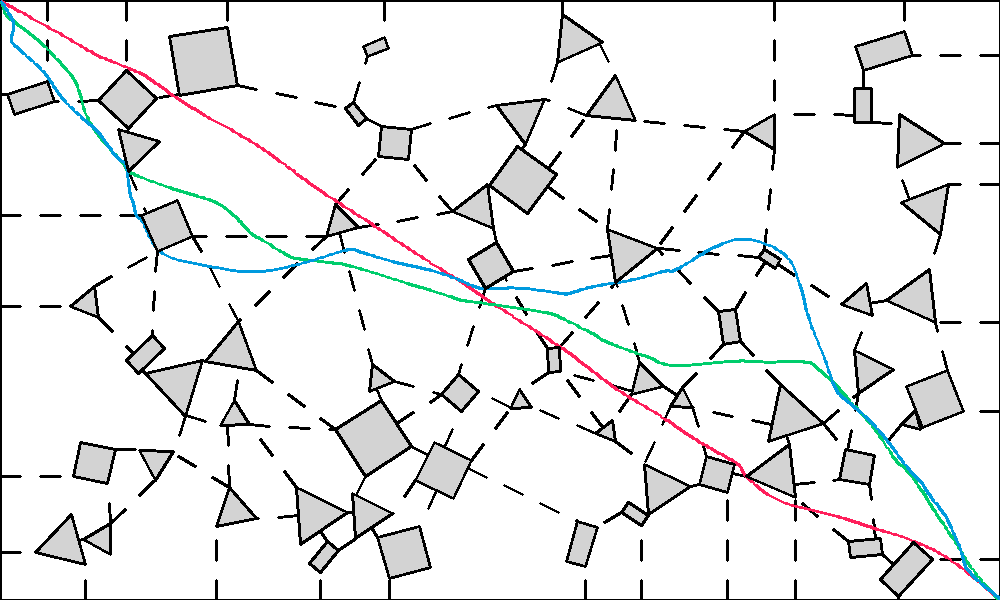}
    \label{Fig: Plannar Planning Examples Obstacle 60}
    }
    \subfigure[]{
    \includegraphics[width= 0.315 \columnwidth]{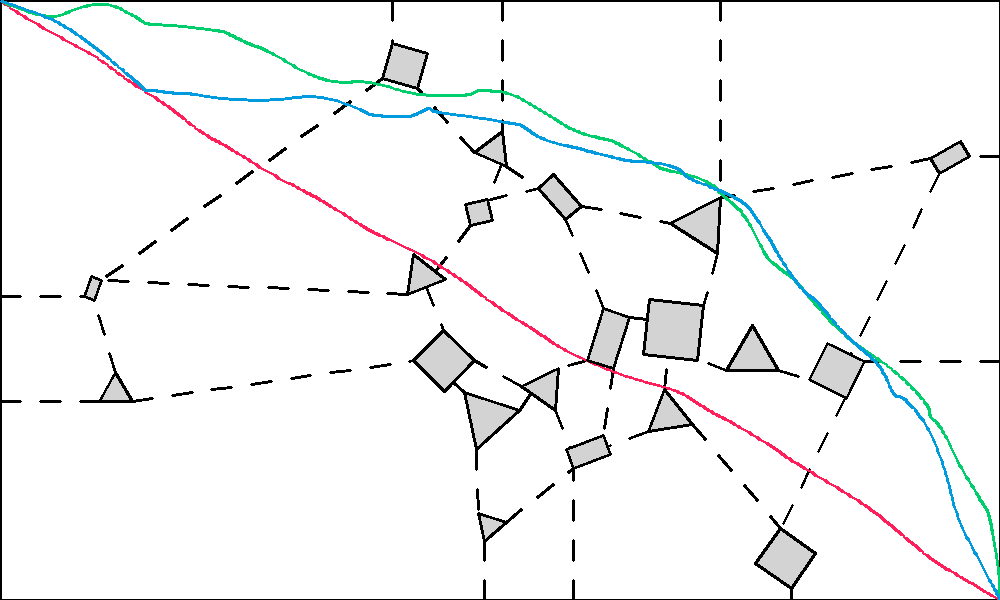}
    \label{Fig: Plannar Planning Examples Gaussian Obstacle 20}
    }
    \subfigure[]{
    \includegraphics[width= 0.315 \columnwidth]{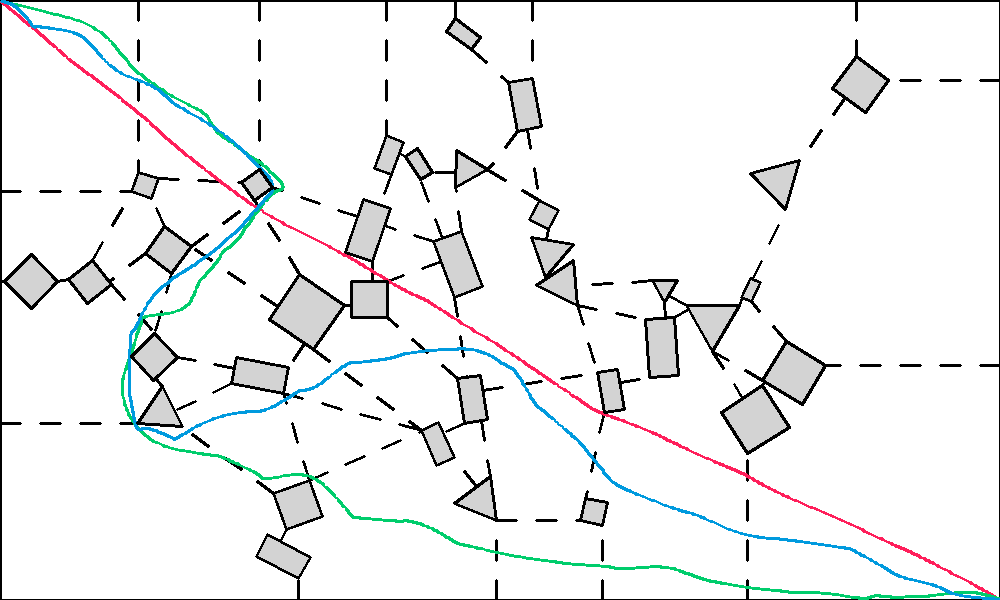}
    \label{Fig: Plannar Planning Examples Gaussian Obstacle 40}
    }
    \subfigure[]{
    \includegraphics[width= 0.315 \columnwidth]{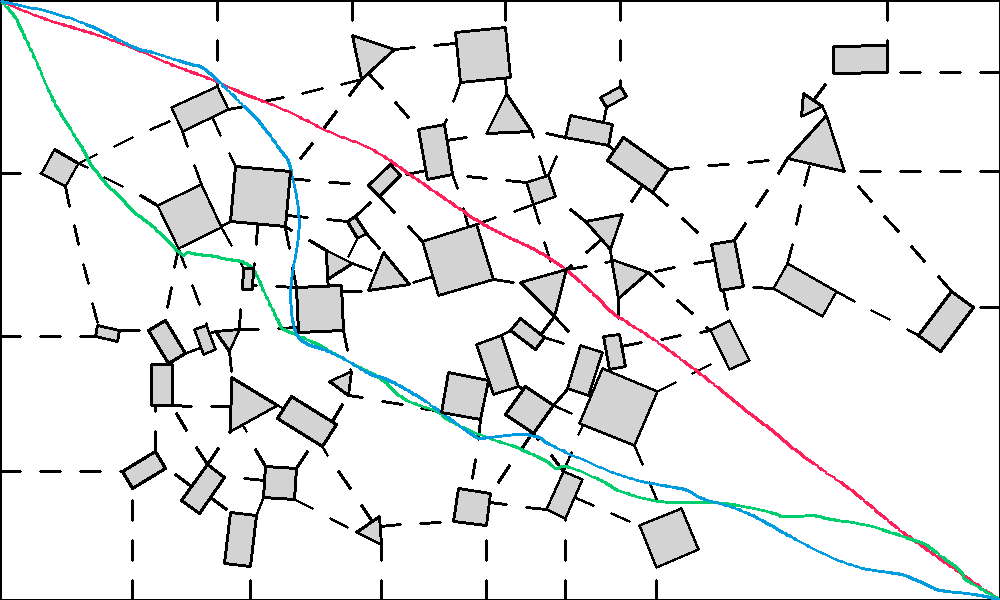}
    \label{Fig: Plannar Planning Examples Gaussian Obstacle 60}
    }    
    \endminipage \hfill    
    \caption{PTOPP examples with different cost and environment setups. (a)-(c) Obstacles are uniformly distributed and $m =$ 20, 40, and 60, respectively. (d)-(f) The same numbers of obstacles respect Gaussian distributions.}
    \label{Fig: Plannar Planning Examples}
\end{figure*}

\hyperref[Fig: Plannar Planning Examples]{Figure \ref{Fig: Plannar Planning Examples}} presents path planning examples in 2D maps. RRT$^*$ for both MPW-PTOPP and GPW-PTOPP is conducted, and shortest path planning acts as the benchmark. The start $\mathbf{x}_0$ and goal $\mathbf{x}_g$ are placed at the top left and bottom right corners, respectively. 3-GPW-PTOPP is implemented with a weight item common ratio of $100$. Varying-size obstacles respect uniform or Gaussian distributions, and environment boundaries are processed as obstacles. Each plan runs until the valid sample number $| V | = 5$k, which is sufficient for optimal path convergence in tests.  
Planned paths commonly differ in traversed passages in different planners. PTOPP selects passages to optimize accessible free space characterized by associated costs. In planar maps, such differences essentially infer different path homotopy classes.
\begin{figure*}[t]
    \minipage{2 \columnwidth}
    \centering
    \subfigure[]{
    \includegraphics[width= 0.315 \columnwidth]{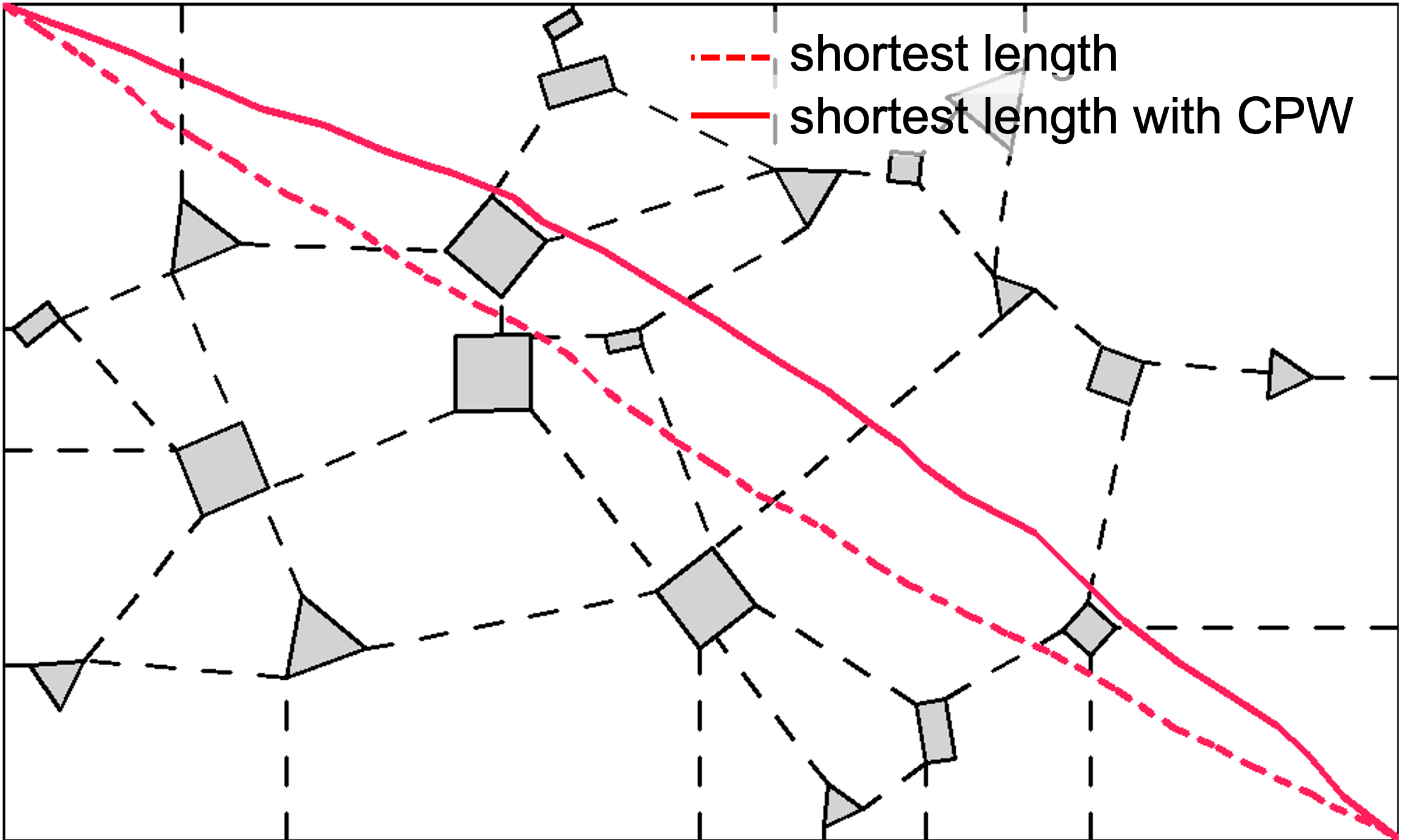}
    \label{Fig: Plannar Planning Examples Constrained a}
    }
    \subfigure[]{
    \includegraphics[width= 0.315 \columnwidth]{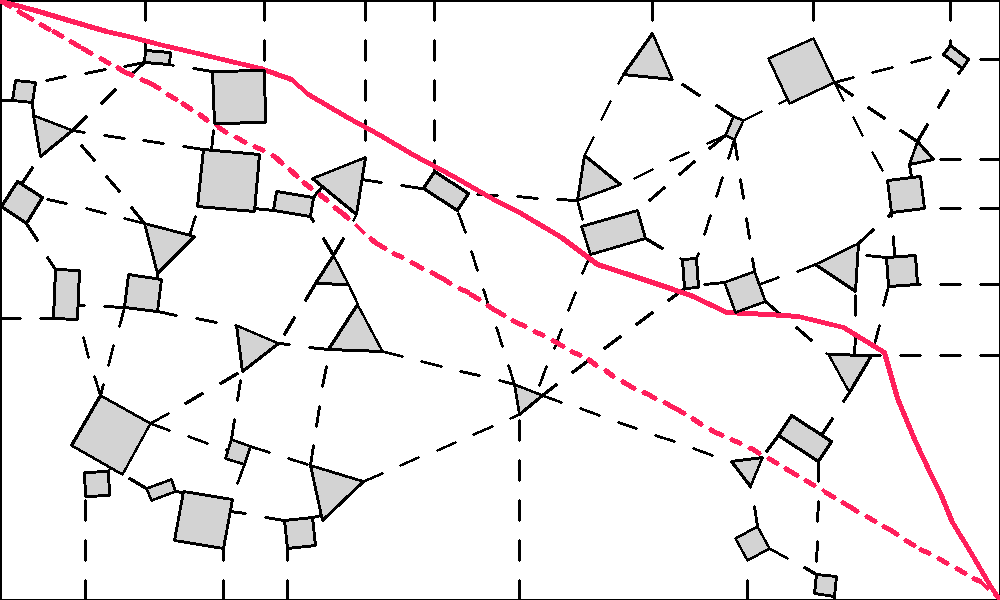}
    \label{Fig: Plannar Planning Examples Constrained b}
    }
    \subfigure[]{
    \includegraphics[width= 0.315 \columnwidth]{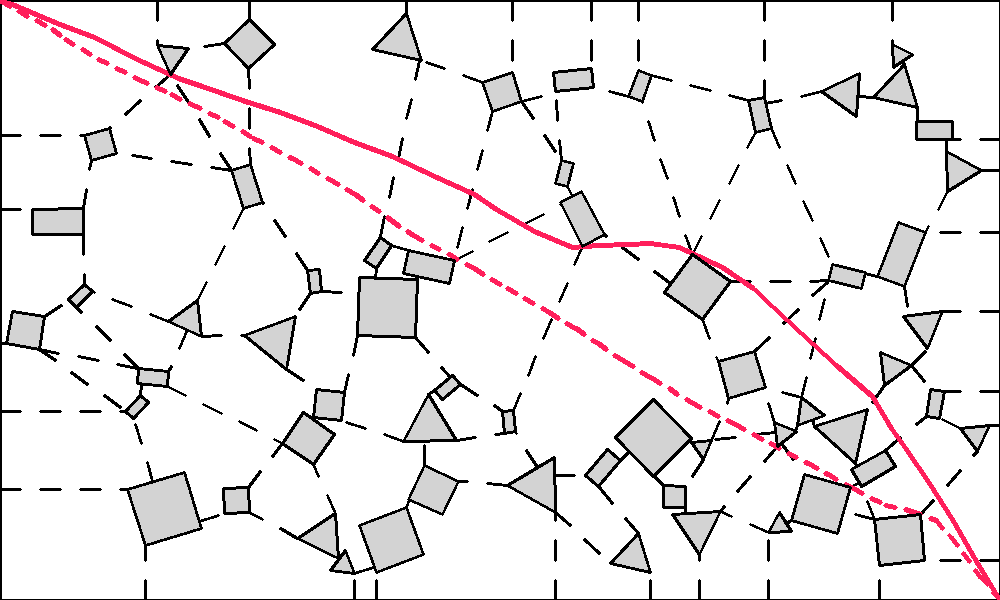}
    \label{Fig: Plannar Planning Examples Constrained c}
    }
    \subfigure[]{
    \includegraphics[width= 0.315 \columnwidth]{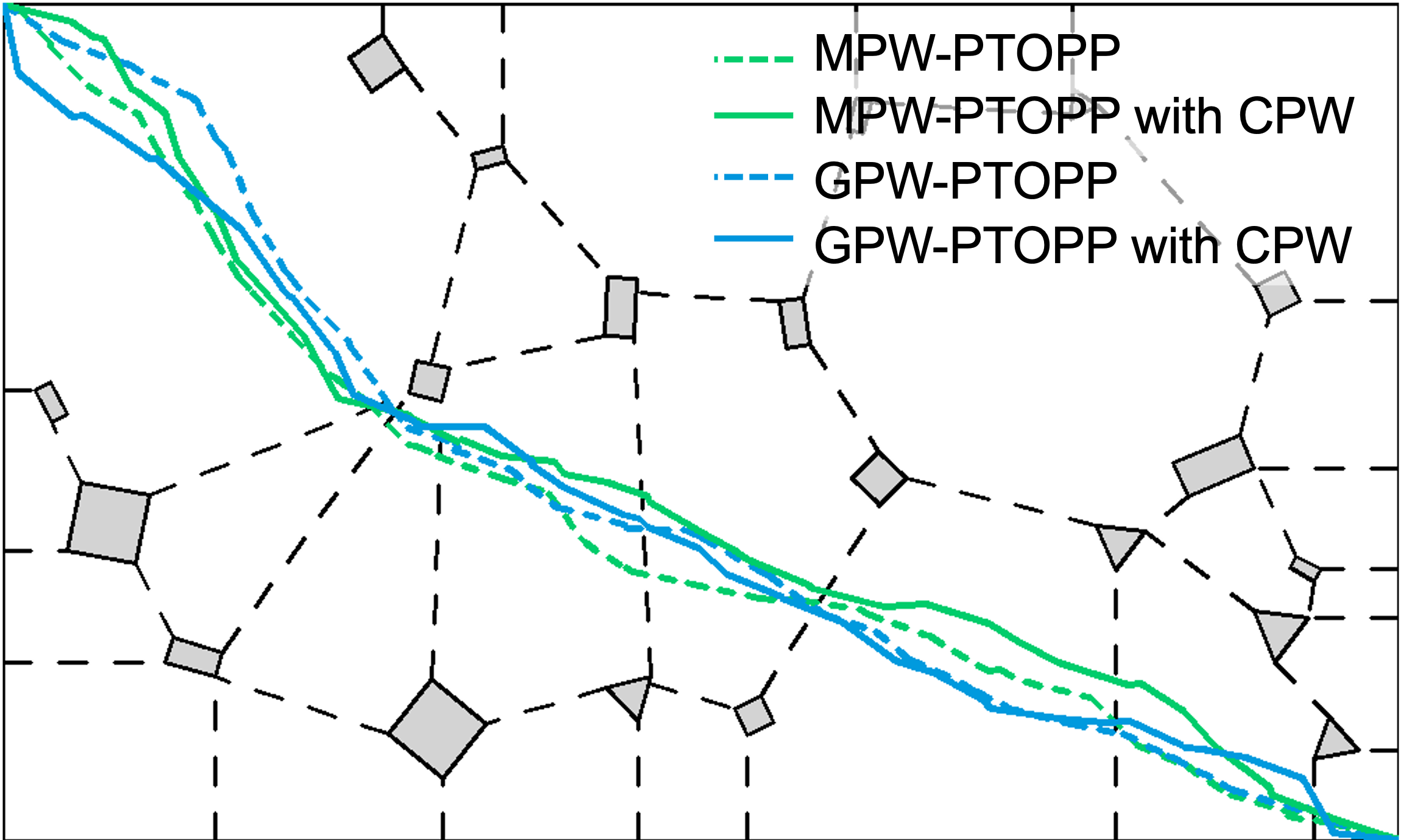}
    \label{Fig: Plannar Planning Examples Constrained d}
    }
    \subfigure[]{
    \includegraphics[width= 0.315 \columnwidth]{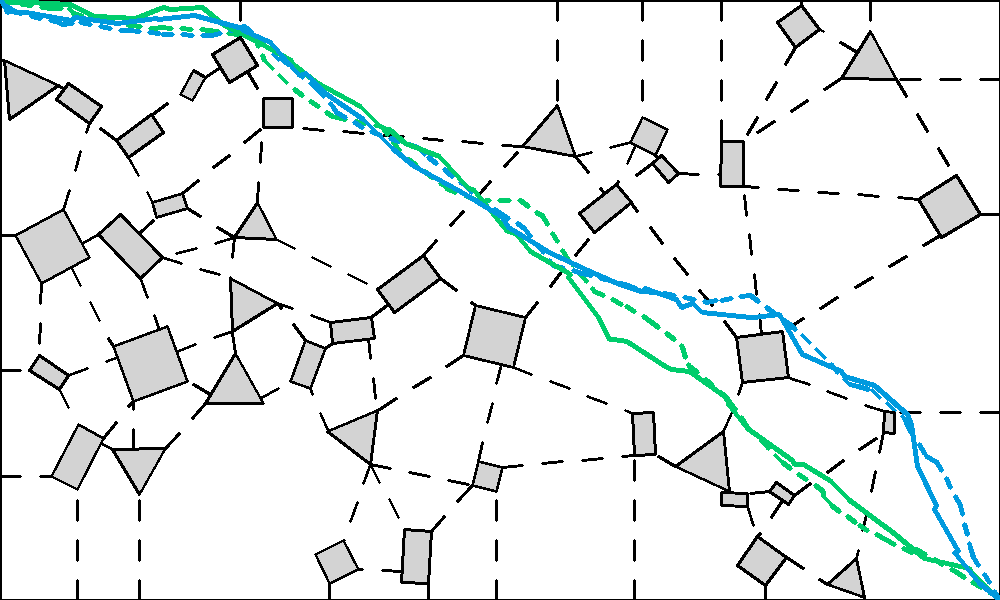}
    \label{Fig: Plannar Planning Examples Constrained e}
    }
    \subfigure[]{
    \includegraphics[width= 0.315 \columnwidth]{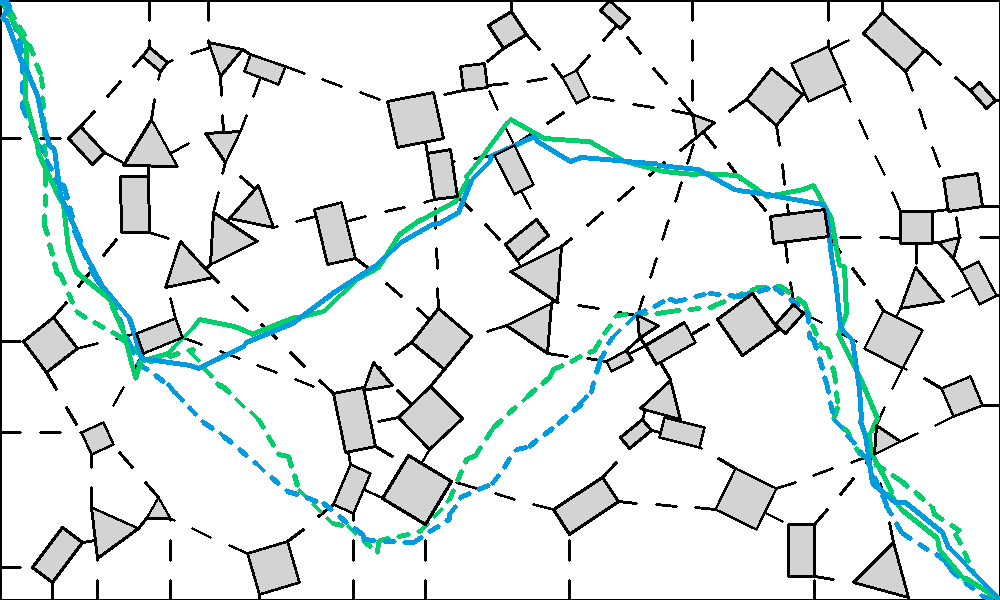}
    \label{Fig: Plannar Planning Examples Constrained f}
    }    
    \endminipage \hfill    
    \caption{CPW-PTOPP examples imposed on different types of planners. Obstacles are uniformly distributed. (a)-(c) CPW-PTOPP is integrated into shortest path planning. (d)-(f) CPW-PTOPP is integrated into MPW-PTOPP and GPW-PTOPP.}
    \label{Fig: Plannar Planning Examples Constrained}
\end{figure*}

The shortest paths (red paths) typically pass through quite narrow passages to achieve the minimum lengths, leading to limited accessible free space alongside. In contrast, paths in PTOPP traverse particular passages to admit large free space. Although both improve the accessibility of free space, MPW-PTOPP (green paths) and GPW-PTOPP (blue paths) behave differently. In general, GPW-PTOPP is more greedy in choosing wide passages. For instance, in \hyperref[Fig: Plannar Planning Examples Obstacle 20]{Figure \ref{Fig: Plannar Planning Examples Obstacle 20}} and \hyperref[Fig: Plannar Planning Examples Obstacle 40]{\ref{Fig: Plannar Planning Examples Obstacle 40}}, narrow passages enclose the start position and thus are inevitable. After passing them at the beginning, MPW-PTOPP degrades in the sense that for passages encountered, only those narrower than the previously passed ones are actively avoided. In consequence, narrow passages are traversed in MPW-PTOPP when they could have been avoided. This problem is alleviated in GPW-PTOPP by optimizing multiple confined passages more than the narrowest one. GPW-PTOPP paths in \hyperref[Fig: Plannar Planning Examples Obstacle 20]{Figure \ref{Fig: Plannar Planning Examples Obstacle 20}} and \hyperref[Fig: Plannar Planning Examples Obstacle 40]{\ref{Fig: Plannar Planning Examples Obstacle 40}} select routes enabling significantly larger free spaces alongside. The sorted passage width vector $\mathbf{p}_\sigma$ in GPW-PTOPP is guaranteed to be lexically no smaller than that in MPW-PTOPP. In many cases, however, two planners return paths traversing identical passages, namely homotopic paths in 2D space, like in \hyperref[Fig: Plannar Planning Examples Gaussian Obstacle 20]{Figure \ref{Fig: Plannar Planning Examples Gaussian Obstacle 20}}. 
\begin{figure*}[t]
    \minipage{2 \columnwidth}
    \centering
    \subfigure[]{
    \includegraphics[width= 0.315 \columnwidth]{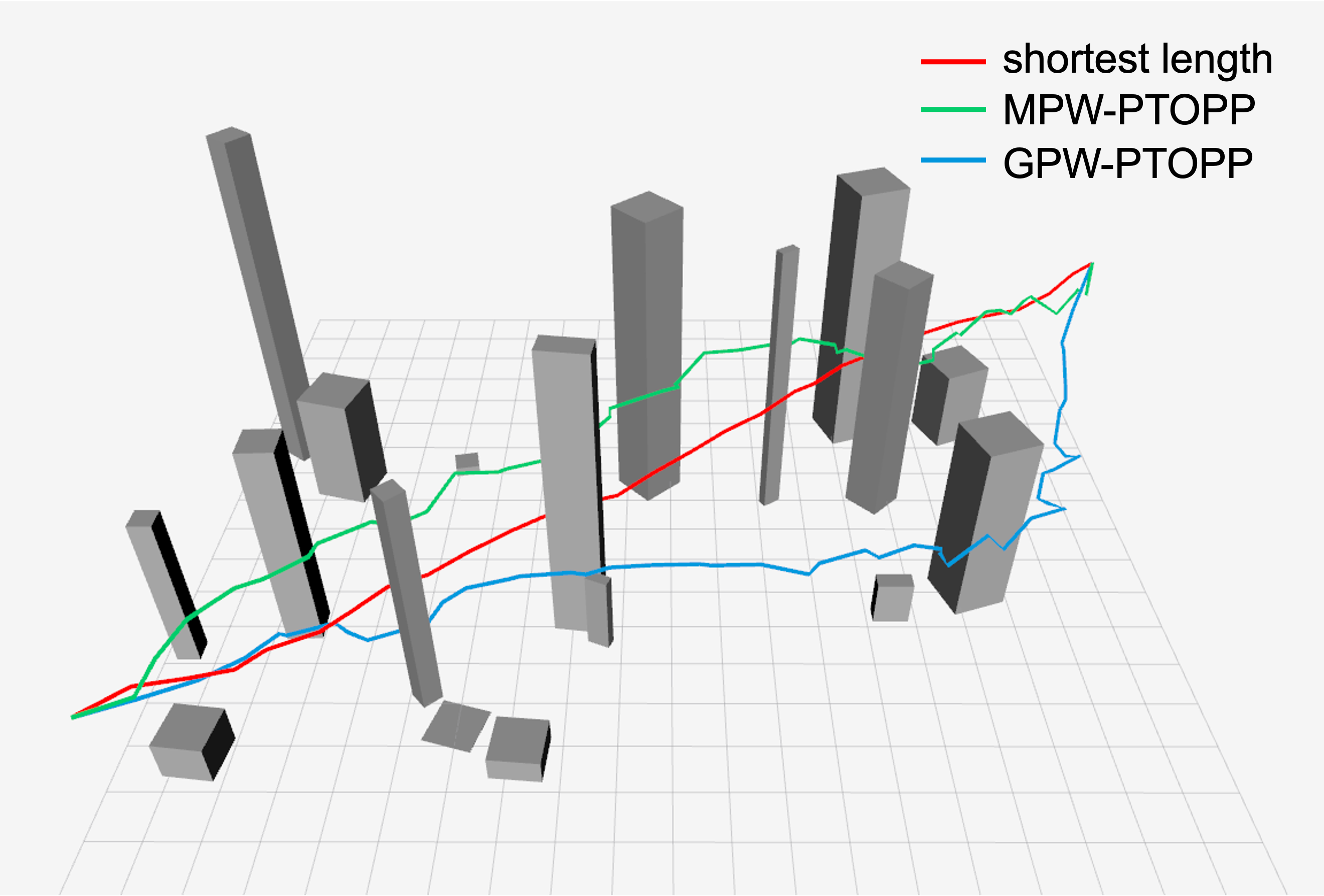}
    \label{Fig: 3D Planning Examples a}
    }
    \subfigure[]{
    \includegraphics[width= 0.315 \columnwidth]{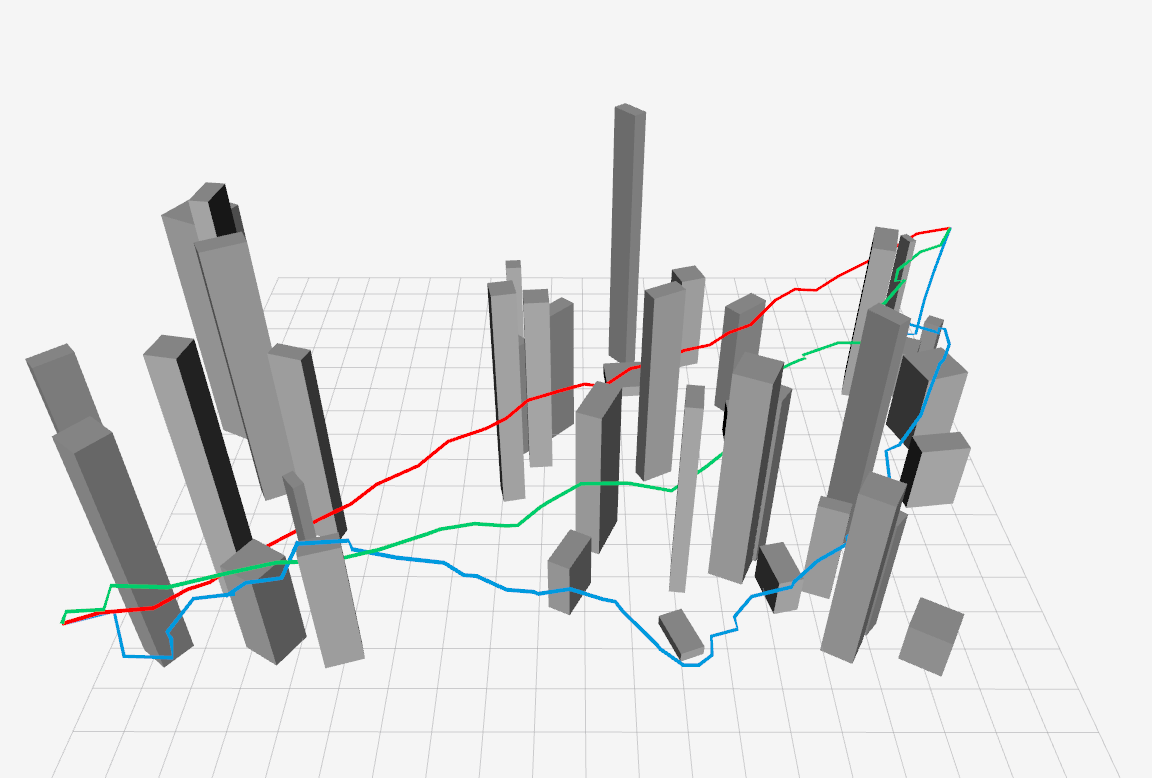}
    \label{Fig: 3D Planning Examples b}
    }
    \subfigure[]{
    \includegraphics[width= 0.315 \columnwidth]{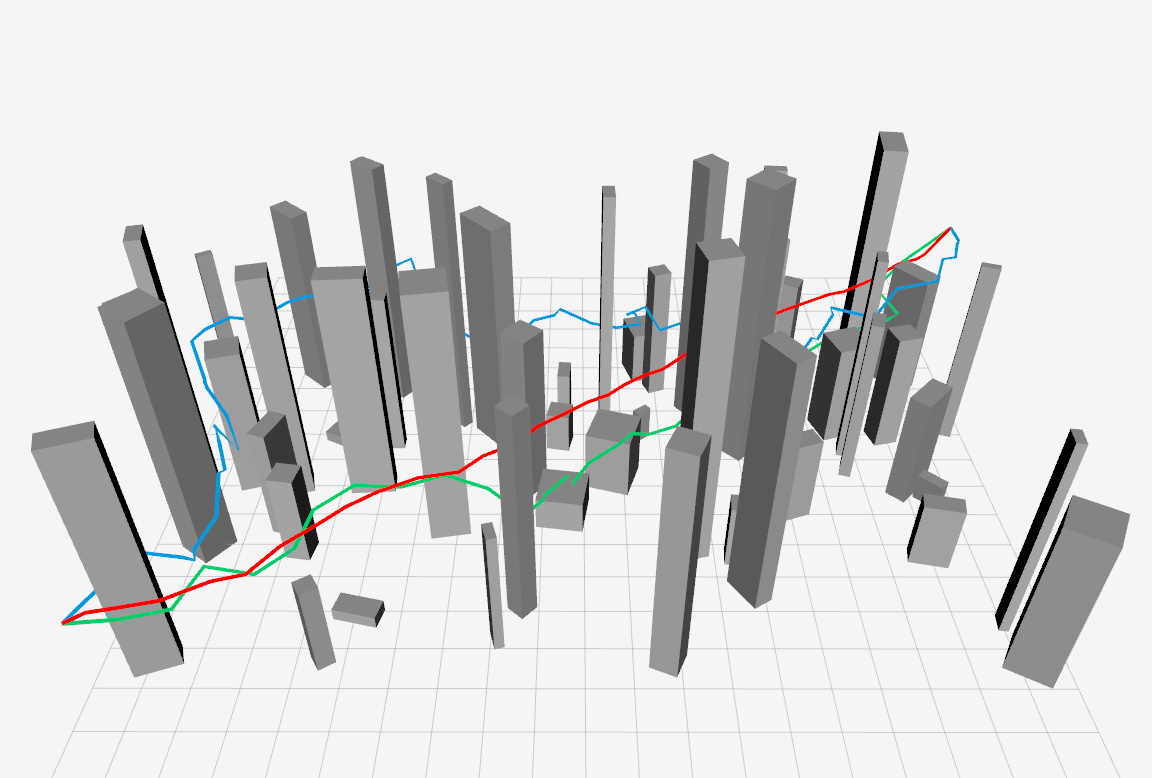}
    \label{Fig: 3D Planning Examples c}
    }   
        \subfigure[]{
    \includegraphics[width= 0.315 \columnwidth]{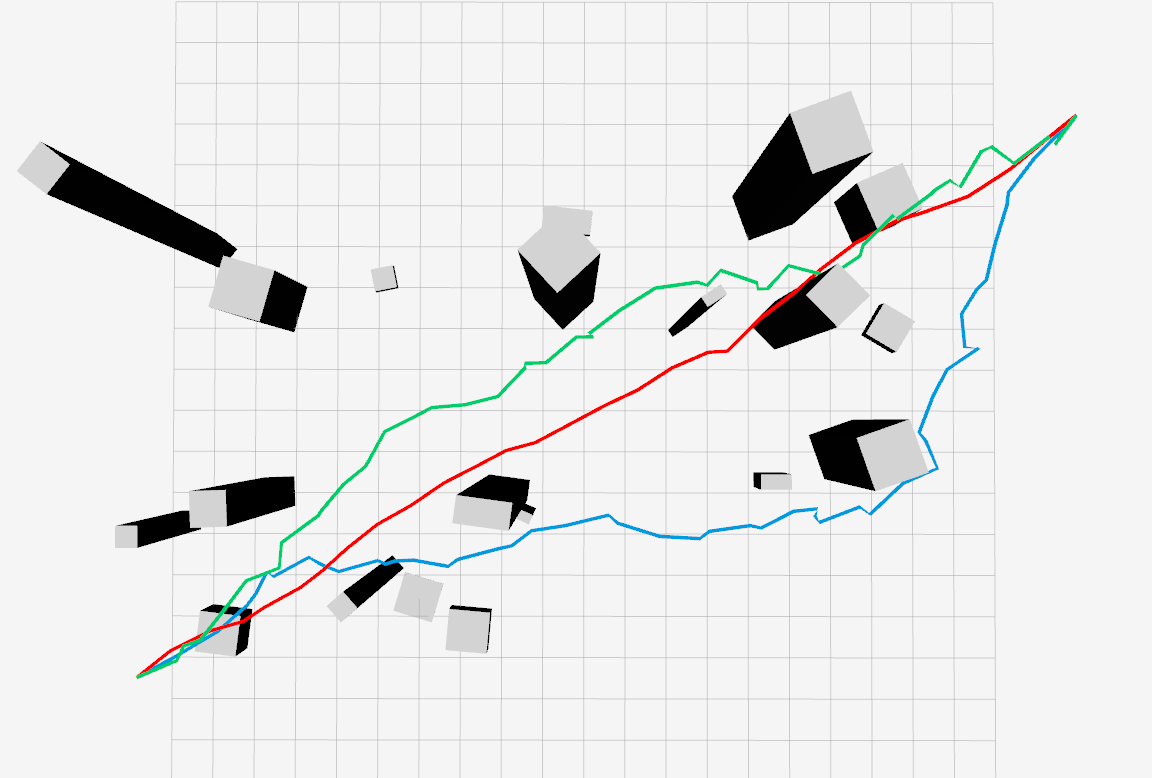}
    \label{Fig: 3D Planning Examples d}
    }
    \subfigure[]{
    \includegraphics[width= 0.315 \columnwidth]{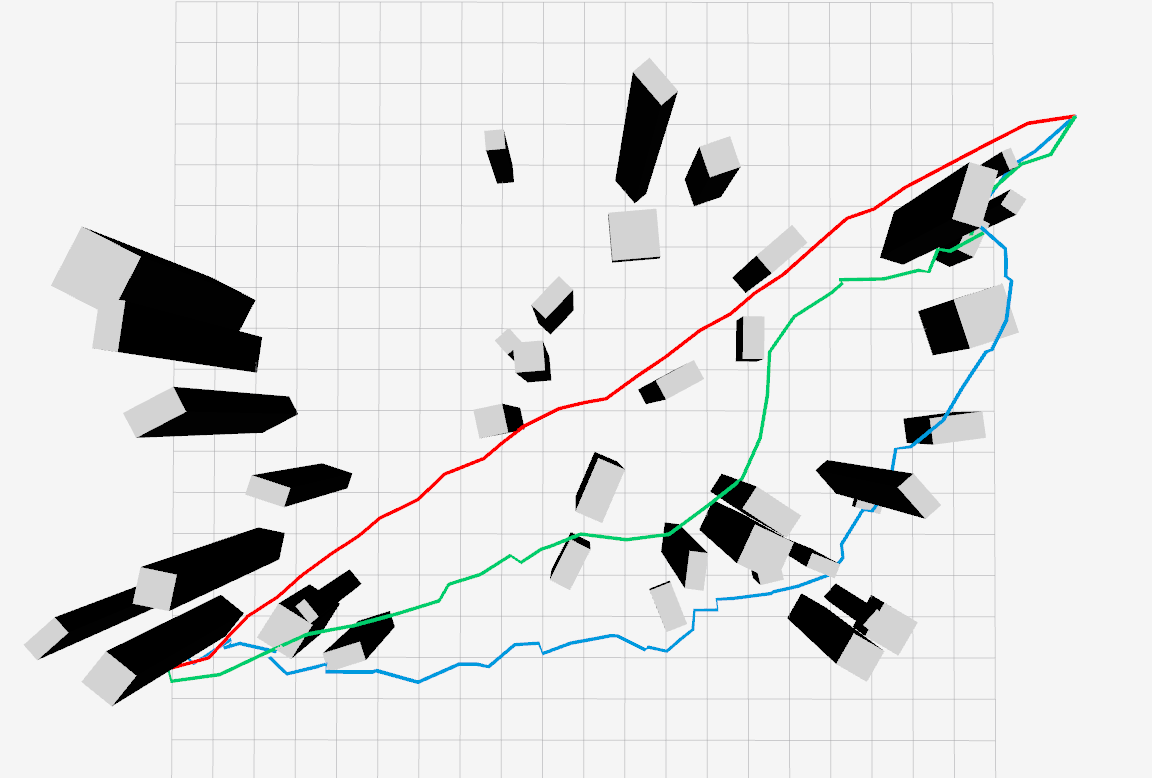}
    \label{Fig: 3D Planning Examples e}
    }
    \subfigure[]{
    \includegraphics[width= 0.315 \columnwidth]{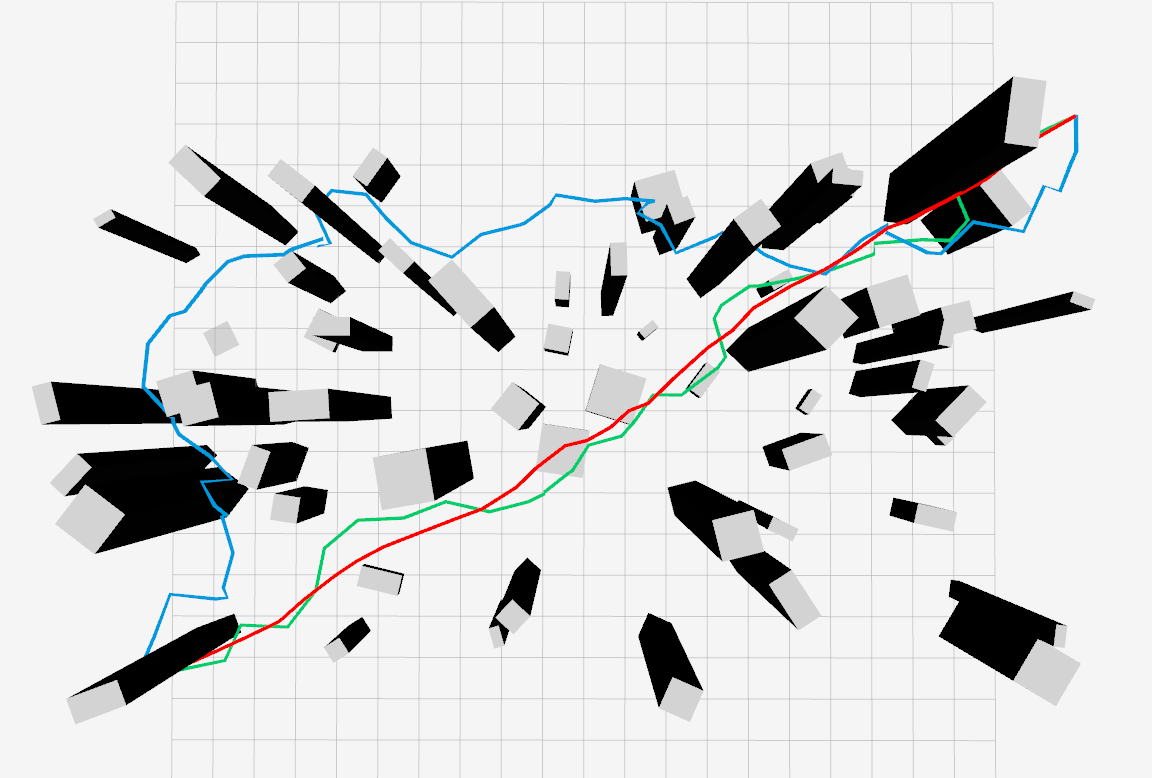}
    \label{Fig: 3D Planning Examples f}
    }  
    \endminipage \hfill    
    \caption{Examples of PTOPP results with different path costs in 3D space. The environment boundaries are not depicted. Obstacles are uniformly distributed. $m =$ 20, 40, and 60, respectively. (a)-(c) Side views. (d)-(f) Corresponding top views.}
    \label{Fig: 3D Planning Examples}
\end{figure*}

Constrained passage width PTOPP is tested by assigning the minimum admissible passage width $\underline{\varepsilon}_p$, which represents a ubiquitous demand to accommodate robots. It is readily integrated into shortest path planning as in (\ref{Eqn: Passage Width Cost on Path Length}) or PTOPP via relabeling passage widths as in (\ref{Eqn: Passage Width Update Rule}) to penalize traversing unqualified narrow passages. \hyperref[Fig: Plannar Planning Examples Constrained a]{Figure \ref{Fig: Plannar Planning Examples Constrained a}} to \hyperref[Fig: Plannar Planning Examples Constrained c]{\ref{Fig: Plannar Planning Examples Constrained c}} present CPW-PTOPP results built on shortest path planning ($\underline{\varepsilon}_p = 60$). Dashed and solid paths are without and with the constrained passage width (CPW), respectively. CPW-PTOPP finds the shortest path avoiding passages narrower than $\underline{\varepsilon}_p$. If there is no feasible path with its minimum passage width $f_p(\sigma) > \underline{\varepsilon}_p$, CPW-PTOPP will pass the fewest passages narrower than $\underline{\varepsilon}_p$. 
\hyperref[Fig: Plannar Planning Examples Constrained d]{Figure \ref{Fig: Plannar Planning Examples Constrained d}} to \hyperref[Fig: Plannar Planning Examples Constrained f]{\ref{Fig: Plannar Planning Examples Constrained f}} display planned paths in MPW-PTOPP and GPW-PTOPP with CPW. $f_p(\sigma)$ is maximized in MPW-PTOPP and GPW-PTOPP even without an admissible threshold $\underline{\varepsilon}_p$. Therefore, homotopic paths are found in most situations since relabeling passage widths does not affect greedy selection of wide passages, e.g., \hyperref[Fig: Plannar Planning Examples Constrained d]{Figure \ref{Fig: Plannar Planning Examples Constrained d}} and \hyperref[Fig: Plannar Planning Examples Constrained e]{\ref{Fig: Plannar Planning Examples Constrained e}}. However, CPW-PTOPP is still beneficial. Specifically, if $f_p(\sigma) \leq \underline{\varepsilon}_p$ for all feasible $\sigma$, CPW-PTOPP finds the path traversing passages narrower than $\underline{\varepsilon}_p$ as few as possible. For instance, there is no feasible path with $f_{p}(\sigma) > \underline{\varepsilon}_p$ in \hyperref[Fig: Plannar Planning Examples Constrained f]{Figure \ref{Fig: Plannar Planning Examples Constrained f}}. Without CPW, the first three items in $\mathbf{p}_{\sigma}$ of the planned path are $[59.1, 59.3, 60.2]\T$. With CPW, items are $[42.3, 60.2, 63.9]\T$, in which only one entry is smaller than $\underline{\varepsilon}_p$ in spite of a smaller $f_p(\sigma)$. The overall path cost decreases due to fewer violations of the passage width constraint.

\subsubsection{PTOPP results in 3D space}
PTOPP algorithms are further tested in 3D space. In implementation, major modifications arise from spatial passage and compound Gabriel cell detection in preprocessing and the adjusted node positioning procedure in planning. Midair passages are added as virtual boundaries of planar cells that their projections intersect with. Node projections are positioned in the planar cells for fast passage traversal check. To obtain the passed passages of an edge, both real and virtual cell boundaries are examined. \hyperref[Fig: 3D Planning Examples]{Figure \ref{Fig: 3D Planning Examples}} exhibits PTOPP results in 3D space returned by RRT$^*$. $\mathbf{x}_0$ and $\mathbf{x}_g$ are set as the bottom right and top left corners respectively with height going from 100 to 200. The maximum valid sample number increases to $|V| = 15$k. PTOPP improves the free space accessibility in different ways. Obstacles of different heights restrict the free space. MPW-PTOPP and GPW-PTOPP enlarge accessible free space by selecting wider passages to travel. MPW-PTOPP only optimizes the narrowest passage width. GPW-PTOPP maximizes top \textit{k} narrowest passages usually at the expense of longer path lengths, making it more greedy in optimizing accessible free space.

\begin{figure}[t]
    \minipage{1 \columnwidth}
    \centering
     \includegraphics[width= 0.76 \columnwidth]{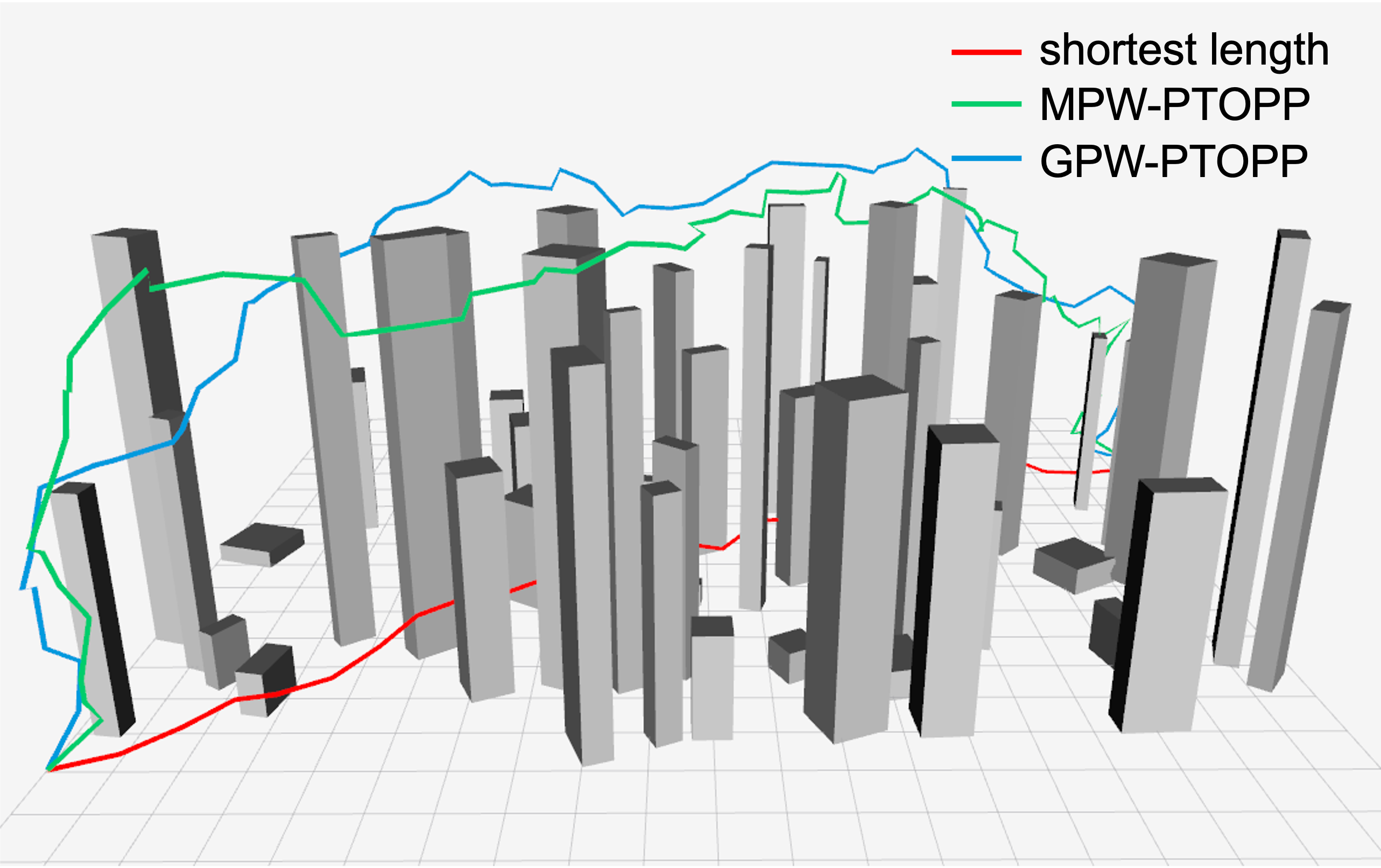}
     \endminipage \hfill
     \caption{$\mathbf{x}_0$ and $\mathbf{x}_g$ are diagonal ends of the ground. Paths in PTOPP first reach high and then go to the target.}
     \label{Fig: PTOPP Paths at High Heights}
\end{figure}

\begin{figure*}[t]
    \minipage{2 \columnwidth}
    \centering
    \subfigure[]{
     \includegraphics[width= 0.315 \columnwidth]{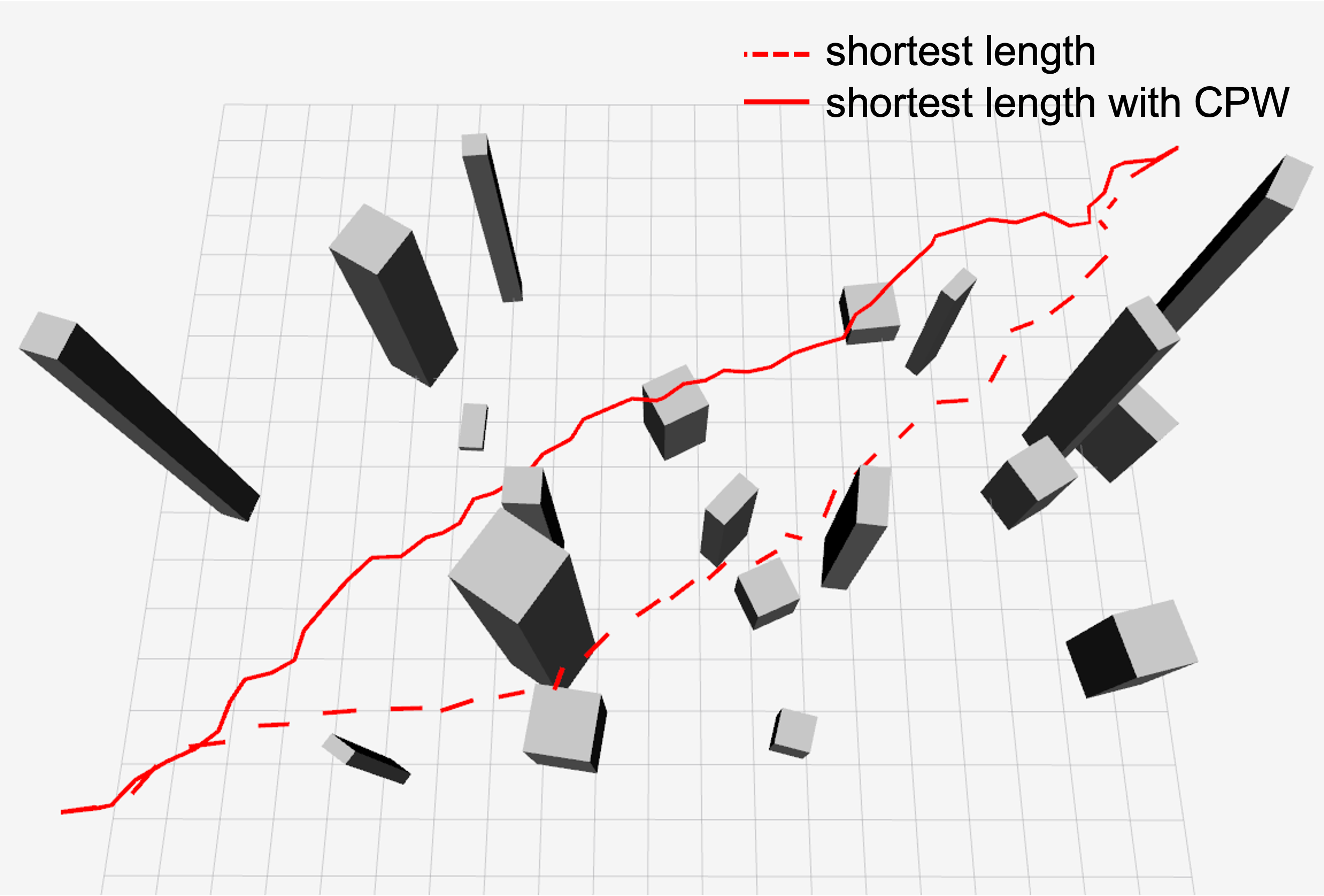}}    
    \subfigure[]{
     \includegraphics[width= 0.315 \columnwidth]{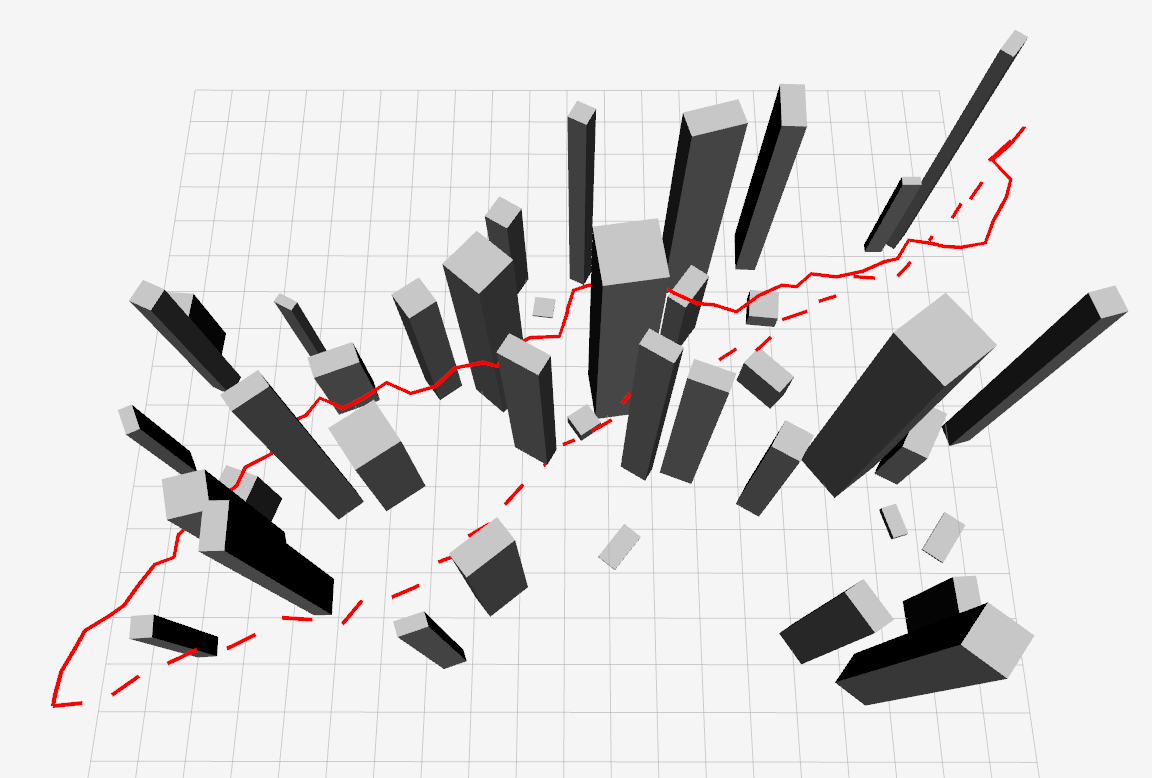}}    
    \subfigure[]{
     \includegraphics[width= 0.315 \columnwidth]{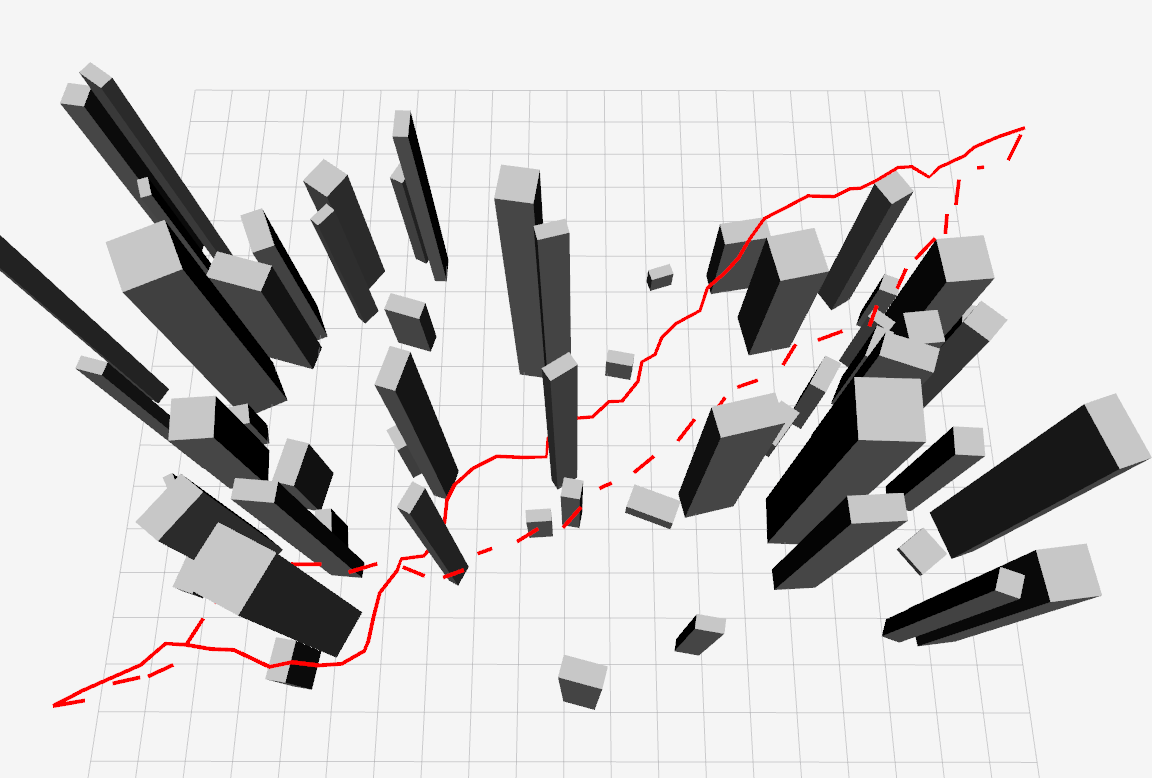}}   
    \subfigure[]{
     \includegraphics[width= 0.315 \columnwidth]{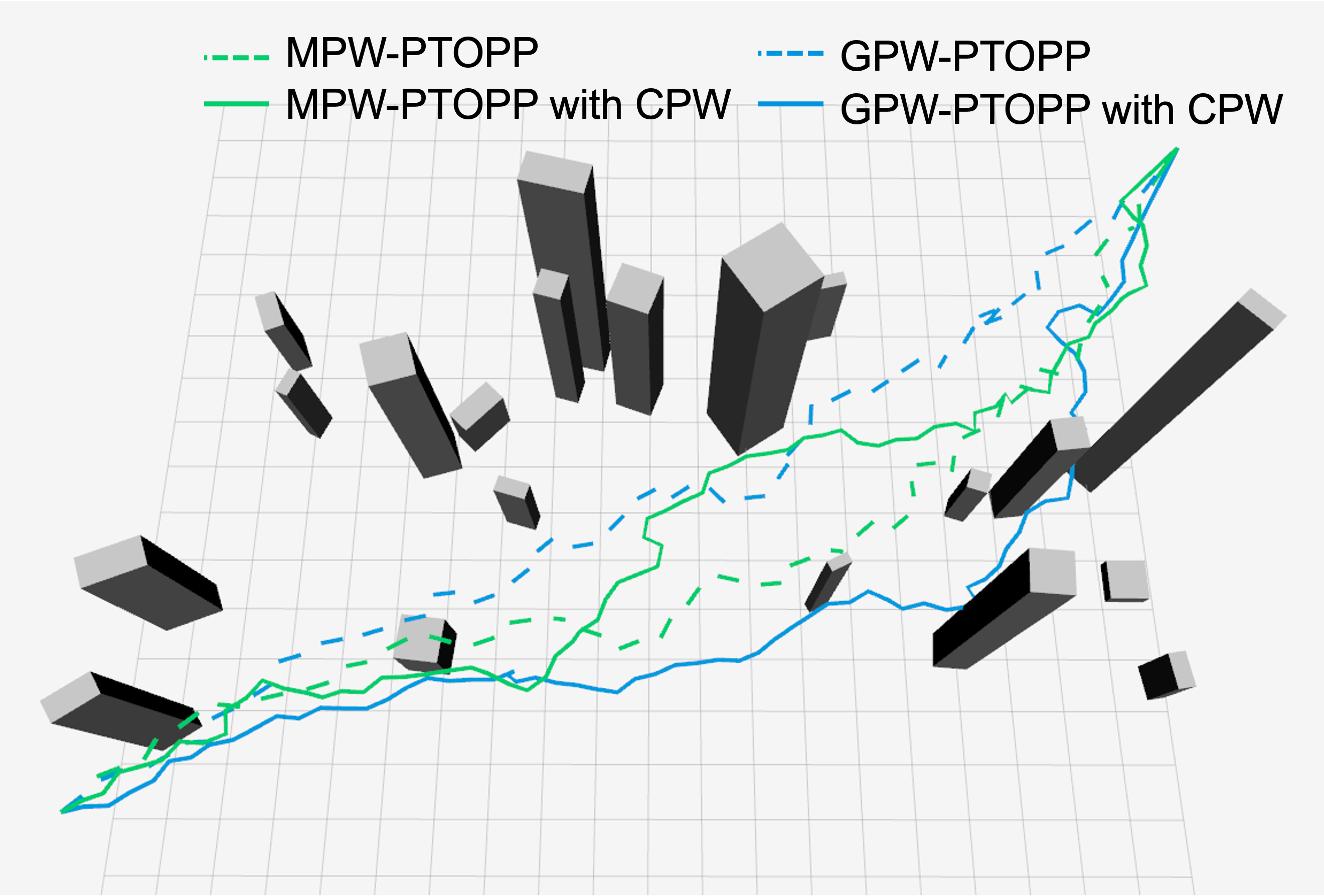}}    
    \subfigure[]{
     \includegraphics[width= 0.315 \columnwidth]{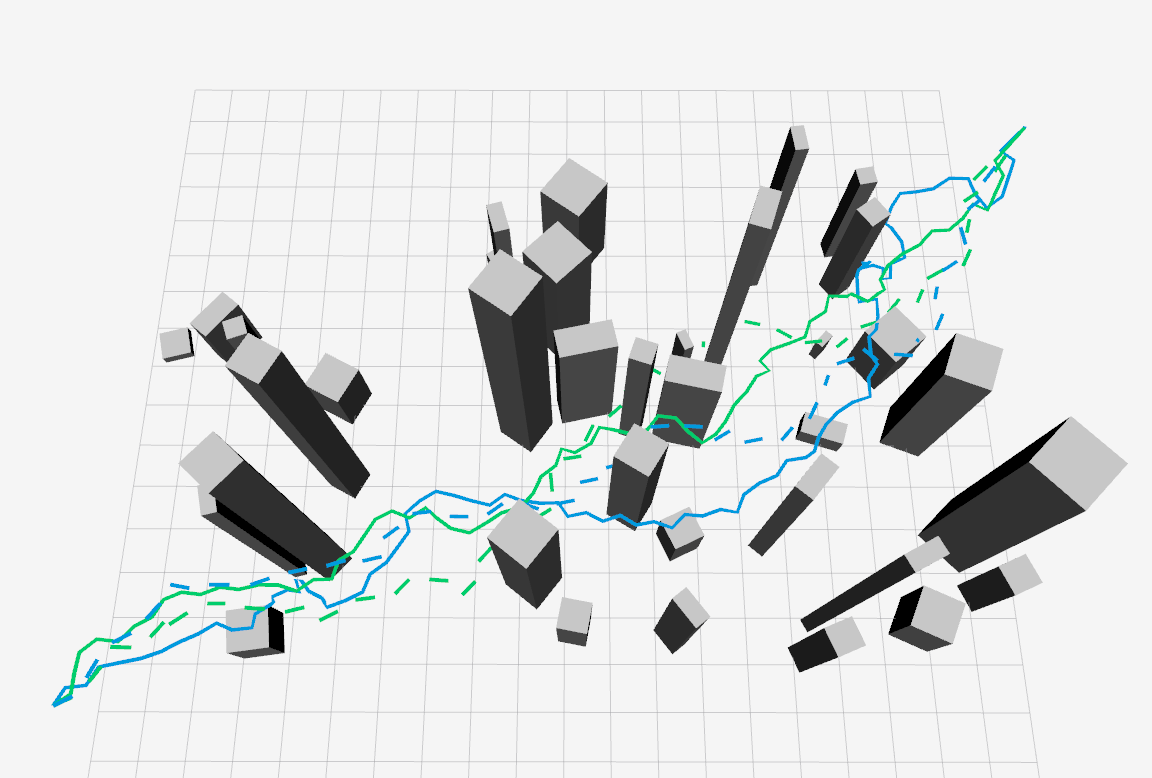}}    
    \subfigure[]{
     \includegraphics[width= 0.315 \columnwidth]{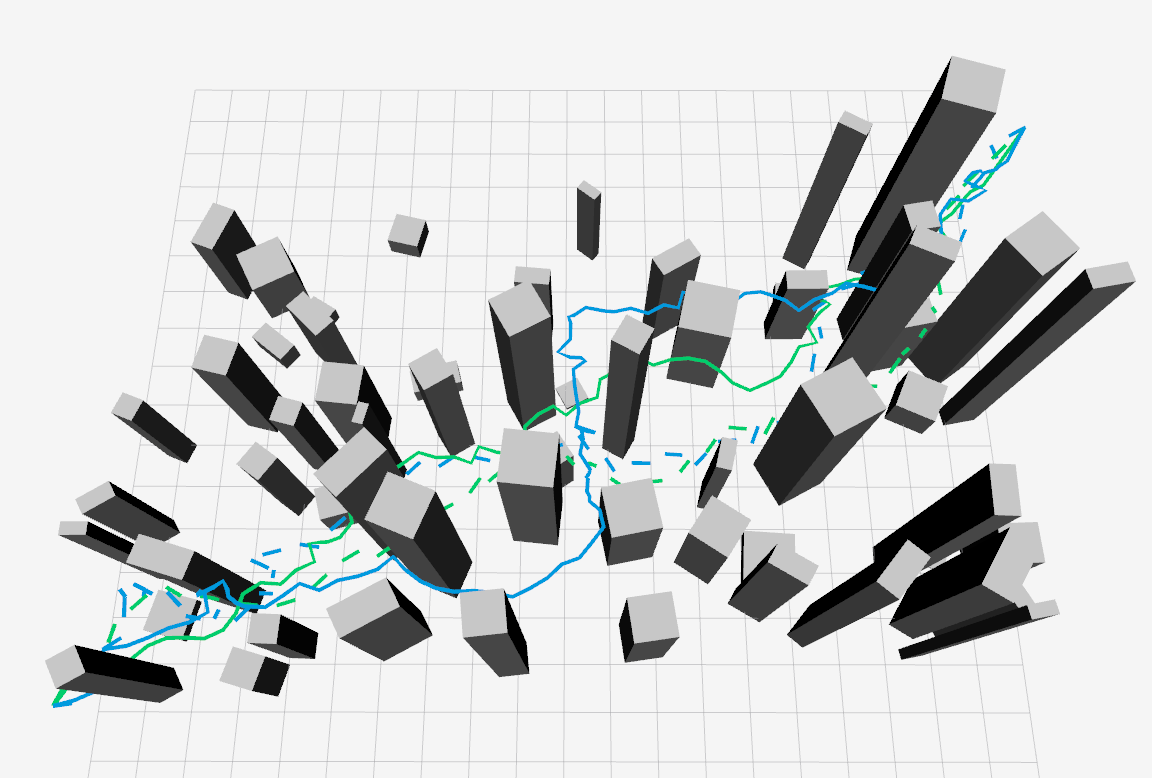}}      
     \endminipage \hfill
     \caption{Examples of CPW-PTOPP results imposed on different planner types in 3D space. $\mathbf{x}_0, \mathbf{x}_g$ are the same as \hyperref[Fig: 3D Planning Examples]{Figure \ref{Fig: 3D Planning Examples}}. (a)-(c) CPW-PTOPP is integrated into shortest path planning. (d)-(f) CPW-PTOPP is integrated into MPW-PTOPP and GPW-PTOPP.}
     \label{Fig: CPW-PTOPP Paths 3D}
\end{figure*}

Paths will go up and down to traverse wide passages in 3D space. The optimal paths prefer large heights because passages are sparse and wide there. This is exemplified in \hyperref[Fig: PTOPP Paths at High Heights]{Figure \ref{Fig: PTOPP Paths at High Heights}} where $\mathbf{x}_0$ and $\mathbf{x}_g$ are low. Optimal paths in PTOPP tend to go upwards first to reach a large height and then lead to the goal. By doing this, the paths only go through a few wide passages at large heights, contributing to a minimized path cost. In practice, this strategy may be undesired if certain path properties are expected. For instance, robots need to traverse different heights for information collection purposes in search tasks. To address this, the sampling strategy can be specified by limiting the admissible height range of samples. The sampled height range in \hyperref[Fig: 3D Planning Examples]{Figure \ref{Fig: 3D Planning Examples}} is $[0, 205]\T$ with the upper bound slightly larger than $\mathbf{x}_g$ height to prevent the optimal paths from going too high. Then, CPW-PTOPP is also tested in 3D space as in \hyperref[Fig: CPW-PTOPP Paths 3D]{Figure \ref{Fig: CPW-PTOPP Paths 3D}}. The impact of a constrained passage width is similar to planar cases. If imposed in shortest path planning, it avoids spatial passages narrower than the admissible width ($\underline{\varepsilon}_p = 60)$ as much as possible while being the shortest. If imposed in PTOPP, it minimizes the number of passed passages whose widths are smaller than $\underline{\varepsilon}_p$.   

\subsection{Computational efficiency of PTOPP}
At last, the computational efficiency of the proposed PTOPP algorithms is reported. Two major aspects are investigated: PTOPP efficiency compared with benchmark algorithms and previous avenues. The benchmarks include the shortest path planning and two common approaches to reaching large free space for paths, i.e., planning under a clearance constraint and to maximize clearance. Planning under a clearance constraint restricts paths inside the interior of $\mathcal{X}_{free}$. For simplicity, this is done by discarding samples of a clearance smaller than the threshold and assuming edges connecting valid samples lie in the interior. Maximizing clearance is achieved by setting the negative path clearance as a compatible path cost, namely $f_c(\sigma) = -\text{clr}(\sigma, \mathcal{X}_{obs})$, where $\text{clr}(\sigma, \mathcal{X}_{obs})$ designates the clearance between $\sigma$ and obstacles. GPW-PTOPP is run as a representative of PTOPP for the full use of all developed modules. The compared previous implementation is in \citep{huang2024homotopic} that has no sample positioning in Gabriel cells and needs to traverse all passages when checking the passed passages of edges. Planners are built both on PRM$^*$ and RRT$^*$.

The preprocessing time of passages and cells is negligible compared to the planning phase and is not taken into account. Each planning time is averaged across 30 random obstacle distributions for the same distant setup of $\mathbf{x}_0$ and $\mathbf{x}_g$. A plan is terminated when the valid sample number reaches $|V| = 2$k in uniform sampling. Clearance calculates the smallest distance between samples and obstacles, which can be rapidly performed using R-trees or geometric computations for simple shapes. A larger clearance constraint incurs more sampling trials to achieve $|V|$ because the feasible sampling space is reduced. To be fair, a relatively small clearance constraint is adopted (10 in tests). Another parameter influencing the planning time is the edge length $\eta$ in RRT$^*$ \texttt{Steer}() subroutine. It decides the range of near neighbors search. In PTOPP, a longer edge will traverse more cells in \texttt{PositionNode}(). A larger $\eta$ thus means increased computational load. We set $\eta = 50$ in 2D tests and $100$ in 3D tests so that they are of similar magnitudes of $r_{PRM^*}$ for better comparability.
\begin{table*}[t]
    \caption{Planners' average path planning time in 2D environments \textnormal{(ms)}. Data are collected from 30 trials for each obstacle number. Each trial is executed in the same environment. The total sample number in each plan is $|V| = 2$k.}
    \label{Tab: 2D Planning Time}
    \centering
    \small
    \begin{tabularx}{2 \columnwidth}{ l Y Y Y Y Y Y Y Y Y Y Y Y Y}
    \toprule
        Obstacle number & & &20 &40 &60 &80 &100 &120 &140 &160 &180 &200 \\
        \toprule         
        \multirow{5}{*}{PRM$^*$-based} &\multicolumn{2}{l}{shortest length} &1231 &2280	&3136	&3953	&4574	&5153	&5680	&6128	&6495	&6920 \\
                                             &\multicolumn{2}{l}{clearance constraint} &1329	&2525	&3551	&4603	&5446	&6301	&7148	&7898	&8495	&9313 \\ 
                                             &\multicolumn{2}{l}{maximum clearance} &1230	&2282	&3154	&3967	&4593	&5168	&5672	&6189	&6540	&6959 \\
                                             &\multicolumn{2}{l}{GPW-PTOPP traversal} &1488	&2750	&3814	&4794	&5554	&6215	&6757	&7378	&7672	&8198 \\
                                             &\multicolumn{2}{l}{GPW-PTOPP} &\textbf{1279}  &\textbf{2332} &\textbf{3203} &\textbf{4009} &\textbf{4649} &\textbf{5221} &\textbf{5732} &\textbf{6193} &\textbf{6577} &\textbf{6966} \\
        \toprule
        \multirow{5}{*}{RRT$^*$-based} &\multicolumn{2}{l}{shortest length} &411  &796  &1151  &1503  &1831  &2163  &2485 &2819 &3148 &3530 \\   
                                             &\multicolumn{2}{l}{clearance constraint} &450  &892  &1313  &1769  &2194  &2771  &3248  &4133 &4237  &5160 \\ 
                                             &\multicolumn{2}{l}{maximum clearance} &413	 &802  &1158  &1521	 &1840 
 &2194  &2537  &2860	 &3215	 &3601 \\
                                             &\multicolumn{2}{l}{GPW-PTOPP traversal} &769	&1492	&2183	&2860	&3488	&4128	&4688	&5325	&5942 &6623 \\
                                             &\multicolumn{2}{l}{GPW-PTOPP} &\textbf{488} &\textbf{878} &\textbf{1239} &\textbf{1595} &\textbf{1940} &\textbf{2281} &\textbf{2623} &\textbf{2948} &\textbf{3322} &\textbf{3667} \\
    \bottomrule         
    \end{tabularx}
\end{table*}
\begin{table*}[t]
    \caption{Average path planning time in 3D environments \textnormal{(ms)}.}
    \label{Tab: 3D Planning Time}
    \centering
    \small
    \begin{tabularx}{2 \columnwidth}{ l Y Y Y Y Y Y Y Y Y Y Y Y Y}
    \toprule
        Obstacle number & & &20 &40 &60 &80 &100 &120 &140 &160 &180 &200 \\
        \toprule         
        \multirow{5}{*}{PRM$^*$-based} &\multicolumn{2}{l}{shortest length} &3193	&5601	&7753	&9732	&11658	&13362	&15112	&16554	&18288	&19724 \\   
                                             &\multicolumn{2}{l}{clearance constraint} &3402	&5983	&8384	&10766	&13026	&15292	&17719	&19897	&22374	&24825 \\ 
                                             &\multicolumn{2}{l}{maximum clearance} &3198	&5580	&7744	&9793	&11700	&13420	&15192	&16608	&18284	&19858 \\
                                             &\multicolumn{2}{l}{GPW-PTOPP traversal} &3804	&6788	&9553	&12092	&14385	&16653	&18849	&20560	&22628	&24406 \\
                                             &\multicolumn{2}{l}{GPW-PTOPP} &\textbf{3537} &\textbf{5945} &\textbf{8143} &\textbf{10278} &\textbf{12152} &\textbf{13993} &\textbf{15759} &\textbf{17253} &\textbf{19094} &\textbf{20517} \\
        \toprule
        \multirow{5}{*}{RRT$^*$-based} &\multicolumn{2}{l}{shortest length} &813	&1464	&2099	&2734	&3356	&3969	&4626	&5337	&5967	&6646 \\   
                                             &\multicolumn{2}{l}{clearance constraint} &881	&1592	&2297	&3030	&3778	&4550	&5357	&6324	&7196	&8122 \\ 
                                             &\multicolumn{2}{l}{maximum clearance} &824	 &1479	&2119	&2769	&3426	&4037	&4729	&5376	&6093	&6781 \\
                                             &\multicolumn{2}{l}{GPW-PTOPP traversal} &1382	&2614	&3855	&5075	&6316	&7503	&8805	&9944	&11284	&12662 \\
                                             &\multicolumn{2}{l}{GPW-PTOPP} &\textbf{1182} &\textbf{1929} &\textbf{2667} &\textbf{3362} &\textbf{4087} &\textbf{4776} &\textbf{5567} &\textbf{6227} &\textbf{7039} &\textbf{7832} \\
    \bottomrule         
    \end{tabularx}
\end{table*}

\subsubsection{Planning efficiency with different obstacle numbers}
\begin{figure*}[t]
    \minipage{2 \columnwidth}
    \centering
    \subfigure[]{
    \includegraphics[width= 0.36 \columnwidth]{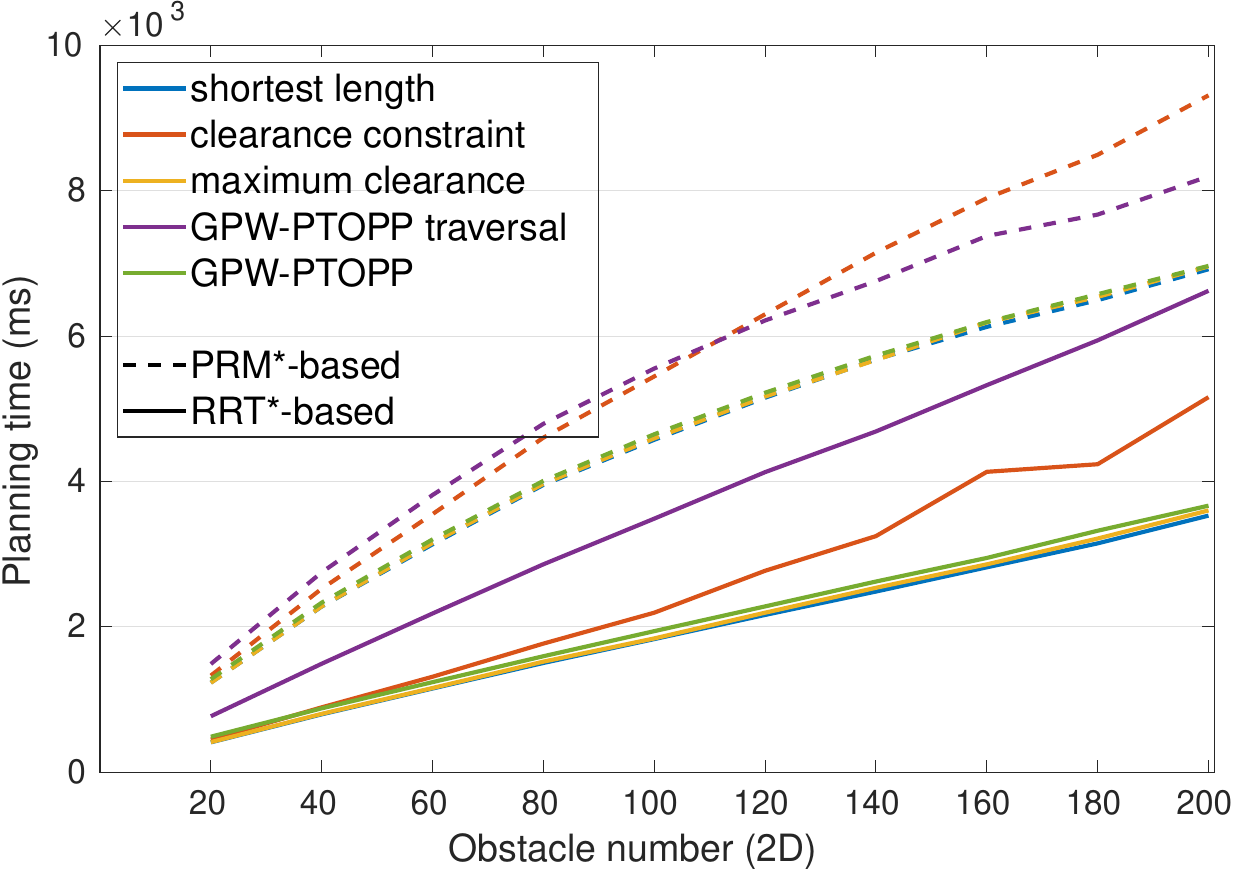}
    \label{Fig: RRTS Planning Time Obstacle 2D}
    }
    \;\;\;\;\;
    \subfigure[]{
    \includegraphics[width= 0.36 \columnwidth]{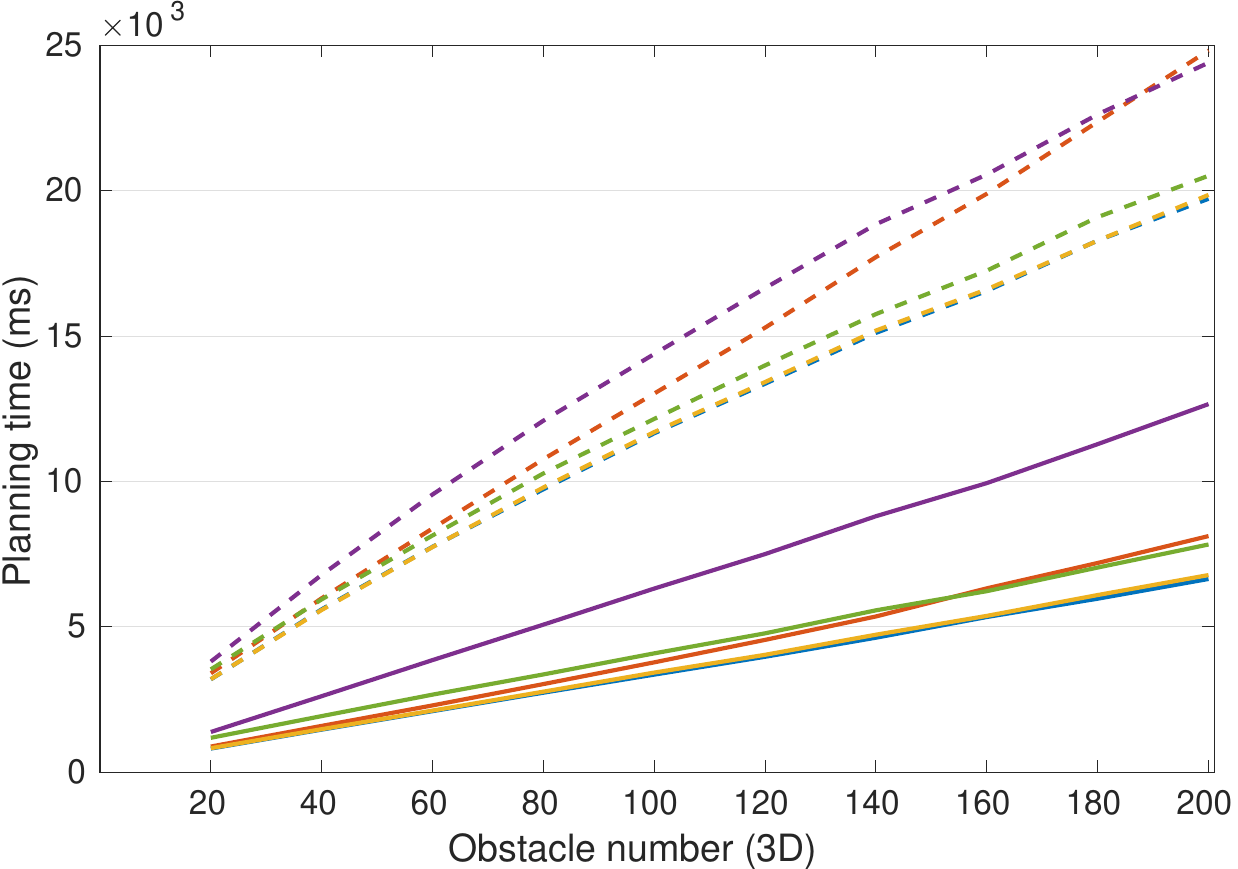}    
    \label{Fig: PRMS Planning Time Obstacle 2D}
    }  
    \endminipage \hfill    
    \caption{Average planning time with varying obstacle number $m$. Dashed and solid lines correspond to PRM$^*$-based and RRT$^*$-based algorithms, respectively. For each $m$, 30 trials in random obstacle distributions are conducted.}
    \label{Fig: Path Planning Time Obstacle}
\end{figure*}

Planning time with different numbers of obstacles is listed in \hyperref[Tab: 2D Planning Time]{Table \ref{Tab: 2D Planning Time}} and \ref{Tab: 3D Planning Time} and visualized in \hyperref[Fig: Path Planning Time Obstacle]{Figure \ref{Fig: Path Planning Time Obstacle}}.
The main results are twofold. The first is that PTOPP has comparable computational efficiency to benchmarks while achieving its advantages. The second is that the proposed algorithms address the high computational demands in the previous PTOPP implementation that has to traverse all passages in passage traversal check. More specifically, vanilla shortest path planning runs the fastest basically in all conditions. Maximum clearance planning takes a similar time as shortest path planning. It affords extra clearance computation for samples, which leads to an insignificant increase in planning time, e.g., less than \SI{2.1}{\percent} across tests. By contrast, imposing a clearance constraint results in obviously longer planning time. The clearance constraint reduces the feasible sampling spaces and more sampling trials are required to reach $|V|$. When obstacles are dense, the feasible sampling space turns particularly confined. In 2D space, the planning time increases \SI{9.5}{\percent}-\SI{46.2}{\percent} in RRT$^*$-based implementation and \SI{7.9}{\percent}-\SI{34.6}{\percent} in PRM$^*$-based implementation as $m$ gets large. In 3D space, the corresponding increases are \SI{8.5}{\percent}-\SI{22.2}{\percent} and \SI{6.5}{\percent}-\SI{25.9}{\percent}, respectively.

In \hyperref[Fig: Path Planning Time Obstacle]{Figure \ref{Fig: Path Planning Time Obstacle}}, the proposed PTOPP algorithms manifest computational efficiency close to the shortest path planning and maximum clearance planning, whereas direct passage traversal check is much more time-consuming. RRT$^*$-based PTOPP using passage traversal has nearly twice the planning time of the shortest path planning baseline in both 2D and 3D space. In PRM$^*$-based PTOPP, increases in running time are also dramatic. This presents a bottleneck in PTOPP computational efficiency, hindering its applicability under limited computational resources, e.g., in onboard systems. The proposed algorithms address this issue effectively. Overall, there is no significant planning time rises in PTOPP. The average planning time increments of PTOPP over the shortest path planning baseline are \SI{1.7}{\percent} in PRM$^*$ and \SI{7.4}{\percent} in RRT$^*$ for 2D space, and \SI{5.3}{\percent} and \SI{24.2}{\percent} respectively for 3D space, substantially lower than direct passage traversal. In brief, PTOPP achieves comparable computational efficiency to the baseline while realizing its advantages in accessible free space optimization. Compared to existing methods, it has similar efficiency with maximum clearance planning and mostly outperforms clearance constraint planning, especially amid dense obstacles.

\subsubsection{Planning efficiency with different valid sample numbers and cost evolution}
\begin{figure*}[t]
    \minipage{2 \columnwidth}
    \centering
    \subfigure[]{
    \includegraphics[width= 0.38 \columnwidth]{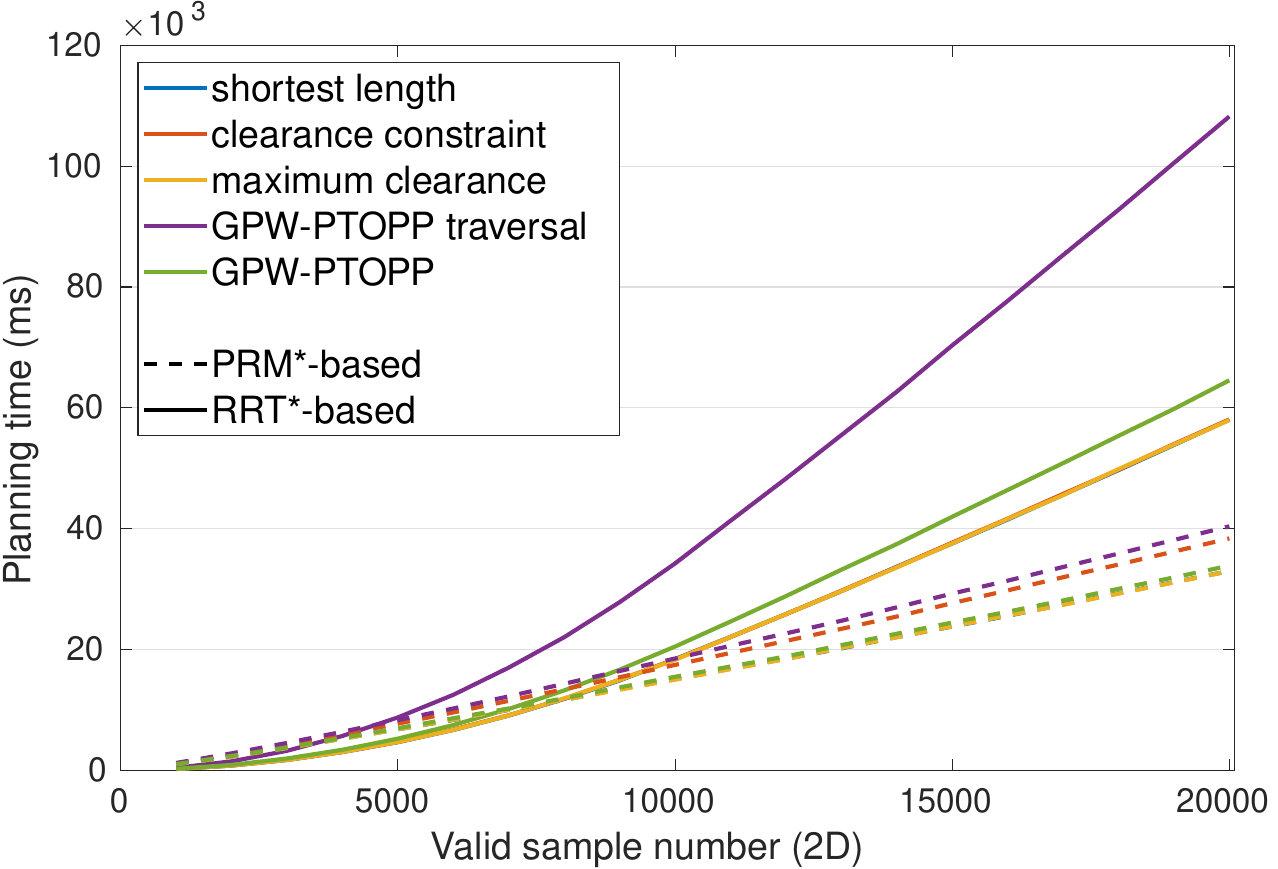}
    \label{Fig: RRTS Planning Time Sample 2D}
    }
    \;\;\;
    \subfigure[]{
    \includegraphics[width= 0.38 \columnwidth]{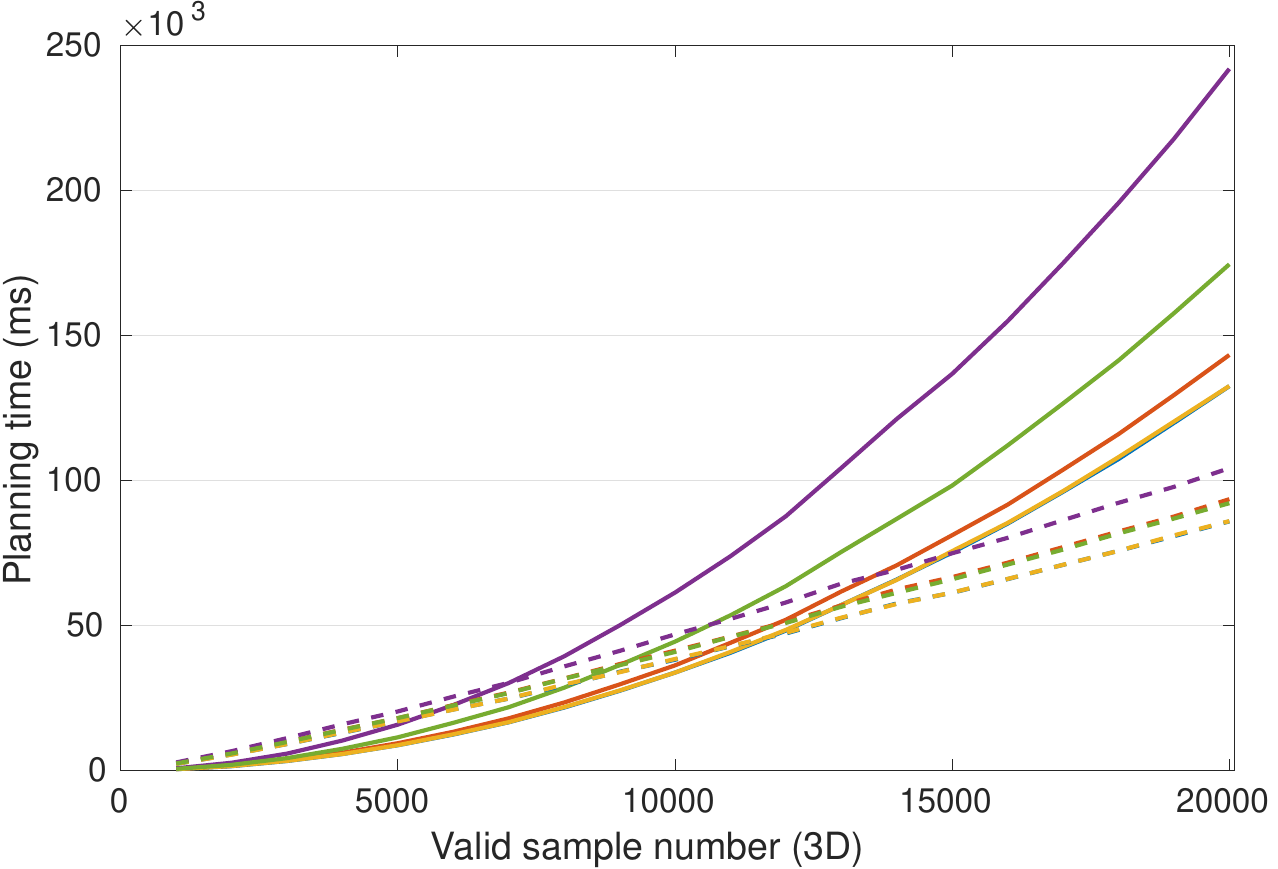}  
    \label{Fig: PRMS Planning Time Sample 2D}
    }  
    \endminipage \hfill    
    \caption{Average path planning time with varying valid sample number $|V|$. For each $|V|$, five trials in the same environment with 40 obstacles are conducted.}
    \label{Fig: Path Planning Time Sample}
\end{figure*}

Planning time with varying valid sample numbers $|V|$ is shown in \hyperref[Fig: Path Planning Time Sample]{Figure \ref{Fig: Path Planning Time Sample}}. The obstacle number $m = 40$ and the planning time is averaged over five runs. PTOPP using passage traversal takes significantly more time than other planners as $|V|$ increases, while the proposed PTOPP remarkably reduces the time expense. In PRM$^*$-based planners, the proposed method has planning time shorter than or similar to clearance constraint and maximum clearance planning. In RRT$^*$-based planners, it consumes slightly more time than benchmarks only when $|V|$ gets extremely large. As for planning time trends, RRT$^*$-based planners exhibit stronger linearity w.r.t. $m$, whereas PRM$^*$-based planners exhibit stronger linearity w.r.t. $|V|$. PRM$^*$-based planners run faster when $|V|$ gets large. 

\begin{figure*}[t]
    \minipage{2 \columnwidth}
    \centering
    \subfigure[]{
    \includegraphics[width= 0.36 \columnwidth]{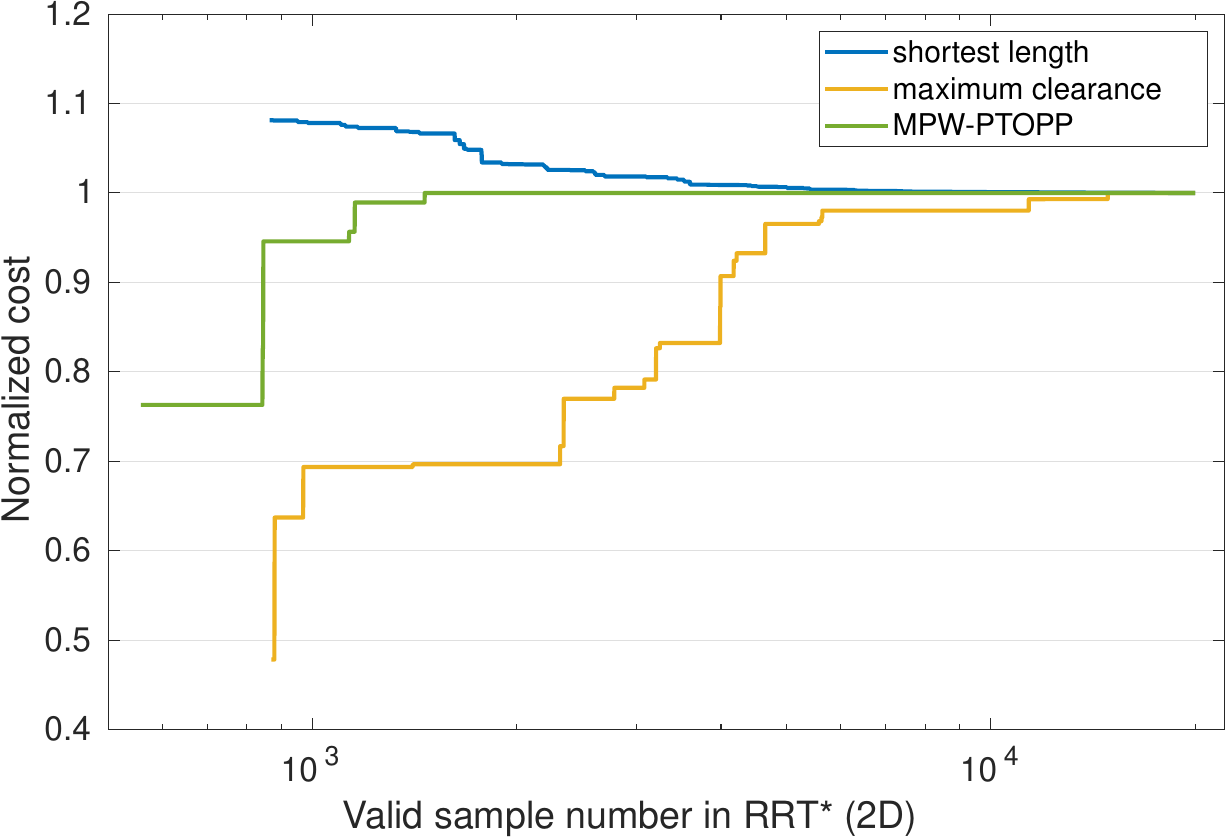}
    \label{Fig: Cost Evolution 2D RRTS}
    }
    \;\;\;\;\;
    \subfigure[]{
    \includegraphics[width= 0.36 \columnwidth]{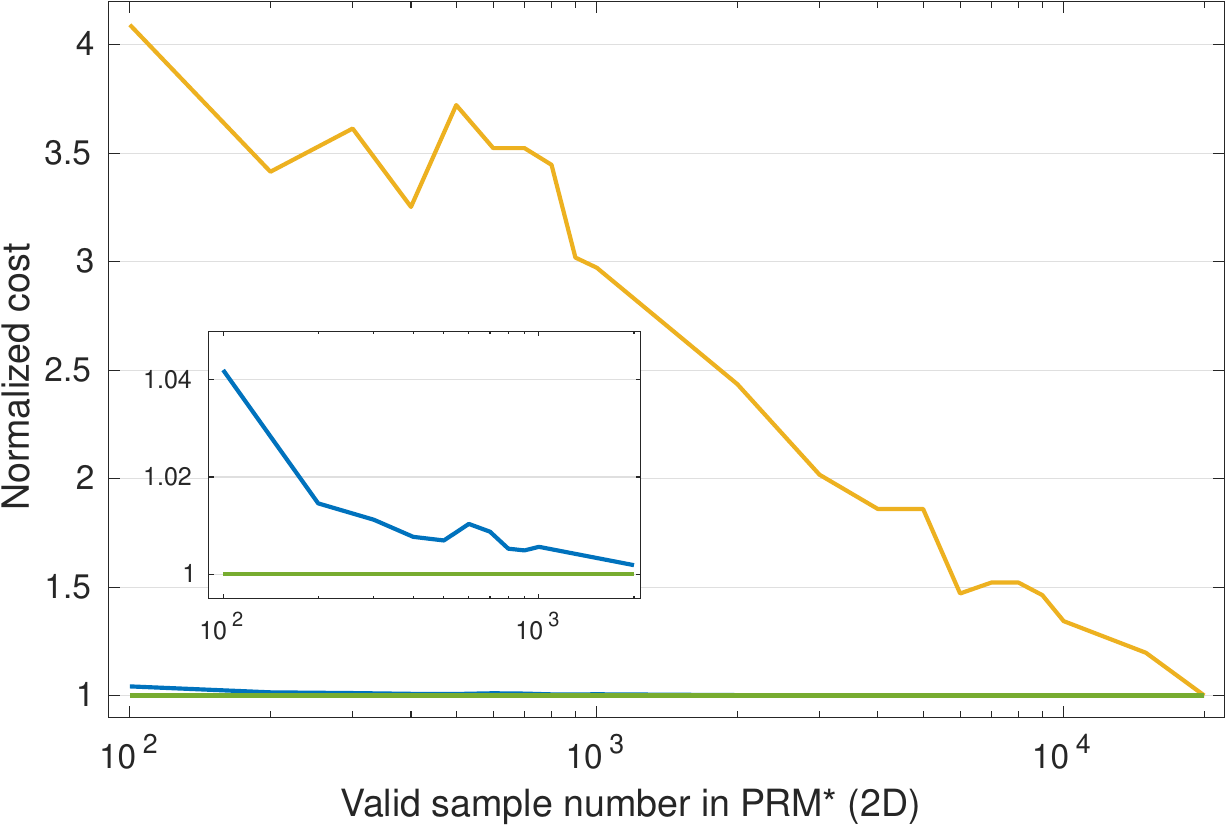}    
    \label{Fig: Cost Evolution 2D PRMS}
    }
    \subfigure[]{
    \includegraphics[width = 0.36 \columnwidth]{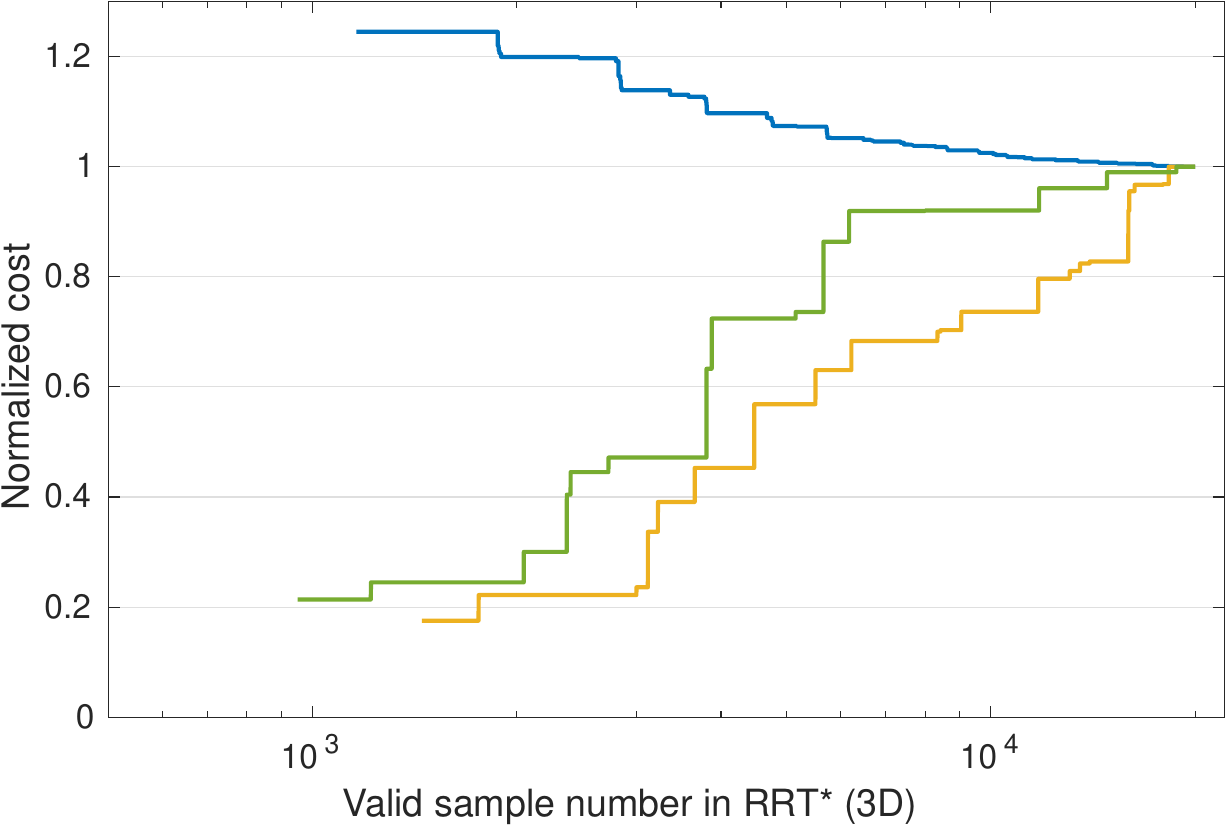}
    \label{Fig: Cost Evolution 3D RRTS}
    }
    \;\;\;\;\;
    \subfigure[]{
    \includegraphics[width = 0.36 \columnwidth]{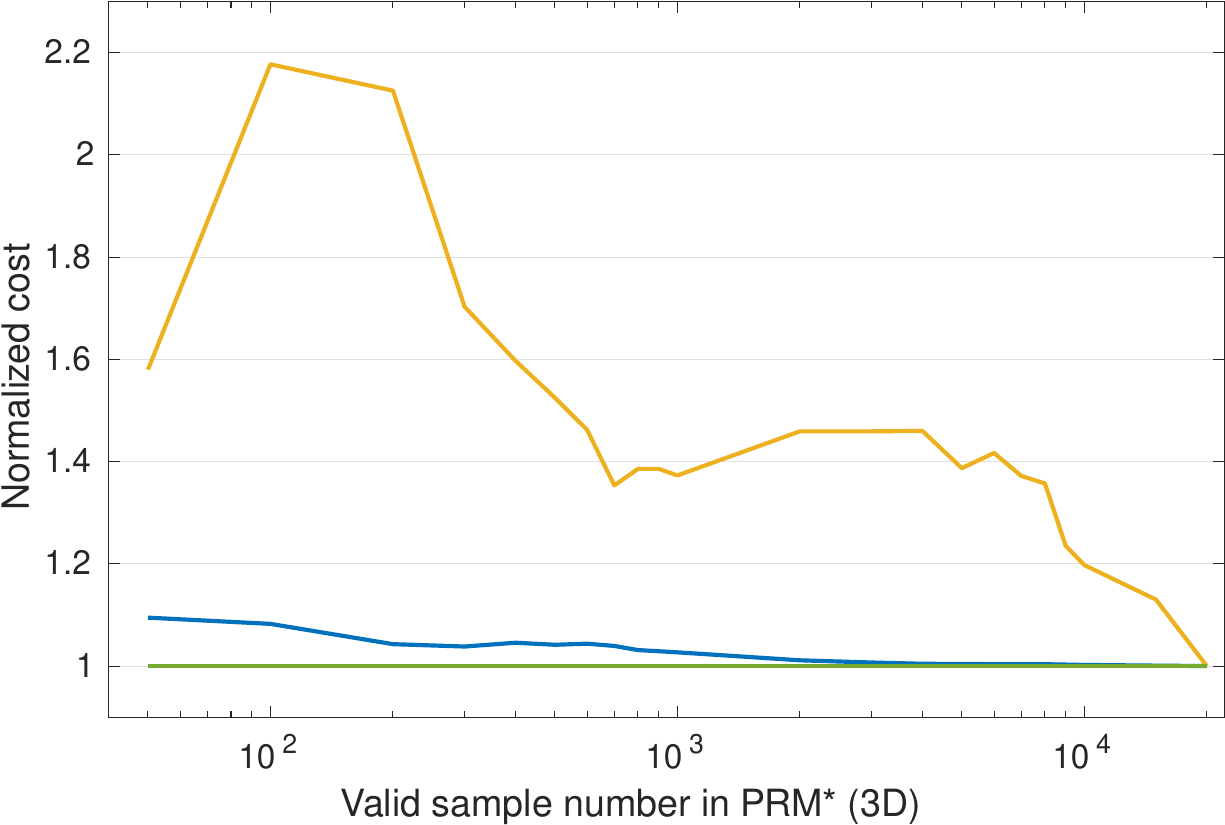}
    \label{Fig: Cost Evolution 3D PRMS}
    }    
    \endminipage \hfill
     \caption{Cost of the best path versus the number of valid samples $|V|$ in different planners and configuration spaces. The values are normalized so that the cost of the final optimal path is one.}
     \label{Fig: Cost Evolution}
\end{figure*}

Lastly, \hyperref[Fig: Cost Evolution]{Figure \ref{Fig: Cost Evolution}} illustrates the evolution of costs of the best paths during planning to highlight the fast convergence of PTOPP. Because MPW-PTOPP and maximum clearance planning find homotopic optimal paths in 2D space, they are compared along with the shortest path planning baseline. MPW-PTOPP converges to the optimal cost much faster than others due to the discreteness of its cost. Finite cost values exist in PTOPP and thus are explored quickly, as illustrated by stages in \hyperref[Fig: Cost Evolution 2D RRTS]{Figure \ref{Fig: Cost Evolution 2D RRTS}} and \hyperref[Fig: Cost Evolution 3D RRTS]{\ref{Fig: Cost Evolution 3D RRTS}}. Clearance also gets updated periodically in RRT$^*$, but is less efficiently explored for its value continuity. PRM$^*$-based planners demonstrate faster convergence than RRT$^*$ counterparts because random roadmaps can explore a much larger configuration space partition than random trees do in RRT$^*$ given the same number of samples. Random trees only grow from leaves. Note that the maximum clearance in PRM$^*$ may decrease with the sample number increasing. This counterintuitive phenomenon is attributed to that the near neighbor search radius $r_{PRM^*}$ is set to decrease with samples. Samples of large clearance may not be reachable from path nodes if $r_{PRM^*}$ is small. 

\section{Discussion and conclusion}
Passages provide advantages over clearance in measuring accessible free space. Many approaches to enlarging path clearance, such as path search in Voronoi graphs, are unable to maximize clearance globally. Though the maximum clearance is attainable in maximum clearance planning, the degradation problem exists. Moreover, path clearance cannot be optimized as flexibly as discrete passage widths in PTOPP. Passage graphs offer a new perspective to organize objects that has potential in scene understanding. Narrow passage regions may also be used to guide sampling for efficient path finding. The Gabriel condition is proposed for passage detection because of its intuitive spatial interpretation in robotics contexts. Another possible choice of proximity graph is the relative neighborhood graph (RNG), a subgraph of the Gabriel graph \citep{toussaint1980relative, melchert2013percolation}. RNG maintains the shortest edges of Delaunay triangles, imposing a stricter passage criterion than the Gabriel condition. Beyond 2D and 3D workspace, the concept of passages is promising in higher dimensional spaces, e.g., the joint space, in describing confined configurations. But characterization of subjects, e.g., obstacles, in high dimensional space is complicated \citep{dai2024certified}.

Accessible free space optimization for paths is studied in this article as one most direct and important application of PTOPP. Other PTOPP problem instances exist since PTOPP has good configurability and applicability to formulate and solve various problems related to passages. For example, passages may be abstracted as toll stations that incur costs to pass. The principles of compatible path costs provide important guidance for cost design in OPP. The presented algorithmic framework of PTOPP is general and aligns with sampling-based optimal planners. Thus, common techniques to accelerate planning and improve the solution quality, such as bidirectional planning and biasing samples to the goal region, are applicable. Apart from sampling-based methods, pure cell-based path planning methods are enabled by Gabriel cells. They can reach a much faster planning speed with no need for a large number of random samples. Their dependent procedures, however, have been covered in the proposed algorithms.

This article presents PTOPP as a new path planning paradigm able to optimize traversed passages of paths. Accessible free space optimization is investigated as an important instantiation for which the entire PTOPP pipeline is developed systematically. In preprocessing of environments, passage detection and free space decomposition based on Gabriel graphs of obstacles are introduced. This enables fast detection of sparse and informative passage distributions and cell decompositions in 2D and 3D space. Typical PTOPP categories defined by path costs are then formulated, in which the compatibility requirements for path costs in sampling-based optimal algorithms are addressed. Next, the primitive procedures PTOPP relies on are elaborated. The core module for planning efficiency is rapid passage traversal check achieved by positioning samples in partitioned cells. Finally, sampling-based optimal algorithms for PTOPP are reported with detailed analysis of probabilistic completeness, asymptotic optimality, and computational complexity. Extensive experimental results verify the effectiveness and efficiency of PTOPP. It not only admits more versatile solutions to various planning problems but also demonstrates superior or comparable efficiency over existing methods. PTOPP offers new insights and tools to resolve fundamental path planning requirements not limited to free space accessibility.
Important future directions include extending current methods to more challenging planning setups, e.g, dynamic environments and higher dimensional configuration spaces, developing theories and planning methods for non-compatible path cost optimization, and deploying PTOPP in downstream robotic tasks, e.g., manipulation planning and navigation in confined environments.

\bibliographystyle{sageh}
\bibliography{references.bib}

\appendix
\section{Appendix}
\subsection{Geodesic distance traversed in Delaunay graphs in passage detection}
\label{Appendix: Geodesic Distance Test}
When extracting the Gabriel graph from a Delaunay graph, it suffices to check all the edges. For edge $(\mathbf{x}_i, \mathbf{x}_j)$, the graph nodes $\mathbf{x}_k$ that possibly violate the Gabriel condition are limited to adjacent nodes of $\mathbf{x}_i$ and $\mathbf{x}_j$. When determining passages from the Delaunay graph $\mathcal{DG}(C_c)$ constructed from the obstacle centroids $C_c$, solely examining edges may miss passages because obstacles are volumetric. The solution is to increase the check range of geodesic distance. The checked geodesic distance $k_{gd}$ in $\mathcal{DG}(C_c)$ is described in (\ref{Eqn: Smallest Geodesic Distance}) and adopted as two in our implementation of (\ref{Eqn: Amended Gabriel Condition Within Geodesic Distances}), which proves to be adequate to find all passages under test conditions. Major factors influencing $k_{gd}$ include obstacle dimensions and distributions. The more obstacles can be treated as points (e.g., with small sizes and in large environments), the smaller $k_{gd}$ is.

Here we report detected passage data in various setups with $k_{gd} = 1$ and $2$. It empirically verifies that a small $k_{gd} > 1$ suffices to detect all passages. As demonstrated in \hyperref[Tab: Passage Geodesic Distance Data]{Table \ref{Tab: Passage Geodesic Distance Data}} and \ref{Tab: Passage Geodesic Distance Percentage Data}, when obstacle sizes have limited variations, $k_{gd} = 2$ reports all passages, e.g., for the side length range $[20, 60]$ used in experiments. As expected, only checking edges of $\mathcal{DG}(C_c)$, i.e., $k_{gd} = 1$, fails to find all passages mostly. If obstacle sizes have extreme variations, $k_{gd} = 2$ may also fail. When the obstacle side length range is $[1, 60]$, obstacle sizes may differ by thousands of times. Failure to detect all passages occurs occasionally for $k_{gd} = 2$, especially when obstacles are dense. Importantly, only a negligible number of passages are missed in failure cases as listed in \hyperref[Tab: Passage Geodesic Distance Percentage Data]{Table \ref{Tab: Passage Geodesic Distance Percentage Data}}. This is acceptable in most practical scenes. To find all passages, $k_{gd}$ only needs to be larger than two. $k_{gd}$ does not change the detection complexity because the expected degree of a Delaunay graph node is $O(1)$ and at most six \citep{mark2008computational}. The computational efficiency is guaranteed to outperform the brute-force check strategy.

\subsection{Compatible GPW-PTOPP path cost design}
\label{Appendix: Path Cost Compatibility}
In general, the cost form in (\ref{Eqn: Weighted GPW-PTOPP Cost}) is not order-preserving as a sum of paired weight and width products. The design objective is to make the order of path costs consistent with the reversed lexicographic order of passage width vectors by setting appropriate weights. The lexicographic order of passage vectors is preserved in path concatenation since
\begin{equation}
\label{Apx Eqn: Insertion of Passage Vector} 
    \mathbf{p}_{\sigma_1} \prec_{\text{lex}} \mathbf{p}_{\sigma_2} \Rightarrow \mathbf{p}_{\sigma_1} \sqcup \{x\} \prec_{\text{lex}} \mathbf{p}_{\sigma_2} \sqcup \{x\}
\end{equation}
where $\sqcup$ stands for inserting a new entry into one sorted vector while maintaining the sorted order. This is easily verified by comparing the new element's insertion positions in two vectors. 
\begin{table}[t]
    \caption{Number of trails that report all passages. For a given obstacle number, 30 trials on random obstacle distributions are conducted. The environment dimension is $1000 \times 600$.}
    \label{Tab: Passage Geodesic Distance Data}
    \centering
    \small
    \begin{tabularx}{1 \columnwidth}{l l Y Y Y Y Y}
    \toprule
         Obstacle number & &40 &80 &120 &160 &200  \\
        \toprule         
        \multirow{2}{*}{Side in [20, 60]} &\mbox{$k_{gd} = 1$} &18  &1 &1 &0 &0 \\   
                                              &\mbox{$k_{gd} = 2$} &30  &30 &30 &30 &30\\ 
        \toprule
        \multirow{2}{*}{Side in [1, 60]} &\mbox{$k_{gd} = 1$}  &13  &1 &0 &0 &0 \\   
                                                &\mbox{$k_{gd} = 2$}  &30  &29 &28 &27 &26\\ 
    \bottomrule         
    \end{tabularx}
\end{table}
\begin{table}[t]
    \caption{Percentages of the average detected passage number in the average total passage number (\SI{}{\percent}).}
    \label{Tab: Passage Geodesic Distance Percentage Data}
    \centering
    \small
    \begin{tabularx}{1 \columnwidth}{l l Y Y Y Y Y}
    \toprule
         Obstacle number & &40 &80 &120 &160 &200  \\
        \toprule         
        \multirow{2}{*}{Side in [20, 60]} &\mbox{$k_{gd} = 1$} &99.26  &98.64 &98.67 &98.74 &98.31 \\   
                                              &\mbox{$k_{gd} = 2$} &100  &100 &100 &100 &100\\ 
        \toprule  
        \multirow{2}{*}{Side in [1, 60]} &\mbox{$k_{gd} = 1$}  &98.15  &97.47 &96.13 &96.32 &95.61 \\   
                                                &\mbox{$k_{gd} = 2$}  &100  &99.98 &99.97 &99.97 &99.97\\ 
    \bottomrule         
    \end{tabularx}
\end{table}

$\mathbf{p}_\sigma$ is fixed sized in PTOPP and the insertion above is accomplished by first inserting the new element and then truncating the last and also the largest element. To ensure the transition in (\ref{Eqn: Lexical Order to Cost}) for path cost (\ref{Eqn: Weighted GPW-PTOPP Cost}), weight values in $\mathbf{w}_p$ are analyzed recursively. Start with the basic case of 2-GPW-PTOPP, i.e., $\textit{k} = 2$. Let $\mathbf{p}_{\sigma_1} = [p_{1,1}, p_{1,2}]\T, \mathbf{p}_{\sigma_2} = [p_{2,1}, p_{2,2}]\T, \mathbf{w}_p = [w_1, w_2]\T$ and $\mathbf{p}_{\sigma_1} \prec_{\text{lex}} \mathbf{p}_{\sigma_2}$. $f_c(\sigma_1) > f_c(\sigma_2)$ implies that
\begin{equation}
\label{Apx Eqn: Basic Case Cost Order} 
    w_1 p_{1,1} + w_2 p_{1,2} <  w_1 p_{2,1} + w_2 p_{2,2}.
\end{equation}
$p_{1,1} = p_{2,1}, p_{1,2} < p_{2,2}$ is trivial. If $p_{1,1} < p_{2,1}$, the full constraint on weights is expressed as
\begin{equation}
\label{Apx Eqn: Weight Constraints in Basic Case}
\begin{split}
    & \ \ \ \ \ \ \ \ \ \frac{w_1}{w_2} > \frac{p_{1,2} - p_{2,2}}{p_{2,1} - p_{1,1}} \\
    \st  \; &p_{1,1} < p_{1, 2}, \, p_{2,1} < p_{2, 2}, \, p_{1,1} < p_{2, 1}, \\
           & p_{1, 1}, p_{1,2}, p_{2,1}, p_{2,2} \in \mathbb{P}, \\
           & w_1, w_2 \in \mathbb{R}_+.
\end{split}
\end{equation}
The elements in $\mathbf{p}_{\sigma}$ are picked from the finite passage width set $\mathbb{P}$. (\ref{Apx Eqn: Weight Constraints in Basic Case}) shows that the ratio between consecutive weight items $w_{p, i+1} / w_{p, i}$ needs to be large enough. In implementation, it is convenient to utilize the following scaled ratio
\begin{equation}
\label{Apx Eqn: Scaled Weight Ratio}
    \frac{w_1}{w_2} > \frac{\max \Delta p}{\min \Delta p} \geq \frac{p_{1,2} - p_{2,2}}{p_{2,1} - p_{1,1}}
\end{equation}
where $\Delta p$ represents the absolute difference between passage widths, i.e., $\Delta p = |p_i - p_j|, p_i, p_j \in \mathbb{P}, i \neq j$.

Extending to $\textit{k} = 3$, a similar ratio constraint on weights is derived. If $p_{1,1} = p_{2,1}$, the constraint in (\ref{Apx Eqn: Weight Constraints in Basic Case}) applies to $w_2, w_3$ except that choices of passage widths decrease by one because $p_{1,1} = p_{2,1}$ is removed. But the scaled value in (\ref{Apx Eqn: Scaled Weight Ratio}) is still applicable to set $w_2 / w_3 > \max \Delta p / \min \Delta p$. If $p_{1, 1} < p_{2, 1}$, the following inequality holds 
\begin{equation}
\label{Apx Eqn: Weight Constraints k = 3}
    \frac{w_1}{w_2} > \frac{p_{1,2} - p_{2,2}}{p_{2,1} - p_{1,1}} + \frac{w_3}{w_2} \frac{p_{1,3} - p_{2,3}}{p_{2,1} - p_{1,1}}. 
\end{equation}
Again, using the scaled $\Delta p$ ratio, $w_1 / w_2$ can be set as 
\begin{equation}
\label{Apx Eqn: Scaled Weight Constraints k = 3}
    \frac{w_1}{w_2} > \frac{\max \Delta p}{\min \Delta p} + 1 > \frac{p_{1,2} - p_{2,2}}{p_{2,1} - p_{1,1}} + \frac{w_3}{w_2} \frac{p_{1,3} - p_{2,3}}{p_{2,1} - p_{1,1}}. 
\end{equation}
Similarly, $w_1/ w_2 > \max \Delta p / \min \Delta p + 2$ suffices when $\textit{k} = 4$. By induction, the ratio between consecutive weights can be assigned as follows 
\begin{equation}
\label{Apx Eqn: Consecutive Weights Ratio Inequality}
    \frac{w_i}{w_{i + 1}} > \frac{\max \Delta p}{\min \Delta p} + k - 1 - i, \, i = 1, 2 ..., k - 1. 
\end{equation}
By this means, the reversed lexical orders of passage width vectors are maintained in (\ref{Eqn: Weighted GPW-PTOPP Cost}). The ratios between contiguous weight items are required to be sufficiently large. In practice, any large ratio that meets (\ref{Apx Eqn: Consecutive Weights Ratio Inequality}) works. Hence, it suffices to use $w_{p, i} / w_{p, i+1} \gg 1$ in GPW-PTOPP.

There are other possible constructions of compatible GPW-PTOPP path costs, such as $f_c(\sigma) = -\sum p_i$, $p_i \in \mathbf{p}_\sigma$ and its combination with the path length $f_c = \text{Len}(\sigma) - \sum p_i$. They have simple structures and good interpretability. But the problem is that while these costs favor wide passages, they are unable to prioritize avoiding narrow passages. For example, $\mathbf{p}_\sigma = [1, 9]\T$ leads to the same cost as $\mathbf{p}_\sigma = [5, 5]\T$ though it contains a much smaller passage width. For this reason, they are not adopted in this paper for accessible free space optimization purposes.

\end{document}